%% file: main.tex
\documentclass[10pt,journal,compsoc]{IEEEtran}
\ifCLASSOPTIONcompsoc
  \usepackage{cite}
\else
  \usepackage{cite}
\fi

\ifCLASSINFOpdf
\else
\fi

\usepackage{url}
\usepackage{times}
\usepackage{epsfig}
\usepackage{graphicx}
\usepackage{amsmath}
\usepackage{amssymb}
\usepackage{multirow}
\usepackage{colortbl}
\usepackage{color}
\usepackage{threeparttable}
\usepackage{booktabs}
\usepackage{adjustbox}
\usepackage{subfig}
\usepackage{caption}
\usepackage[misc]{ifsym}
\usepackage{xcolor,colortbl}
\usepackage{makecell}
\usepackage{tikz}
\usepackage[edges]{forest}
\definecolor{hidden-draw}{RGB}{20,68,106}
\definecolor{hidden-pink}{RGB}{255,245,247}
\usepackage{amssymb}
\usepackage{placeins}

\usepackage[normalem]{ulem}

\usepackage{cancel}

\usepackage{hyperref}
\usepackage{url}
\usepackage{color}
\usepackage{colortbl}
\usepackage{soul}
\usepackage{multirow}

\usepackage{rotating,siunitx} %
\usepackage{tabularx}
\usepackage{hyphenat}
\usepackage{longtable}

\usepackage{float}

\usepackage{etoolbox}
\usepackage{algorithm}      
\usepackage[noend]{algpseudocode} 

\usepackage{pifont}

\definecolor{LightRed}{rgb}{1,0.92,0.92}
\definecolor{LightOrange}{rgb}{1,0.95,0.88}
\definecolor{LightYellow}{rgb}{1.0,1.0,0.84}
\definecolor{LightGreen}{rgb}{0.9,1.0,0.88}
\definecolor{LightCyan}{rgb}{0.9,1,1}
\definecolor{LightBlue}{rgb}{0.9,0.94,1}
\definecolor{LightIndigo}{rgb}{0.92,0.9,1}
\definecolor{LightMagenta}{rgb}{0.96,0.86,1}
\definecolor{DirtyWhite}{rgb}{0.96,0.96,0.96}
\definecolor{Training}{rgb}{0.9568627450980393, 0.7803921568627451, 0.6901960784313725}
\definecolor{Finetuning}{rgb}{0.6274509803921569, 0.8470588235294118, 0.9372549019607843}
\definecolor{Free}{rgb}{0.596078431372549, 0.8509803921568627, 0.5568627450980392}
\newcommand{\TrainingColor}{60}
\newcommand{\FinetuningColor}{40}
\newcommand{\FreeColor}{40}

\usepackage{printlen}
\uselengthunit{pt}

\begin{document}

\title{Diffusion Model-Based Image Editing: A Survey}

\author{Yi Huang$^*$, Jiancheng Huang$^*$, Yifan Liu$^*$, Mingfu Yan$^*$, Jiaxi Lv$^*$, Jianzhuang Liu$^*$,~\IEEEmembership{Senior Member,~IEEE},\\
Wei Xiong, He Zhang, Liangliang Cao,~\IEEEmembership{Fellow,~IEEE}, and Shifeng Chen
\IEEEcompsocitemizethanks{
\IEEEcompsocthanksitem 
Y. Huang, J. Huang, M. Yan, and J. Lv are with Shenzhen Institute of Advanced Technology, Chinese Academy of Sciences, Shenzhen, China, and also with University of Chinese Academy of Sciences, Beijing, China. (E-mail: yi.huang@siat.ac.cn)
\IEEEcompsocthanksitem 
Y. Liu is with Southern University of Science and Technology, Shenzhen, China, and also with Shenzhen Institute of Advanced Technology, Chinese Academy of Sciences, Shenzhen, China.
\IEEEcompsocthanksitem 
J. Liu and S. Chen are with Shenzhen Institute of Advanced Technology, Chinese Academy of Sciences, Shenzhen, China. (E-mail: jz.liu@siat.ac.cn, shifeng.chen@siat.ac.cn)
\IEEEcompsocthanksitem W. Xiong is with NVIDIA, Santa Clara, USA.
\IEEEcompsocthanksitem H. Zhang is with Adobe Inc, San Jose, USA.
\IEEEcompsocthanksitem L. Cao is with Apple Inc, Cupertino, USA.
\IEEEcompsocthanksitem $^*$ denotes equal contributions. S. Chen is the corresponding author.}
}


\markboth{IEEE Transactions on Pattern Analysis and Machine Intelligence}%
{Shell \MakeLowercase{\textit{et al.}}: Bare Demo of IEEEtran.cls for Computer Society Journals}

\IEEEtitleabstractindextext{%
\begin{abstract}
Denoising diffusion models have emerged as a powerful tool for various image generation and editing tasks, facilitating the synthesis of visual content in an unconditional or input-conditional manner. The core idea behind them is learning to reverse the process of gradually adding noise to images, allowing them to generate high-quality samples from a complex distribution. In this survey, we provide an exhaustive overview of existing methods using diffusion models for image editing, covering both theoretical and practical aspects in the ﬁeld. 
We delve into a thorough analysis and categorization of these works from multiple perspectives, including learning strategies, user-input conditions, and the array of specific editing tasks that can be accomplished.
In addition, we pay special attention to image inpainting and outpainting, and explore both earlier traditional context-driven and current multimodal conditional methods, offering a comprehensive analysis of their methodologies.
To further evaluate the performance of text-guided image editing algorithms, we propose a systematic benchmark, EditEval, featuring an innovative metric, LMM Score.
Finally, we address current limitations and envision some potential directions for future research. The accompanying repository is released at \href{https://github.com/SiatMMLab/Awesome-Diffusion-Model-Based-Image-Editing-Methods}{https://github.com/SiatMMLab/\\Awesome-Diffusion-Model-Based-Image-Editing-Methods}.
\end{abstract}

\begin{IEEEkeywords}
Diffusion Model, Image Editing, AIGC
\end{IEEEkeywords}}

\maketitle

\IEEEdisplaynontitleabstractindextext

\IEEEpeerreviewmaketitle

\input{sec/sec1_intro}

\input{sec/sec2_background}

\input{sec/sec3_overview}

\input{sec/sec4_train}

\input{sec/sec5_finetune}

\input{sec/sec6_free}

\input{sec/sec7_inpainting}

\input{sec/sec8_benchmark}

\input{sec/sec9_future}








\ifCLASSOPTIONcaptionsoff
  \newpage
\fi

{\small
		\bibliographystyle{ieeetr}
		\bibliography{main}
}

\vfill


\end{document}

%% file: sec/sec1_intro.tex

 \begin{figure}[!t]
		\small
		\setlength{\abovecaptionskip}{-0.1cm}
		\centering
		\includegraphics[width=0.46\textwidth]{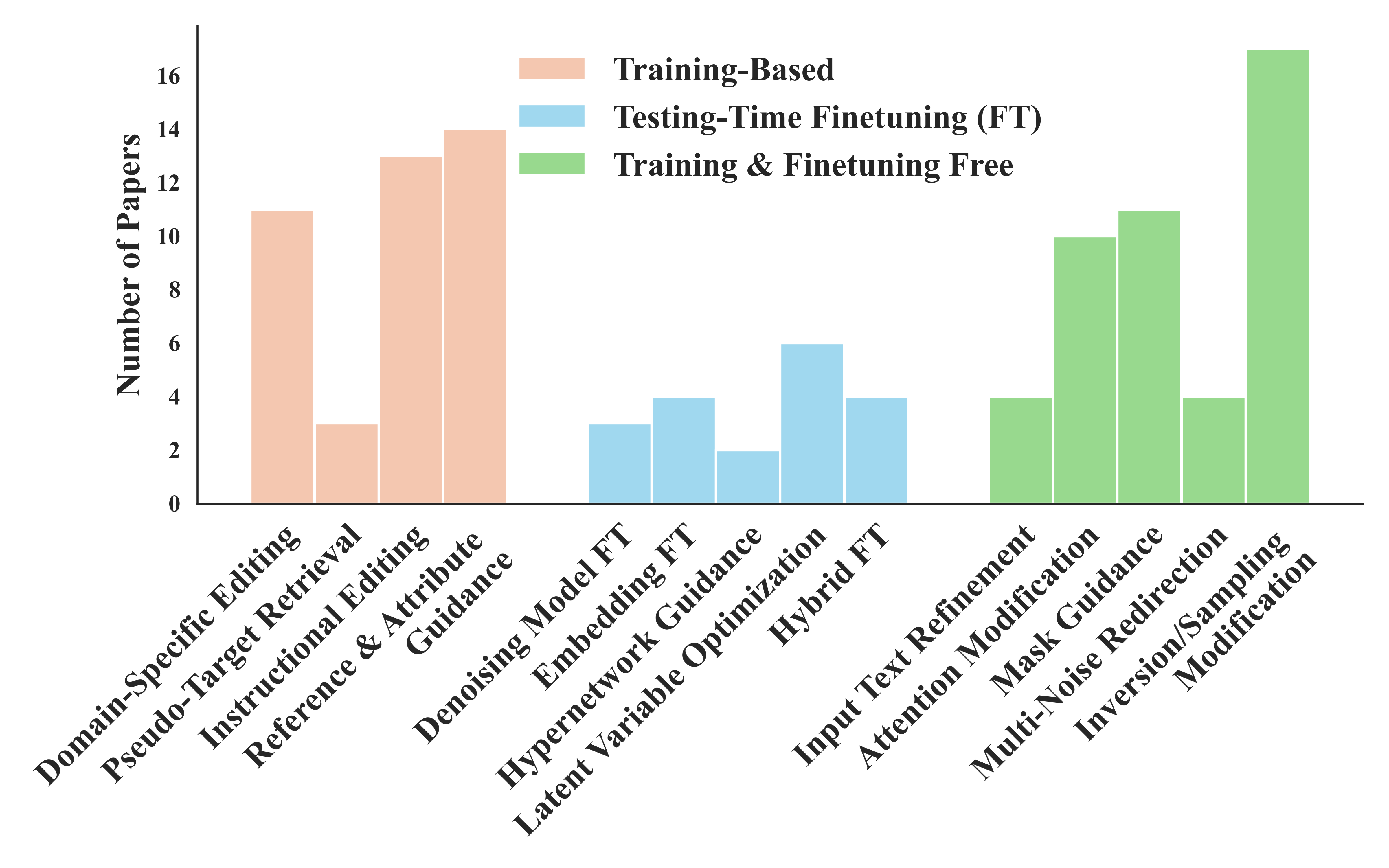}
		\includegraphics[width=0.42\textwidth]{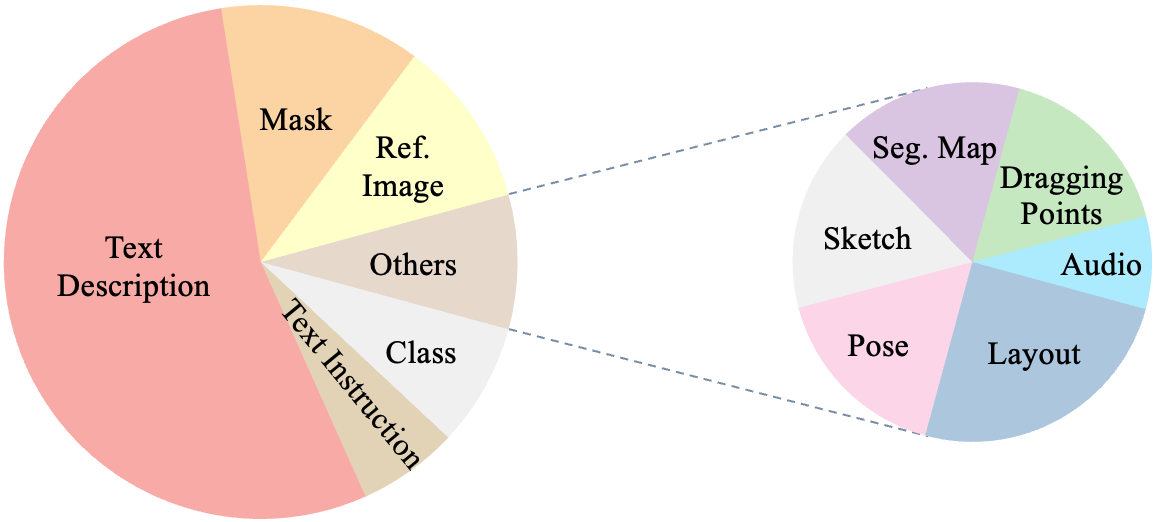}
		\includegraphics[width=0.45\textwidth]{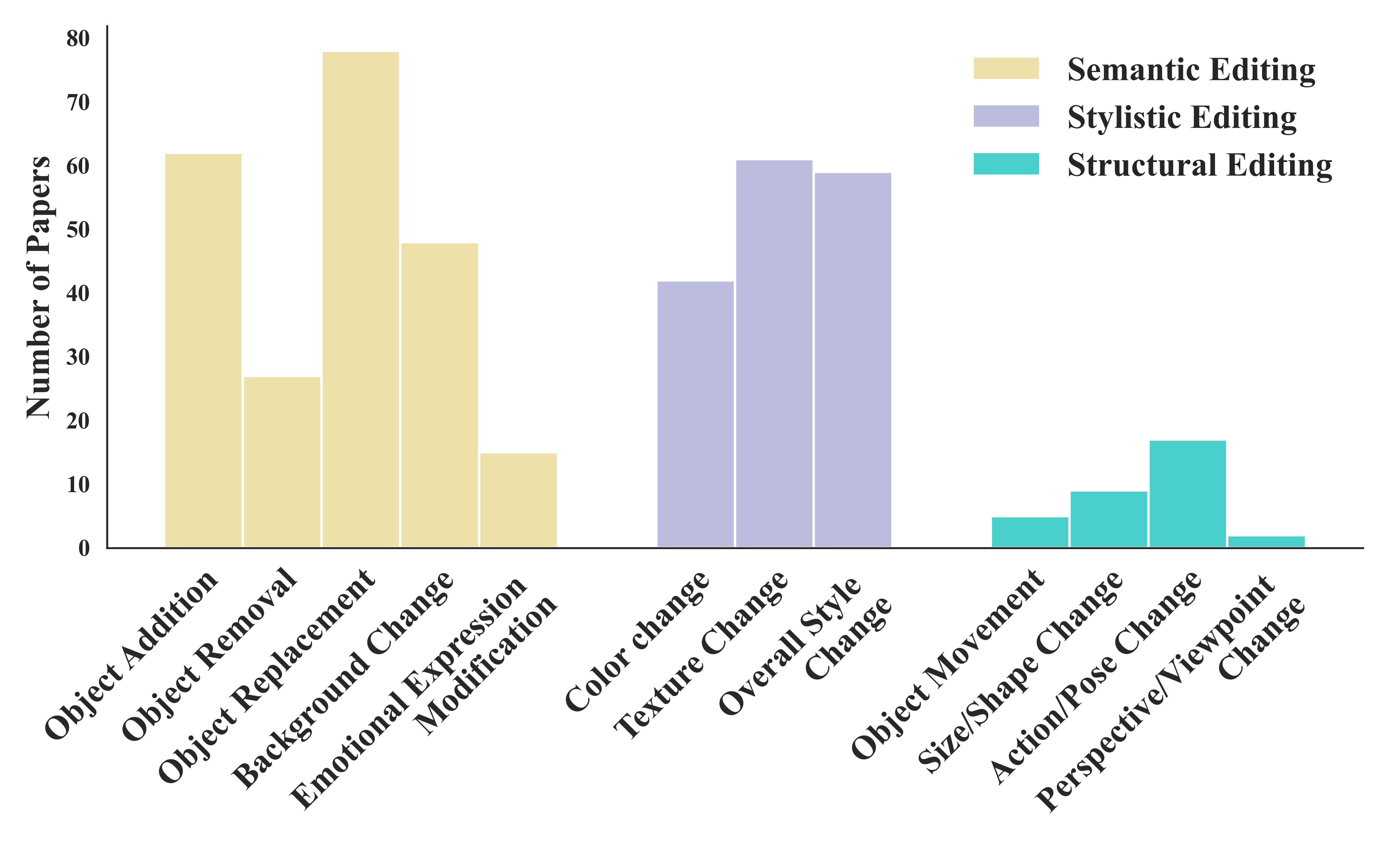}
		\caption{Statistical overview of research publications in diffusion model-based image editing. Top: learning strategies. Middle: input conditions. Bottom: editing tasks.}
		\label{teaser}
			\vspace{-0.8cm}
\end{figure}
    \IEEEraisesectionheading{\section{Introduction}\label{sec:introduction}}

	\IEEEPARstart{I}{n} the realm of AI-generated content (AIGC)~\cite{kingma2013auto, van2016pixel, rezende2015variational, van2017neural, papamakarios2017masked, esser2021taming, ramesh2021zero, zhao2016energy, goodfellow2014generative, sohl2015deep}, image generation and editing are recognized as significant for practical applications, including digital media, advertising, and scientific research. Distinct from image generation, which creates entirely new images from minimal inputs, image editing involves modifying the appearance, structure, or content of an existing image to achieve a desired outcome. This process can encompass a range of changes from subtle adjustments to moderate transformations, all without fundamentally altering the original image. The evolution of image editing has progressed from manual, labor-intensive processes to advanced learning-based algorithms. A pivotal advancement in this evolution is the introduction of Generative Adversarial Networks (GANs) \cite{goodfellow2014generative, karras2019style, mirza2014conditional, isola2017image, zhu2017unpaired, brock2019large}, which significantly enhance the possibilities for creative image manipulation.
 
	More recently, diffusion models have emerged as a powerful tool in this domain. Inspired by principles from non-equilibrium thermodynamics \cite{sohl2015deep}, diffusion models work by gradually adding noise to data and then learning to reverse this process from random noise until generating desired data that matches the source data distribution \cite{ho2020denoising, nichol2021improved, song2021denoising, hyvarinen2005estimation, vincent2011connection, song2019generative, song2021score, song2020improved}. The application of diffusion models in image editing has garnered growing interest, as evidenced by the increasing number of research publications. This interest highlights the potential and versatility of diffusion models in improving image editing performance compared to previous methods. Given this significant advancement, it is essential to systematically review and summarize these contributions. However, despite this progress, there is a notable lack of surveys specifically dedicated to diffusion model-based image editing. Existing surveys \cite{kazerouni2022diffusion, xing2023survey, li2023diffusion, moser2024diffusion} on diffusion models tend to concentrate on other visual tasks, such as image generation \cite{shi2022divae, ramesh2022hierarchical,rombach2022high,saharia2022photorealistic, ho2021classifier, batzolis2021conditional, bao2022analytic, ho2022cascaded, dhariwal2021diffusion, liu2023more,  nichol2022glide, rombach2022text, bansal2022cold, meng2023distillation, phung2023wavelet, xu2023versatile}, restoration and enhancement \cite{saharia2022image, saharia2022palette, ozdenizci2023restoring, shang2023resdiff, gao2023implicit, guo2023shadowdiffusion, luo2023image, xia2023diffir, choi2021ilvr, kawar2022denoising, huang2023wavedm, wang2023zero, yue2022difface, chung2022improving, chung2023diffusion}, or video generation \cite{ho2022video, singer2023make, ho2022imagen, ge2023preserve, blattmann2023align, zhou2022magicvideo, yin2023nuwa, esser2023structure, an2023latent, wang2023modelscope, li2023videogen, he2022latent, wang2023lavie, wu2023tune, lv2023gpt4motion, zhang2023show, blattmann2023stable, girdhar2023emu}. Those that do mention image editing often provide only a cursory overview \cite{yang2023diffusion, cao2022survey, croitoru2023diffusion, ulhaq2022efficient, zhang2023text,  zhang2023survey, po2023state, koo2023comprehensive}, lacking a detailed and focused exploration of the methods.
    To address this gap, we conduct this survey to offer an in-depth and comprehensive analysis specifically focused on 2D image editing using diffusion models. Due to the extensive related literature, works that either focus on other relevant tasks or employ other techniques are not included as the primary review target.
	
	Specifically, we delve deeply into the methodologies, input conditions, and a wide range of editing tasks achieved by diffusion models in this field. The survey critically reviews over 100 research papers, organizing them into three primary classes—\textit{training-based approaches}, \textit{testing-time finetuning approaches}, and \textit{training and finetuning free approaches}—based on whether their learning strategies require training, finetuning during inference, or can operate effectively without either. Each class is further divided based on its core techniques, with detailed discussions presented in Sections~\ref{train},~\ref{finetune}, and~\ref{free}, respectively. We also explore 10 distinct types of input conditions used in these methods, including text, mask, reference (Ref.) image, class, layout, pose, sketch, segmentation (Seg.) map, audio, and dragging points to show the adaptability of diffusion models in diverse image editing scenarios. Furthermore, our survey presents a new classification of image editing tasks into three broad categories, \textit{semantic editing}, \textit{stylistic editing}, and \textit{structural editing}, covering 12 specific types. Fig.~\ref{teaser} visually represents the statistical distributions of research across \textit{learning strategies}, \textit{input conditions}, and \textit{editing task categories}.
	In addition, we pay special attention to inpainting and outpainting, which together stand out as a unique type of editing. We explore both earlier traditional and current multimodal conditional methods, offering a comprehensive analysis of their methodologies in Section~\ref{inpainting}.
    We also introduce EditEval, a benchmark designed to evaluate text-guided image editing algorithms, as detailed in Section~\ref{benchmark}. In particular, an effective evaluation metric, LMM Score, is proposed by leveraging the advanced visual-language understanding capabilities of large multimodal models (LMMs).
Finally, we present some current challenges and potential future trends as an outlook in Section~\ref{future}.
 
	In summary, this survey aims to systematically categorize and critically assess the extensive body of research in diffusion model-based image editing. Our goal is to provide a comprehensive resource that not only synthesizes current findings but also guides future research directions in this rapidly advancing field.

%% file: sec/sec2_background.tex
\section{Background}
	
\subsection{Diffusion Models}
Diffusion Models have exerted a profound influence on the field of generative AI, giving rise to a plethora of approaches falling under their umbrella. Essentially, these models are grounded in a pivotal principle known as diffusion, which gradually adds noise to data samples of some distribution, transforming them into a predefined typically simple distribution, such as Gaussian, and this process is then reversed iteratively to generate data matching the original distribution.
What distinguishes diffusion models from earlier generative models is their dynamic execution across iterative time steps, covering both forward and backward movements in time.

For each time step $t$, the noisy latent $\mathbf{z}_t$ delineates the current state. The time step $t$ increments progressively during forward diffusion process, and it reduces towards 0 during reverse diffusion process. Notably, the literature lacks a distinct differentiation between $\mathbf{z}_t$ in forward diffusion and $\mathbf{z}_t$ in reverse diffusion.
In the context of forward diffusion, let $\mathbf{z}_t \sim q \left( \mathbf{z}_t \mid \mathbf{z}_{t-1} \right)$, and in reverse diffusion, let $\mathbf{z}_{t-1} \sim p \left( \mathbf{z}_{t-1} \mid \mathbf{z}_t \right)$. Herein, we denote $T$ with $0 < t \leq T$ as the maximal time step for finite cases. The initial data distribution at $t=0$ is represented by $\mathbf{z}_0 \sim q\left( \mathbf{z}_0\right)$, slowly contaminated by additive noise to $\mathbf{z}_0$.
Diffusion models gradually eliminate noise by a parameterized model $p_\theta\left( \mathbf{z}_{t-1} \mid \mathbf{z}_t \right)$ in the reverse time direction. This model approximates the ideal, albeit unattainable, denoised distribution $p \left( \mathbf{z}_{t-1} \mid \mathbf{z}_t \right)$.
Denoising Diffusion Probabilistic Models (DDPMs), as introduced in Ho \textit{et al.} \cite{ho2020denoising}, effectively utilize Markov chains to facilitate both the forward and backward processes over a finite series of time steps.

\noindent \textbf{Forward Diffusion Process.} This process serves as transforming the data distribution into a predefined distribution, such as the Gaussian distribution. The transformation is represented as:
\begin{equation}
    q(\mathbf{z}_{t} \mid \mathbf{z}_{t-1}) = \mathcal{N}(\mathbf{z}_{t} \mid \sqrt{1-\beta_t}\, \mathbf{z}_{t-1}, \beta_t \mathbf{I} ),
\end{equation}
where the set of hyper-parameters $0 < \beta_{1:T} < 1$ is indicative of the noise variance introduced at each sequential step. This diffusion process can be briefly expressed via a single-step equation:
\begin{equation}
    q(\mathbf{z}_t \mid \mathbf{z}_0) = \mathcal{N}(\mathbf{z}_t \mid \sqrt{\bar\alpha_t}\, \mathbf{z}_0, (1-\bar\alpha_t) \mathbf{I}),
\end{equation}
with $\alpha_t=1-\beta_t$ and $\bar\alpha_t = \prod_{i=1}^t \alpha_i$, as elaborated in Sohl-Dickstein \textit{et al.} \cite{sohl2015deep}. As a result, bypassing the need to consider intermediate time steps, $\mathbf{z}_t$ can be directly sampled by:
\begin{equation}
    \label{eq_sr3yt}
    \mathbf{z}_t = \sqrt{\bar\alpha_t} \cdot \mathbf{z}_0  + \sqrt{1-\bar\alpha_t} \cdot \epsilon, \quad \epsilon \sim \mathcal{N} \left( \mathbf{0}, \mathbf{I} \right). 
\end{equation}

\noindent \textbf{Reverse Diffusion Process.} The reverse process, also called ancestral sampling~\cite{cao2024survey}, is defined by sampling a random Gaussian noise $\mathbf{z}_T$ and then for $t$ in $T,t-1,...,1$, sampling $\mathbf{z}_{t-1}$, until getting to $\mathbf{z}_{0}$. The primary objective here is to learn the reverse of the forward diffusion process, with the aim of generating a distribution that closely aligns with the original unaltered data samples $\mathbf{z}_0$. In the context of image editing, $\mathbf{z}_0$ represents the edited image. Practically, this is achieved using a UNet architecture to learn a parameterized version of $p$. Given that the forward diffusion process is approximated by $q(\mathbf{z}_{T}) \approx \mathcal{N} \left( \mathbf{0}, \mathbf{I} \right)$, the formulation of the learnable transition is expressed as:
\begin{equation}
    p_\theta \left( \mathbf{z}_{t-1} \mid \mathbf{z}_t \right) = \mathcal{N} \left( \mathbf{z}_{t-1} \mid \mu_{\theta}(\mathbf{z}_{t}, \bar\alpha_t), \Sigma_\theta(\mathbf{z}_{t}, \bar\alpha_t) \right),
\end{equation}
where the functions $\mu_{\theta}$ and $\Sigma_\theta$ are learnable parameters. In addition, for the conditional formulation $p_\theta \left( \mathbf{z}_{t-1} \mid \mathbf{z}_t, \mathbf{c} \right)$, which is conditioned on an external variable $\mathbf{c}$ (in image editing, $\mathbf{c}$ can be the source image), the model becomes $\mu_{\theta}(\mathbf{z}_{t}, \mathbf{c}, \bar\alpha_t)$ and $\Sigma_\theta(\mathbf{z}_{t}, \mathbf{c}, \bar\alpha_t)$.

\noindent \textbf{Optimization.} The optimization strategy for guiding the reverse diffusion in learning the forward process involves minimizing the Kullback-Leibler (KL) divergence between the joint distributions of the forward and reverse sequences. These are mathematically defined as:
\begin{align}
    p_\theta \left( \mathbf{z}_{0}, ..., \mathbf{z}_{T}\right) &= p \left(\mathbf{z}_{T}\right)\prod^T_{t=1}p_\theta \left( \mathbf{z}_{t-1} \mid \mathbf{z}_t \right), \\
    q \left( \mathbf{z}_{0}, ..., \mathbf{z}_{T}\right) &= q \left(\mathbf{z}_{0}\right)\prod^T_{t=1}q \left( \mathbf{z}_{t} \mid \mathbf{z}_{t-1} \right),
\end{align}
leading to the minimization of:
\begin{align}
\label{eq:vlb_ddpm}
    \text{KL} ( q \left( \mathbf{z}_{0}, ..., \mathbf{z}_{T}\right) \| p_\theta \left( \mathbf{z}_{0}, ..., \mathbf{z}_{T}\right)) \\
    \geq \mathbb{E}_{q(\mathbf{z}_{0})}\left[ - \log p_\theta \left( \mathbf{z}_{0}\right)\right] + c, \nonumber
\end{align}
which is detailed in Ho \textit{et al.}~\cite{ho2020denoising} and the constant $c$ is irrelevant for optimizing $\theta$. The KL divergence of Eq.~\ref{eq:vlb_ddpm} represents the variational bound of the log-likelihood of the data ($\log p_\theta( \mathbf{z}_{0})$). This KL divergence serves as the loss and is minimized in DDPMs. Practically, Ho \textit{et al.} \cite{ho2020denoising} adopt a reweighed version of this loss as a simpler denoising loss:
\begin{equation}
    \label{eq:ddpm_pos}
\mathbb{E}_{t\sim\mathcal{U}(1,T), \mathbf{z}_0\sim q(\mathbf{z}_0), \epsilon \sim \mathcal{N} \left( \mathbf{0}, \mathbf{I} \right)} \left[ \lambda(t) \| \epsilon - \epsilon_\theta(\mathbf{z}_t, t)\|^2\right],
\end{equation}
where $\lambda(t) > 0$ denotes a weighting function, $\mathbf{z}_t$ is obtained using Eq. \ref{eq_sr3yt}, and $\epsilon_\theta$ represents a network designed to predict the noise $\epsilon$ based on $\mathbf{z}_t$ and $t$.

\noindent \textbf{DDIM Sampling and Inversion.} When working with a real image $\mathbf{z}_0$, prevalent editing methods~\cite{meng2021sdedit,cao2023masactrl} initially invert this $\mathbf{z}_0$ into a corresponding $\mathbf{z}_T$ utilizing a specific inversion scheme. Subsequently, the sampling begins from this $\mathbf{z}_T$, employing some editing strategy to produce an edited outcome $\tilde{\mathbf{z}}_0$. In an ideal scenario, direct sampling from $\mathbf{z}_T$, without any edits, should yield a $\tilde{\mathbf{z}}_0$ that closely resembles $\mathbf{z}_0$. A significant deviation of $\tilde{\mathbf{z}}_0$ from $\mathbf{z}_0$, termed reconstruction failure, indicates the inability of the edited image to maintain the integrity of unaltered regions in $\mathbf{z}_0$. Therefore, using an inversion method that ensures $\tilde{\mathbf{z}}_0 \approx \mathbf{z}_0$ is crucial.
The DDIM sampling equation~\cite{song2021denoising} is:
\begin{equation}\label{ddim sampling}
\small
    \mathbf{z}_{t-1} =
\sqrt {\bar\alpha_{t-1}} \frac{\mathbf{z}_t - \sqrt{1-\bar\alpha_t}   \epsilon_\theta(\mathbf{z}_t,t) }{\sqrt {\bar\alpha_t}}
+ \sqrt{1-\bar\alpha_{t-1}}   \epsilon_\theta(\mathbf{z}_t,t),
\end{equation}
which is alternatively expressed as:
\begin{equation}\label{ddim invert ideal}
\small
\mathbf{z}_t =   \sqrt{\bar\alpha_t}  \frac{\mathbf{z}_{t-1} - \sqrt{1-\bar\alpha_{t-1}}   \epsilon_\theta(\mathbf{z}_{t},t) }{\sqrt {\bar\alpha_{t-1}}}+ \sqrt{1-\bar\alpha_{t}}   \epsilon_\theta(\mathbf{z}_{t},t).
\end{equation}
Although Eq. \ref{ddim invert ideal} appears to provide an ideal inversion from $\mathbf{z}_{t-1}$ to $\mathbf{z}_t$, the problem arises from the unknown nature of $\mathbf{z}_t$, which is also used as an input for the network $\epsilon_\theta(\mathbf{z}_{t},t)$. To address this, DDIM Inversion\cite{song2021denoising} operates under the assumption that $\mathbf{z}_{t-1} \approx \mathbf{z}_t$ and replaces $\mathbf{z}_t$ on the right-hand side of Eq.~\ref{ddim invert ideal} with $\mathbf{z}_{t-1}$, leading to the following approximation:
\begin{equation}\label{ddim invert}
\small
\begin{aligned}
   \mathbf{z}_t = \sqrt{\bar\alpha_t}  \frac{\mathbf{z}_{t-1} - \sqrt{1-\bar\alpha_{t-1}}   \epsilon_\theta(\mathbf{z}_{t-1},t) }{\sqrt {\bar\alpha_{t-1}}}+\sqrt{1-\bar\alpha_{t}}   \epsilon_\theta(\mathbf{z}_{t-1},t).
\end{aligned}
\end{equation}

\noindent \textbf{Text Condition and Classifier-Free Guidance.} Text-conditional diffusion models are designed to synthesize outcomes from random noise $\mathbf{z}_T$ guided by a text prompt $P$. During inference in the sampling process, the noise estimation network $\epsilon_\theta(\mathbf{z}_t,t,C)$ is utilized to predict the noise $\epsilon$, where $C = \psi(P)$ represents the text embedding. This process methodically removes noise from $\mathbf{z}_t$ across $T$ steps until the final result $\mathbf{z}_0$ is obtained.

In the realm of text-conditional image generation, it is vital to ensure substantial textual influence and control over the generated output. To this end, Ho \textit{et al.} \cite{ho2021classifier} introduce the concept of classifier-free guidance, a technique that amalgamates conditional and unconditional predictions. More specifically, let $\varnothing = \psi(``")$\footnote[1]{The placeholder $\varnothing$ is typically used for negative prompts to prevent certain attributes from manifesting in the generated image.} denote the null text embedding. When combined with a guidance scale $w$, the classifier-free guidance prediction is formalized as:
\begin{equation}\label{cfg}
\epsilon_\theta(\mathbf{z}_t,t,C,\varnothing) = w \epsilon_\theta(\mathbf{z}_t,t,C) + (1-w) \ \epsilon_\theta(\mathbf{z}_t,t,\varnothing).
\end{equation}
In this formulation, $\epsilon_\theta(\mathbf{z}_t,t,C,\varnothing)$ replaces $\epsilon_\theta(\mathbf{z}_t,t)$ in the sampling Eq.~\ref{ddim sampling}. The value of $w$, typically ranging from $[1, 7.5]$ as suggested in \cite{rombach2022high,saharia2022photorealistic}, dictates the extent of textual control. A higher $w$ value correlates with a stronger text-driven influence in the generation process.

\subsection{Attention in Stable Diffusion}

\noindent \textbf{Cross-Attention in Stable Diffusion.}
In Stable Diffusion, cross-attention layers play a crucial role in fusing images and text, allowing T2I models to generate images that are consistent with textual descriptions. A cross-attention layer receives the query, key, and value matrices, i.e., $Q_{cross}$, $K_{cross}$, and $V_{cross}$, from the noisy image and prompt. Specifically, $Q_{cross}$ is derived from the spatial features of the noisy image by a linear layer, while $K_{cross}$ and $V_{cross}$ are projected from the textual embedding of the input prompt via a linear layer.
The cross-attention map is defined as:
\begin{equation}
 M_{cross} = \text{Softmax}\left(\frac{Q_{cross}{K_{cross}}^{\mathsf{T}}}{\sqrt{d_{cross}}}\right),
\label{equation4}
\end{equation}
where $d_{cross}$ is the dimensionality of the keys and queries.
The final output is defined as the fused feature of the text and image, denoted as $M_{cross}V_{cross}$.
Intuitively, each cell $M_{cross}^{ij}$ in the cross-attention map determines the weights attributed to the value of the $j$-th text token relative to the spatial feature $i$ of the image. The cross-attention map enables the diffusion model to align the tokens of the prompt in the image area.

\noindent \textbf{Self-Attention in Stable Diffusion.}
Unlike cross-attention, a self-attention layer receives the key matrix $K_{self}$ and the query matrix $Q_{self}$ from the noisy image through linear layers.
The self-attention map is defined as:
\begin{equation}
 M_{self} = \text{Softmax}\left(\frac{Q_{self}{K_{self}}^{\mathsf{T}}}{\sqrt{d_{self}}}\right),
\label{equation6}
\end{equation}
where $d_{self}$ is the dimensionality of $K_{self}$ and $Q_{self}$. $M_{self}^{ij}$ of $M_{self}$ determines the weights assigned to the relevance of the $i$-th and $j$-th spatial features in the image and can affect the spatial layout and shape details of the generated image. Consequently, the self-attention map can be utilized to preserve the spatial structure characteristics of the original image during the editing process.

\subsection{Related Tasks}

While this survey focuses exclusively on diffusion model-based image editing, it is important to acknowledge related tasks that share similarities with image editing but differ in their objectives and methodologies. These tasks, such as conditional image generation, image restoration, and image composition, involve the creation, improvement, or compositing of images rather than the direct modification of existing content. Although these tasks are not the primary focus of this survey, understanding their distinctions and connections to image editing is crucial for a comprehensive view of how diffusion models are applied across various vision tasks.

\subsubsection{Conditional Image Generation}
    \noindent \textbf{Class-Conditioned Image Generation.} Early efforts \cite{dhariwal2021diffusion, ho2022cascaded, chao2022denoising, karras2022elucidating, lu2022dpm,salimans2022progressive} usually incorporate the class-induced gradients via an additional pretrained classifier during sampling. However, Ho \textit{et al.} \cite{ho2021classifier} introduce the classifier-free guidance, which does not rely on an external classifier and allows for more versatile conditions, e.g., text, as guidance.
	
	\noindent \textbf{Text-to-Image (T2I) Generation.} GLIDE \cite{nichol2022glide} is the first work that uses text to guide image generation directly from the high-dimensional pixel level, replacing the label in class-conditioned diffusion models. Similarly, Imagen \cite{saharia2022photorealistic} uses a cascaded framework to generate high-resolution images more efficiently in pixel space. A different line of research first projects the image into a lower-dimensional space and then applies diffusion models in this latent space. Representative works include Stable Diffusion (SD) \cite{rombach2022high}, VQ-diffusion \cite{gu2022vector}, and DALL-E 2 \cite{ramesh2022hierarchical}. Following these pioneering studies, a large number of works \cite{meng2023distillation, podell2023sdxl, chen2023pixart, dai2023emu, sohn2023styledrop, balaji2022ediffi, ge2023expressive} are proposed, advancing this field over the past two years.
	
	\noindent \textbf{Additional Conditions.} Beyond text, more specific conditions are also used to achieve higher fidelity and more precise control in image synthesis. GLIGEN \cite{li2023gligen} inserts a gated self-attention layer between the original self-attention and cross-attention layers in each block of a pretrained T2I diffusion model for generating images conditioned on grounding boxes.
	Make-A-Scene \cite{gafni2022make} and SpaText \cite{avrahami2023spatext} use segmentation masks to guide the generation process. In addition to segmentation maps, ControlNet \cite{zhang2023adding} can also incorporate other types of input, such as depth maps, normal maps, canny edges, pose, and sketches as conditions. Other methods \cite{zhao2023uni, qin2023unicontrol, huang2023composer, mou2023t2i, xu2024prompt} integrate diverse conditional inputs and include additional layers, enhancing the generative process controlled by these conditions.
	
	\noindent \textbf{Personalized Image Generation.} Closely related to image editing within conditional image generation is the task of creating personalized images. This task focuses on generating images that maintain a certain identity, typically guided by a few reference images of the same subject. Two early approaches are Textual Inversion \cite{gal2023image} and DreamBooth \cite{ruiz2023dreambooth}. Specifically, Textual Inversion learns a unique identifier word to represent a new subject and incorporates this word into the text encoder's dictionary. DreamBooth binds a new rare word with a specific subject by finetuning the entire Imagen \cite{saharia2022photorealistic} model with a few reference images. For efficient training, some methods\cite{gu2023mix, ruiz2024hyperdreambooth, kong2024omg} depend on low-rank adaptation (LoRA) that focuses on adapting only a small subset of the model’s parameters. For example, Mix-of-Show adopts embedding-decomposed LoRA (ED-LoRA) for single-concept learning and gradient fusion for multi-concept fusion. Following these foundational works, subsequent methods \cite{kumari2023multi, voronov2023loss, wei2023elite, liu2023cones, chen2023subject, shi2023instantbooth, tewel2023key, chen2023disenbooth, gu2023photoswap, yuan2023customnet, lu2023specialist, ye2023ip} provide more refined control over the generated images, enhancing the precision and accuracy of the output.
 

\noindent \textbf{Vector Graphics Generation.} Scalable Vector Graphics (SVGs) can be scaled to any size, and are compact for digital icons, graphics, and stickers. VectorFusion~\cite{Jain_2023_CVPR} shows that a text-conditioned diffusion model trained on pixel representations of images can be used to generate SVG-exportable vector graphics. DiffSketcher~\cite{xing2023diffsketcher} creates vectorized sketches by optimizing a set of Bézier curves with the score distillation sampling loss. A recent work~\cite{Zhao_2024_CVPR} proposes a dual-domain (vector-pixel) diffusion with cross-modality impulse signals from each other.

\subsubsection{Image Restoration and Enhancement}


\noindent \textbf{Input Image as a Condition.} Generative models have significantly contributed to diverse image restoration tasks, such as super-resolution (SR) and deblurring \cite{srdiff, batzolis2021conditional, esser2021taming, ramesh2021zero, ozbey2022unsupervised}. Super-Resolution via Repeated Refinement (SR3)~\cite{saharia2022image} utilizes DDPM for conditional image generation through a stochastic, iterative denoising process. Cascaded Diffusion Model~\cite{ho2022cascaded} adopts multiple diffusion models in sequence, each generating images of increasing resolution. SRDiff \cite{srdiff} follows a close realization of SR3. The main distinction between SRDiff and SR3 is that SR3 predicts the target image directly, whereas SRDiff predicts the difference between the input and output images.

\noindent \textbf{Restoration in Non-Spatial Spaces.} Some diffusion model-based IR methods focus in other spaces. For example, Refusion \cite{luo2023refusion, luo2023image} employs a mean-reverting Image Restoration (IR)-SDE to transform target images into their degraded counterparts. 
They leverage an autoencoder to compress the input image into its latent representation, with skip connections for accessing multi-scale details. Chen \textit{et al.} \cite{chen2023hierarchical} employ a similar approach by proposing a two-stage strategy called Hierarchical Integration Diffusion Model. The conversion from the spatial to the wavelet domain is lossless and offers significant advantages. For instance, WaveDM \cite{huang2023wavedm} modifies the low-frequency band, whereas WSGM \cite{guth2022wavelet} or ResDiff \cite{shang2023resdiff} conditions the high-frequency bands relative to the low-resolution image. BDCE~\cite{huang2023bootstrap} designs a bootstrap diffusion model in the depth curve space for high-resolution low-light image enhancement.

\noindent \textbf{T2I Prior Usage.} The incorporation of T2I information proves advantageous as it allows the usage of pretrained T2I models. 
These models can be finetuned by adding specific layers or encoders tailored to the IR task. Wang \textit{et al.} put this concept into practice with StableSR \cite{wang2023exploiting}.
Central to StableSR is a time-aware encoder trained in tandem with a frozen Stable Diffusion model~\cite{rombach2022high}. 
This setup seamlessly integrates trainable spatial feature transform layers, enabling conditioning based on the input image. DiffBIR~\cite{lin2023diffbir} uses pretrained T2I diffusion models for blind image restoration, with a two-stage pipeline and a controllable module. CoSeR~\cite{sun2023coser} introduces Cognitive Super-Resolution, merging image appearance and language understanding. SUPIR~\cite{yu2024scaling} leverages generative priors, model scaling, and a dataset of 20 million images for advanced restoration guided by textual prompts, featuring negative-quality prompts and a restoration-guided sampling method.

\noindent \textbf{Projection-Based Methods.}
These methods aim to extract inherent structures or textures from input images to complement the generated images at each step and to ensure data consistency. ILVR \cite{choi2021ilvr}, which projects the low-frequency information from the input image to the output image, ensures data consistency and establishes an improved condition. To address this and enhance data consistency, some recent works \cite{chung2023diffusion, chung2022improving, song2022pseudoinverse} take a different approach by aiming to estimate the posterior distribution using the Bayes theorem.

\noindent \textbf{Decomposition-Based Methods.}
These methods view IR tasks as a linear reverse problem. Denoising Diffusion Restoration Models (DDRM)~\cite{kawar2022denoising} utilizes a pretrained denoising diffusion generative model to address linear inverse problems, showcasing versatility across super-resolution, deblurring, inpainting, and colorization under varying levels of measurement noise. Denoising Diffusion Null-space Model (DDNM)~\cite{wang2023zero} represents another decomposition-based zero-shot approach applicable to a broad range of linear IR problems beyond image SR, such as colorization, inpainting, and deblurring. 
It leverages the range-null space decomposition methodology \cite{schwab2019deep, wang2023gan} to tackle diverse IR challenges effectively.

\subsubsection{Image Composition}
\noindent \textbf{Image-Based Virtual Try-On.}
    The goal of image-based virtual try-on (VTON) is to generate images of a person wearing a chosen garment while preserving the person’s identity and ensuring visual coherence \cite{islam2024deep, song2023image}. Traditional warp-based VTON methods \cite{han2018viton, wang2018toward, han2019clothflow, ge2021parser, xie2023gp} generally involve a two-step process: first, the garment is adjusted to match the target person's pose, and then it is seamlessly overlaid onto the person's image. Recently, with the rise of diffusion models, several VTON approaches \cite{li2023warpdiffusion, zhu2023tryondiffusion, morelli2023ladi, kim2024stableviton, zhang2024mmtryon, chong2024catvton} have emerged, leading to significant improvements in generating realistic and high-quality try-on images. For example, TryOnDiffusion \cite{zhu2023tryondiffusion} unifies two UNets to preserve garment details and warp the garment for significant pose and body change in a single network. LaDIVTON \cite{morelli2023ladi} maps garment features to the CLIP embeddings to condition the latent diffusion model along with the warped input. StableVITON \cite{kim2024stableviton} employs a specialized zero cross-attention block to establish a semantic connection between the garment and the person’s features.

\input{table/big_table_half_fulltask}

    \noindent \textbf{Image-Guided Object Composition.}
    Image-guided object composition \cite{song2023objectstitch,yang2023paint,lu2023tf,Sarukkai_2024_WACV,song2024imprint} incorporates specific objects and scenarios from user-provided photos, potentially with the assistance of a text prompt. Recent studies \cite{song2023objectstitch,yang2023paint} effectively incorporate additional guiding images into diffusion models through retraining pre-trained models on their datasets. TF-ICON \cite{lu2023tf} involves the gradual injection of composite self-attention maps through multiple samplers. PrimeComposer~\cite{wang2024primecomposer} formulates this task as a subject-guided local editing problem and depicts the object through attention steering.

%% file: table/big_table_half_fulltask.tex
\begin{table*}[!th]\footnotesize
\renewcommand\arraystretch{1.15}
\setlength{\tabcolsep}{2.3mm}
 \caption{Comprehensive categorization of diffusion model-based image editing methods from multiple perspectives. The methods are color-rendered according to \colorbox{Training!\TrainingColor}{training},  \colorbox{Finetuning!\FinetuningColor}{testing-time finetuning}, and \colorbox{Free!\FreeColor}{training \& finetuning free}. Input conditions include text, class, reference (Ref.) image, segmentation (Seg.) map, pose, mask, layout, sketch, dragging points, and audio. Task capabilities—semantic, stylistic, and structural—are marked with checkmarks (\checkmark) according to the experimental results provided in the source papers.} 
 \label{tab_taxonomy}
 {
  \begin{tabular}{c|m{1.9cm}<{\centering}|m{0.5cm}<{\centering}m{0.5cm}<{\centering}m{0.5cm}<{\centering}m{0.5cm}<{\centering}m{0.5cm}<{\centering}|m{0.5cm}<{\centering}m{0.5cm}<{\centering}m{0.5cm}<{\centering}|m{0.5cm}<{\centering}m{0.5cm}<{\centering}m{0.5cm}<{\centering}m{0.5cm}<{\centering}}

   \toprule
   
   \multirow{4}{*}{\makecell[c]{\textbf{Method}}} & \multirow{4}{*}{\makecell[c]{ \textbf{Condition(s)}}}&\multicolumn{5}{c|}{\textbf{Semantic Editing}} & \multicolumn{3}{c|}{\textbf{Stylistic Editing}} & \multicolumn{4}{c}{\textbf{Structural Editing}} \\ \cline{3-14}
   ~& ~&\multirow{3}*{\makecell[c]{Obj. \\ Add.}}&\multirow{3}*{\makecell[c]{Obj. \\ Remo.}}&\multirow{3}*{ \makecell[c]{Obj. \\ Repl.}}&
   \multirow{3}*{\makecell[c]{Bg. \\ Chg.}}&  \multirow{3}*{\makecell[c]{Emo. \\ Expr. \\ Mod.}}& 
   \multirow{3}*{\makecell[c]{Color \\ Chg.}} &\multirow{3}*{ \makecell[c]{Style \\ Chg.}}& \multirow{3}*{\makecell[c]{Text. \\ Chg.}}&
   \multirow{3}*{\makecell[c]{Obj. \\ Move.} }& \multirow{3}*{\makecell[c]{Obj. \\ Size. \\ Chg.}}&\multirow{3}*{ \makecell[c]{Obj. \\ Act. \\ Chg.}}& \multirow{3}*{\makecell[c]{Persp. \\ /View. \\ Chg.}} \\ 
   ~& ~&~& ~&~& ~&~& ~&~& ~&~& ~&~\\
   ~& ~&~& ~&~& ~&~& ~&~& ~&~& ~&~\\
   \bottomrule

\rowcolor{Training!\TrainingColor}
DiffusionCLIP ~\cite{kim2022diffusionclip} & Text, Class & & & ~~\checkmark & & ~~\checkmark & ~~\checkmark & ~~\checkmark & ~~\checkmark & & & & \\
\rowcolor{Training!\TrainingColor}
Asyrp ~\cite{kwon2023diffusion} & Text, Class & & & ~~\checkmark & & ~~\checkmark & & & ~~\checkmark & & & & \\
\rowcolor{Training!\TrainingColor}
EffDiff ~\cite{starodubcev2023towards} & Class & & & ~~\checkmark & & ~~\checkmark & ~~\checkmark & ~~\checkmark & ~~\checkmark & & & & \\
\rowcolor{Training!\TrainingColor}
DiffStyler ~\cite{huang2024diffstyler} & Text, Class & & & & & & & ~~\checkmark & & & & & \\
\rowcolor{Training!\TrainingColor}
StyleDiffusion ~\cite{wang2023stylediffusion} & Ref. Image, Class & & & & & & & ~~\checkmark & & & & & \\
\rowcolor{Training!\TrainingColor}
UNIT-DDPM ~\cite{sasaki2021unit} & Class & & & & & & & ~~\checkmark & ~~\checkmark & & & & \\
\rowcolor{Training!\TrainingColor}
CycleNet ~\cite{xu2023cyclenet} & Class & & & ~~\checkmark & & & & ~~\checkmark & & & & & \\
\rowcolor{Training!\TrainingColor}
Diffusion Autoencoders ~\cite{preechakul2022diffusion} & Class & & & ~~\checkmark & & ~~\checkmark & & & ~~\checkmark & & & & \\
\rowcolor{Training!\TrainingColor}
HDAE ~\cite{lu2024hierarchical} & Class & & & ~~\checkmark & & ~~\checkmark & & & ~~\checkmark & & & & \\
\rowcolor{Training!\TrainingColor}
EGSDE ~\cite{zhao2022egsde} & Class & & & ~~\checkmark & & & & & & & & & \\
\rowcolor{Training!\TrainingColor}
Pixel-Guided Diffusion ~\cite{matsunaga2022fine} & Seg. Map, Class & & & & & & & & ~~\checkmark & & ~~\checkmark & ~~\checkmark & \\
\hline
\rowcolor{Training!\TrainingColor}
PbE ~\cite{yang2023paint} & Ref. Image & ~~\checkmark & & ~~\checkmark & & & & & & & & & \\
\rowcolor{Training!\TrainingColor}
RIC ~\cite{kim2023reference} & Ref. Image, Sketch & ~~\checkmark & & ~~\checkmark & ~~\checkmark & & ~~\checkmark & & & & ~~\checkmark & & \\
\rowcolor{Training!\TrainingColor}
ObjectStitch ~\cite{song2023objectstitch} & Ref. Image & ~~\checkmark & & & & & & & & & & & \\
\rowcolor{Training!\TrainingColor}
PhD ~\cite{zhang2023paste} & Ref. Image, Layout & ~~\checkmark & & ~~\checkmark & & & & ~~\checkmark & & & & & \\
\rowcolor{Training!\TrainingColor}
DreamInpainter ~\cite{xie2023dreaminpainter} & Ref. Image, Mask, Text & ~~\checkmark & ~~\checkmark & & & & & & ~~\checkmark & & & ~~\checkmark & \\
\rowcolor{Training!\TrainingColor}
Anydoor ~\cite{chen2023anydoor} & Ref. Image, Mask & ~~\checkmark & & ~~\checkmark & & & & & & ~~\checkmark & & & \\
\rowcolor{Training!\TrainingColor}
FADING ~\cite{Chen_2023_BMVC} & Text & & & & & & & & ~~\checkmark & & & & \\
\rowcolor{Training!\TrainingColor}
PAIR Diffusion ~\cite{goel2023pair} & Ref. Image, Text & ~~\checkmark & & ~~\checkmark & & & ~~\checkmark & & ~~\checkmark & & & & \\
\rowcolor{Training!\TrainingColor}
SmartBrush ~\cite{xie2023smartbrush} & Text, Mask & ~~\checkmark & & ~~\checkmark & ~~\checkmark & & & & & & & & \\
\rowcolor{Training!\TrainingColor}
IIR-Net ~\cite{zhang2024text} & Text & & & & ~~\checkmark & & ~~\checkmark & ~~\checkmark & & & & & \\
\rowcolor{Training!\TrainingColor}
PowerPaint ~\cite{zhuang2023task} & Text, Mask & ~~\checkmark & ~~\checkmark & ~~\checkmark & & & & & & & & & \\
\rowcolor{Training!\TrainingColor}
Imagen Editor ~\cite{wang2023imagen} & Text, Mask & ~~\checkmark & & ~~\checkmark & & & ~~\checkmark & & ~~\checkmark & & ~~\checkmark & & \\
\rowcolor{Training!\TrainingColor}
SmartMask ~\cite{singh2023smartmask} & Text & ~~\checkmark & & & & & & & & & & & \\
\rowcolor{Training!\TrainingColor}
Uni-paint ~\cite{yang2023uni} & Text, Mask, Ref. Image & ~~\checkmark & ~~\checkmark & ~~\checkmark & & & & & & & & & \\
\hline
\rowcolor{Training!\TrainingColor}
InstructPix2Pix ~\cite{brooks2023instructpix2pix} & Text & ~~\checkmark & & ~~\checkmark & ~~\checkmark & & ~~\checkmark & ~~\checkmark & ~~\checkmark & & & & \\
\rowcolor{Training!\TrainingColor}
MoEController ~\cite{li2023moecontroller} & Text & ~~\checkmark & & ~~\checkmark & ~~\checkmark & & ~~\checkmark & ~~\checkmark & ~~\checkmark & & & & \\
\rowcolor{Training!\TrainingColor}
FoI ~\cite{guo2023focus} & Text & ~~\checkmark & & ~~\checkmark & ~~\checkmark & & ~~\checkmark & ~~\checkmark & & & & & \\
\rowcolor{Training!\TrainingColor}
LOFIE ~\cite{chakrabarty2023learning} & Text & ~~\checkmark & & ~~\checkmark & ~~\checkmark & & & & & & & ~~\checkmark & \\
\rowcolor{Training!\TrainingColor}
InstructDiffusion ~\cite{geng2023instructdiffusion} & Text & ~~\checkmark & ~~\checkmark & ~~\checkmark & & & ~~\checkmark &~~\checkmark &~~\checkmark & & & & \\
\rowcolor{Training!\TrainingColor}
Emu Edit ~\cite{sheynin2023emu} & Text & ~~\checkmark & ~~\checkmark & ~~\checkmark & ~~\checkmark & ~~\checkmark & & & ~~\checkmark & & & & \\
\rowcolor{Training!\TrainingColor}
DialogPaint ~\cite{wei2023dialogpaint} & Text & ~~\checkmark & ~~\checkmark & ~~\checkmark & ~~\checkmark & & ~~\checkmark & & & & & & \\
\rowcolor{Training!\TrainingColor}
Inst-Inpaint ~\cite{yildirim2023inst} & Text & & ~~\checkmark & & & & & & & & & & \\
\rowcolor{Training!\TrainingColor}
HIVE ~\cite{zhang2023hive} & Text & ~~\checkmark & ~~\checkmark &~~\checkmark & & ~~\checkmark & ~~\checkmark & ~~\checkmark & & & & & \\
\rowcolor{Training!\TrainingColor}
ImageBrush ~\cite{yasheng2023imagebrush} & ~~Ref. Image, Seg. Map, Pose & & ~~\checkmark & ~~\checkmark & ~~\checkmark & & & ~~\checkmark & ~~\checkmark & & & & \\
\rowcolor{Training!\TrainingColor}
InstructAny2Pix ~\cite{li2023instructany2pix} & Ref. Image, Audio, Text & ~~\checkmark & ~~\checkmark & & ~~\checkmark & & & ~~\checkmark & & & & & \\
\rowcolor{Training!\TrainingColor}
MGIE ~\cite{fu2023guiding} & Text & & ~~\checkmark & ~~\checkmark & & & & & & & & & \\
\rowcolor{Training!\TrainingColor}
SmartEdit ~\cite{huang2023smartedit} & Text & & ~~\checkmark & ~~\checkmark & & & & & & & & ~~\checkmark & \\
\hline
\rowcolor{Training!\TrainingColor}
iEdit ~\cite{bodur2023iedit} & Text & ~~\checkmark & & ~~\checkmark & & ~~\checkmark & ~~\checkmark & ~~\checkmark & ~~\checkmark & & & ~~\checkmark & \\
\rowcolor{Training!\TrainingColor}
TDIELR ~\cite{lin2023text} & Text & ~~\checkmark & & ~~\checkmark & & & & & ~~\checkmark & & & ~~\checkmark & \\
\rowcolor{Training!\TrainingColor}
ChatFace ~\cite{yue2023chatface} & Text & ~~\checkmark  & &  & & ~~\checkmark & ~~\checkmark & & & & & & \\
\hline
\rowcolor{Finetuning!\FinetuningColor}
UniTune ~\cite{valevski2022unitune} & Text & ~~\checkmark & & ~~\checkmark & ~~\checkmark & & ~~\checkmark & ~~\checkmark & ~~\checkmark & & & & \\
\rowcolor{Finetuning!\FinetuningColor}
Custom-Edit ~\cite{choi2023custom} & Text, Ref. Image & & & ~~\checkmark & & & & ~~\checkmark & & & & & \\
\rowcolor{Finetuning!\FinetuningColor}
KV-Inversion ~\cite{huang2023kv} & Text & & & & & & & & & & & ~~\checkmark & \\ \hline
\rowcolor{Finetuning!\FinetuningColor}
Null-Text Inversion ~\cite{mokady2023null} & Text & & & ~~\checkmark & ~~\checkmark & ~~\checkmark & ~~\checkmark & ~~\checkmark & ~~\checkmark & & & ~~\checkmark & \\
\rowcolor{Finetuning!\FinetuningColor}
DPL ~\cite{wang2023dynamic} & Text & & & ~~\checkmark & & & & & & & & & \\
\rowcolor{Finetuning!\FinetuningColor}
DiffusionDisentanglement ~\cite{Wu_2023_CVPR} & Text & ~~\checkmark & & ~~\checkmark & & ~~\checkmark & ~~\checkmark & ~~\checkmark & ~~\checkmark & & & & \\
\rowcolor{Finetuning!\FinetuningColor}
Prompt Tuning Inversion ~\cite{dong2023prompt} & Text & ~~\checkmark & & ~~\checkmark & & & ~~\checkmark & ~~\checkmark & ~~\checkmark & & & ~~\checkmark & \\
\hline
   \bottomrule
  \end{tabular}
 }
 \newline
 (Continued on the next page)
\end{table*}


\begin{table*}[!th]\footnotesize
{\normalsize {TABLE 1 (continued)}}
\newline
\newline
\renewcommand\arraystretch{1.15}
\setlength{\tabcolsep}{2.3mm}
 {
   \begin{tabular}{m{3.75cm}<{\centering}|m{1.9cm}<{\centering}|m{0.5cm}<{\centering}m{0.5cm}<{\centering}m{0.5cm}<{\centering}m{0.5cm}<{\centering}m{0.5cm}<{\centering}|m{0.5cm}<{\centering}m{0.5cm}<{\centering}m{0.5cm}<{\centering}|m{0.5cm}<{\centering}m{0.5cm}<{\centering}m{0.5cm}<{\centering}m{0.5cm}<{\centering}}

   \toprule
   
   \multirow{4}{*}{\makecell[c]{\textbf{Method}}} & \multirow{4}{*}{\makecell[c]{ \textbf{Condition(s)}}}&\multicolumn{5}{c|}{\textbf{Semantic Editing}} & \multicolumn{3}{c|}{\textbf{Stylistic Editing}} & \multicolumn{4}{c}{\textbf{Structural Editing}} \\ \cline{3-14}
   ~& ~&\multirow{3}*{\makecell[c]{Obj. \\ Add.}}&\multirow{3}*{\makecell[c]{Obj. \\ Remo.}}&\multirow{3}*{ \makecell[c]{Obj. \\ Repl.}}&
   \multirow{3}*{\makecell[c]{Bg. \\ Chg.}}&  \multirow{3}*{\makecell[c]{Emo. \\ Expr. \\ Mod.}}& 
   \multirow{3}*{\makecell[c]{Color \\ Chg.}} &\multirow{3}*{ \makecell[c]{Style \\ Chg.}}& \multirow{3}*{\makecell[c]{Text. \\ Chg.}}&
   \multirow{3}*{\makecell[c]{Obj. \\ Move.} }& \multirow{3}*{\makecell[c]{Obj. \\ Size. \\ Chg.}}&\multirow{3}*{ \makecell[c]{Obj. \\ Act. \\ Chg.}}& \multirow{3}*{\makecell[c]{Persp. \\ /View. \\ Chg.}} \\ 
   ~& ~&~& ~&~& ~&~& ~&~& ~&~& ~&~\\
   ~& ~&~& ~&~& ~&~& ~&~& ~&~& ~&~\\
   \bottomrule
\rowcolor{Finetuning!\FinetuningColor}
StyleDiffusion ~\cite{li2023stylediffusion} & Text & ~~\checkmark & & ~~\checkmark & & ~~\checkmark & & ~~\checkmark & ~~\checkmark & & & & \\
\rowcolor{Finetuning!\FinetuningColor}
InST ~\cite{Zhang_2023_inst} & Text & & & & & & & ~~\checkmark & & & & & \\ \hline
\rowcolor{Finetuning!\FinetuningColor}
DragonDiffusion ~\cite{mou2023dragondiffusion} & Dragging Points & & & ~~\checkmark & & ~~\checkmark & & & ~~\checkmark & ~~\checkmark & ~~\checkmark & ~~\checkmark & \\
\rowcolor{Finetuning!\FinetuningColor}
DragDiffusion ~\cite{shi2023dragdiffusion} & Dragging Points & & & & & ~~\checkmark & & & & & ~~\checkmark & ~~\checkmark & \\
\rowcolor{Finetuning!\FinetuningColor}
DDS ~\cite{hertz2023delta} & Text & ~~\checkmark & & ~~\checkmark & ~~\checkmark & & ~~\checkmark & ~~\checkmark & ~~\checkmark & & & ~~\checkmark & \\
\rowcolor{Finetuning!\FinetuningColor}
DiffuseIT ~\cite{kwon2022diffusion} & Text & & & ~~\checkmark & ~~\checkmark & & ~~\checkmark & ~~\checkmark & ~~\checkmark & & & & \\
\rowcolor{Finetuning!\FinetuningColor}
CDS ~\cite{nam2023contrastive} & Text & ~~\checkmark & & ~~\checkmark & & ~~\checkmark & ~~\checkmark & & ~~\checkmark & & & & \\
\rowcolor{Finetuning!\FreeColor}
MagicRemover ~\cite{yang2023magicremover} & Text & & ~~\checkmark & & & & & ~~\checkmark & & & & & \\
\hline
\rowcolor{Finetuning!\FinetuningColor}
Imagic ~\cite{kawar2023imagic} & Text & ~~\checkmark & & ~~\checkmark & & & ~~\checkmark & ~~\checkmark & & ~~\checkmark & & & \\
\rowcolor{Finetuning!\FinetuningColor}
LayerDiffusion ~\cite{li2023layerdiffusion} & Mask, Text & & & & ~~\checkmark & & & & &~~\checkmark & ~~\checkmark& ~~\checkmark & \\
\rowcolor{Finetuning!\FinetuningColor}
Forgedit ~\cite{zhang2023forgedit} & Text & ~~\checkmark & ~~\checkmark& ~~\checkmark &~~\checkmark &~~\checkmark & ~~\checkmark & ~~\checkmark&~~\checkmark & & & ~~\checkmark& ~~\checkmark \\
\rowcolor{Finetuning!\FinetuningColor}
SINE ~\cite{zhang2023sine} & Text & ~~\checkmark & & & ~~\checkmark & & & ~~\checkmark & ~~\checkmark & & & & \\
\hline
\rowcolor{Free!\FreeColor}
PRedItOR ~\cite{ravi2023preditor} & Text & ~~\checkmark & & ~~\checkmark & ~~\checkmark & & ~~\checkmark & ~~\checkmark & ~~\checkmark & & & & \\
\rowcolor{Free!\FreeColor}
ReDiffuser ~\cite{lin2023regeneration} & Text & ~~\checkmark & ~~\checkmark & & & & & ~~\checkmark & & & ~~\checkmark & & \\
\rowcolor{Free!\FreeColor}
Captioning and Injection ~\cite{kim2023user} & Text & ~~\checkmark & & ~~\checkmark & & & & & & & & & \\
\rowcolor{Free!\FreeColor}
InstructEdit ~\cite{wang2023instructedit} & Text & & & ~~\checkmark & ~~\checkmark & & ~~\checkmark & ~~\checkmark & ~~\checkmark & & & & \\
\hline
\rowcolor{Free!\FreeColor}
Direct Inversion ~\cite{elarabawy2022direct} & Text & ~~\checkmark & & ~~\checkmark & ~~\checkmark & & ~~\checkmark & ~~\checkmark & & & & & \\
\rowcolor{Free!\FreeColor}
DDPM Inversion ~\cite{huberman2023edit} & Text & ~~\checkmark & ~~\checkmark & ~~\checkmark & ~~\checkmark & & ~~\checkmark & ~~\checkmark & ~~\checkmark & & & & \\
\rowcolor{Free!\FreeColor}
SDE-Drag ~\cite{nie2023blessing} & Layout & & & & & & & & & ~~\checkmark & & ~~\checkmark & \\
\rowcolor{Free!\FreeColor}
LEDITS++ ~\cite{brack2023ledits++} & Text & ~~\checkmark & ~~\checkmark & ~~\checkmark & ~~\checkmark & & ~~\checkmark & ~~\checkmark & ~~\checkmark & & & & \\
\rowcolor{Free!\FreeColor}
FEC ~\cite{chen2023fec} & Text & ~~\checkmark & & ~~\checkmark & ~~\checkmark & & & ~~\checkmark & & & & & \\
\rowcolor{Free!\FreeColor}
EMILIE ~\cite{joseph2024iterative} & Text & ~~\checkmark & ~~\checkmark & ~~\checkmark & ~~\checkmark & & & ~~\checkmark & ~~\checkmark & & & & \\
\rowcolor{Free!\FreeColor}
Negative Inversion ~\cite{miyake2023negative} & Text & ~~\checkmark & ~~\checkmark & ~~\checkmark & ~~\checkmark & & ~~\checkmark & ~~\checkmark & ~~\checkmark & & & & \\
\rowcolor{Free!\FreeColor}
ProxEdit ~\cite{han2024proxedit} & Text & ~~\checkmark & ~~\checkmark & ~~\checkmark & ~~\checkmark & & ~~\checkmark & ~~\checkmark & ~~\checkmark & & & & \\
\rowcolor{Free!\FreeColor}
Null-Text Guidance ~\cite{zhao2023null} & Ref. Image, Text & & & & & & & ~~\checkmark & & & & & \\
\rowcolor{Free!\FreeColor}
EDICT ~\cite{wallace2023edict} & Text & ~~\checkmark & ~~\checkmark & ~~\checkmark & ~~\checkmark & & ~~\checkmark & ~~\checkmark & ~~\checkmark & & & & \\
\rowcolor{Free!\FreeColor}
AIDI ~\cite{pan2023effective} & Text & ~~\checkmark & ~~\checkmark & ~~\checkmark & ~~\checkmark & & ~~\checkmark & ~~\checkmark & ~~\checkmark & & & & \\
\rowcolor{Free!\FreeColor}
CycleDiffusion ~\cite{wu2023latent} & Text & ~~\checkmark & & ~~\checkmark & ~~\checkmark & & & ~~\checkmark & ~~\checkmark & & & & \\
\rowcolor{Free!\FreeColor}
InjectFusion ~\cite{jeong2024training} & Ref. Image & & & ~~\checkmark & & & & ~~\checkmark & ~~\checkmark & & & & \\
\rowcolor{Free!\FreeColor}
Fixed-point inversion ~\cite{meiri2023fixed} & Text & & & ~~\checkmark & ~~\checkmark & & & ~~\checkmark & ~~\checkmark & & & & \\
\rowcolor{Free!\FreeColor}
TIC ~\cite{duan2023tuning} & Text, Mask & & & & & & & & & & & ~~\checkmark & \\
\rowcolor{Free!\FreeColor}
Diffusion Brush ~\cite{gholami2023diffusion} & Text, Mask & & & ~~\checkmark & & & & & & & & & \\
\rowcolor{Free!\FreeColor}
Self-guidance ~\cite{epstein2023diffusion} & Layout & & & & & & & & & ~~\checkmark & ~~\checkmark & & \\
\hline
\rowcolor{Free!\FreeColor}
P2P ~\cite{hertz2022prompt} & Text & ~~\checkmark & & ~~\checkmark & ~~\checkmark & & ~~\checkmark & ~~\checkmark & ~~\checkmark & & & & \\
\rowcolor{Free!\FreeColor}
Pix2Pix-Zero ~\cite{parmar2023zero} & Text & ~~\checkmark & & ~~\checkmark & ~~\checkmark & & & ~~\checkmark & ~~\checkmark & & & & \\
\rowcolor{Free!\FreeColor}
MasaCtrl ~\cite{cao2023masactrl} & Text, Pose, Sketch & & & & & & & & & & & ~~\checkmark & \\
\rowcolor{Free!\FreeColor}
PnP ~\cite{tumanyan2023plug} & Text & & & & & & & ~~\checkmark & ~~\checkmark & & & & \\
\rowcolor{Free!\FreeColor}
TF-ICON ~\cite{lu2023tf} & Text, Ref. Image, Mask & ~~\checkmark & & & & & & & & & & & \\
\rowcolor{Free!\FreeColor}
Object-Shape Variations ~\cite{patashnik2023localizing} & Text & & & & & & & & & & ~~\checkmark & & \\
\rowcolor{Free!\FreeColor}
Conditional Score Guidance\cite{lee2023conditional} & Text & & & ~~\checkmark & ~~\checkmark & & ~~\checkmark & ~~\checkmark & ~~\checkmark & & & & \\
\rowcolor{Free!\FreeColor}
EBMs ~\cite{park2023energy} & Text & & & ~~\checkmark & ~~\checkmark & & ~~\checkmark & ~~\checkmark & ~~\checkmark & & & & \\
\rowcolor{Free!\FreeColor}
Shape-Guided Diffusion ~\cite{park2024shape} & Text, Mask & & & & & & & & ~~\checkmark & & & & \\
\rowcolor{Free!\FreeColor}
HD-Painter ~\cite{manukyan2023hd} & Text, Mask & ~~\checkmark & & ~~\checkmark & & & & & & & & & \\
\hline
\rowcolor{Free!\FreeColor}
FISEdit ~\cite{yu2023fisedit} & Text & ~~\checkmark & & ~~\checkmark & ~~\checkmark & & & ~~\checkmark & ~~\checkmark & & & & \\
\rowcolor{Free!\FreeColor}
Blended Latent Diffusion ~\cite{avrahami2023blended} & Text, Mask & ~~\checkmark & & ~~\checkmark & ~~\checkmark & & ~~\checkmark & & ~~\checkmark & & & & \\
\rowcolor{Free!\FreeColor}
PFB-Diff ~\cite{huang2023pfb} & Text, Mask & ~~\checkmark & & ~~\checkmark & ~~\checkmark & & & & ~~\checkmark & & & & \\
\rowcolor{Free!\FreeColor}
DiffEdit ~\cite{couairon2023diffedit} & Text & ~~\checkmark & ~~\checkmark & ~~\checkmark & ~~\checkmark & & & ~~\checkmark & ~~\checkmark & & & & \\
\rowcolor{Free!\FreeColor}
RDM ~\cite{huang2023region} & Text & ~~\checkmark & & ~~\checkmark & ~~\checkmark & & & ~~\checkmark & ~~\checkmark & & & & \\
\rowcolor{Free!\FreeColor}
MFL ~\cite{liu2023text} & Text & & ~~\checkmark & ~~\checkmark & ~~\checkmark & & ~~\checkmark & & ~~\checkmark & & & & \\
\rowcolor{Free!\FreeColor}
Differential Diffusion ~\cite{levin2023differential} & Text, Mask & ~~\checkmark & & ~~\checkmark & ~~\checkmark & & & ~~\checkmark & ~~\checkmark & & & & \\
\rowcolor{Free!\FreeColor}
Watch Your Steps ~\cite{mirzaei2023watch} & Text & ~~\checkmark & & ~~\checkmark & ~~\checkmark & & ~~\checkmark & ~~\checkmark & ~~\checkmark & & & & \\
\rowcolor{Free!\FreeColor}
Blended Diffusion ~\cite{avrahami2022blended} & Text, Mask & ~~\checkmark & ~~\checkmark & ~~\checkmark & ~~\checkmark & & ~~\checkmark & & ~~\checkmark & & & & \\
\rowcolor{Free!\FreeColor}
ZONE ~\cite{li2023zone} & Text, Mask & ~~\checkmark & ~~\checkmark & ~~\checkmark & ~~\checkmark & & & & ~~\checkmark & & & & \\
\rowcolor{Free!\FreeColor}
Inpaint Anything ~\cite{yu2023inpaint} & Text, Mask & & & ~~\checkmark & & & & & & & & & \\
\hline
\rowcolor{Free!\FreeColor}
The Stable Artist ~\cite{brack2022stable} & Text & ~~\checkmark & ~~\checkmark & ~~\checkmark & ~~\checkmark & & ~~\checkmark & ~~\checkmark & ~~\checkmark & & & & \\
\rowcolor{Free!\FreeColor}
SEGA ~\cite{brack2023sega} & Text & ~~\checkmark & ~~\checkmark & ~~\checkmark & ~~\checkmark & & ~~\checkmark & ~~\checkmark & ~~\checkmark & & & & \\
\rowcolor{Free!\FreeColor}
LEDITS ~\cite{tsaban2023ledits} & Text & ~~\checkmark & & ~~\checkmark & ~~\checkmark & & & ~~\checkmark & ~~\checkmark & & & & \\
\rowcolor{Free!\FreeColor}
OIR-Diffusion ~\cite{yang2023object} & Text & & & ~~\checkmark & ~~\checkmark & & ~~\checkmark & ~~\checkmark & ~~\checkmark & & & & \\
\hline
   \bottomrule
  \end{tabular}
 }
\end{table*}

%% file: sec/sec3_overview.tex
\section{Scope and Categorization}\label{overview}
\input{table/taxonomy_of_training}

\noindent \textbf{Scope.}
In this survey, we concentrate on those works that most directly contribute to the field of image editing using diffusion models. To ensure a clear and focused discussion, this survey applies two primary criteria for the inclusion of works: (1) the task must focus on image editing, defined as modifying the appearance, structure, or content of existing images without significant transformations that would constitute new image generation, and (2) the methodology must depend on diffusion models performed on 2D images. Hence, works that utilize traditional image editing techniques or other generative models are excluded. Furthermore, works that rely on 3D knowledge are also not reviewed.


\noindent \textbf{Categorization.}
In this survey, we organize diffusion model-based image editing papers into three principle groups based on their learning strategies: \textit{training-based approaches}, \textit{testing-time finetuning approaches}, and \textit{training and finetuning free approaches}, which are elaborated in Sections~\ref{train},~\ref{finetune}, and~\ref{free}, respectively. These learning strategies are defined by whether the methods require a substantial training phase, involve finetuning during the inference stage, or can perform without either, thus providing a comprehensive understanding of the resources and processes involved in applying these methods.
Additionally, we explore 10 types of input conditions employed by these methods to control the editing process, including text, mask, reference (Ref.) image, class, layout, pose, sketch, segmentation (Seg.) map, audio, and dragging points. Furthermore, we investigate 12 most common editing types that can be accomplished by these methods, which are organized into three broad categories defined as follows.

\begin{itemize}
    \item \textit{Semantic Editing}: This category encompasses alterations to the content and narrative of an image, affecting the depicted scene's story, context, or thematic elements. Tasks within this category include object addition (Obj. Add.), object removal (Obj. Remo.), object replacement (Obj. Repl.), background change (Bg. Chg.), and emotional expression modification (Emo. Expr. Mod.).

    \item \textit{Stylistic Editing}: This category focuses on enhancing or transforming the visual style and aesthetic elements of an image without altering its narrative content. Tasks within this category include color change (Color Chg.), texture change (Text. Chg.), and overall style change (Style Chg.) encompassing both artistic and realistic styles.

    \item \textit{Structural Editing}: This category pertains to changes in the spatial arrangement, positioning, viewpoints, and characteristics of elements within an image, emphasizing the organization and presentation of objects within the scene. Tasks within this category include object movement (Obj. Move.), object size and shape change (Obj. Size. Chg.), object action and pose change (Obj. Act. Chg.), and perspective/viewpoint change (Persp./View. Chg.).
\end{itemize}

Table~\ref{tab_taxonomy} summarizes the multi-perspective categorization of the surveyed papers comprehensively, providing a quick search.

%% file: table/taxonomy_of_training.tex
\newcommand{\taxsize}{\scriptsize}

\tikzstyle{my-box}=[
    rectangle,
    draw=hidden-draw,
    rounded corners,
    text opacity=1,
    minimum height=1.5em,
    minimum width=5em,
    inner sep=2pt,
    align=center,
    fill opacity=.5,
    line width=0.8pt,
]
\tikzstyle{leaf}=[my-box, minimum height=1.5em,
    fill=hidden-pink!80, text=black, align=left,font=\taxsize,
    inner xsep=2pt,
    inner ysep=4pt,
    line width=0.8pt,
]
\begin{figure*}[t!]
    \centering
    \scalebox{1}{
        \begin{forest}
            forked edges,
            for tree={
                grow=east,
                reversed=true,
                anchor=base west,
                parent anchor=east,
                child anchor=west,
                base=left,
                font=\taxsize,
                rectangle,
                draw=hidden-draw,
                rounded corners,
                align=left,
                minimum width=2em,
                edge+={darkgray, line width=1pt},
                s sep=3pt,
                inner xsep=2pt,
                inner ysep=3pt,
                line width=0.8pt,
                ver/.style={rotate=90, child anchor=north, parent anchor=south, anchor=center},
                transform shape, 
            },
            where level=1{text width=4.5em,font=\taxsize,}{},
            where level=2{text width=3.1em,font=\taxsize,}{},
            where level=3{text width=3.0em,font=\taxsize,}{},
            [
                Training-Based \\ Approaches (Section~\ref{train} )
                [
                    Domain-Specific Editing, text width=7.5em
                    [
                        DiffusionCLIP~\cite{kim2022diffusionclip}{,} Asyrp~\cite{kwon2023diffusion}{,} EffDiff~\cite{starodubcev2023towards}{,} DiffStyler~\cite{huang2024diffstyler}{,} StyleDiffusion~\cite{wang2023stylediffusion}{,} UNIT-DDPM~\cite{sasaki2021unit}{,} \\CycleNet~\cite{xu2023cyclenet}{,} Diffusion Autoencoders~\cite{preechakul2022diffusion}{,} HDAE~\cite{lu2024hierarchical}{,} EGSDE~\cite{zhao2022egsde}{,} Pixel-Guided Diffusion~\cite{matsunaga2022fine}
                        , leaf, text width=33em
                    ]
                ]
                [
                    Reference and Attribute\\ Guided Editing, text width=7.5em
                    [
                        PbE~\cite{yang2023paint}{,} RIC~\cite{kim2023reference}{,} ObjectStitch~\cite{song2023objectstitch}{,} PhD~\cite{zhang2023paste}{,} DreamInpainter~\cite{xie2023dreaminpainter}{,} Anydoor~\cite{chen2023anydoor}{,} FADING~\cite{Chen_2023_BMVC}{,} \\PAIR Diffusion~\cite{goel2023pair}{,} SmartBrush~\cite{xie2023smartbrush}{,} IIR-Net~\cite{zhang2024text}{,} PowerPaint~\cite{zhuang2023task}{,} Imagen Editor~\cite{wang2023imagen}{,} \\SmartMask~\cite{singh2023smartmask}{,}Uni-paint~\cite{yang2023uni}
                        , leaf, text width=33em
                    ]
                ]
                [
                    Instructional Editing, text width=6.5em
                    [
                        InstructPix2Pix~\cite{brooks2023instructpix2pix}{,} MoEController~\cite{li2023moecontroller}{,} FoI~\cite{guo2023focus}{,} LOFIE~\cite{chakrabarty2023learning}{,} InstructDiffusion~\cite{geng2023instructdiffusion}{,}  Emu Edit~\cite{sheynin2023emu}{,} \\DialogPaint~\cite{wei2023dialogpaint}{,} Inst-Inpaint~\cite{yildirim2023inst}{,} HIVE~\cite{zhang2023hive}{,} ImageBrush~\cite{yasheng2023imagebrush}{,} InstructAny2Pix~\cite{li2023instructany2pix}{,} MGIE~\cite{fu2023guiding}{,} \\SmartEdit~\cite{huang2023smartedit}
                        , leaf, text width=34em
                    ]
                ]
                [
                    Pseudo-Target Retrieval Based Editing, text width=12em
                    [
                        iEdit~\cite{bodur2023iedit}{,} TDIELR~\cite{lin2023text}{,} ChatFace~\cite{yue2023chatface}
                        , leaf, text width=14em
                    ]
                ]
            ]
        \end{forest}
    }
     \vspace{-5pt}
    \caption{Taxonomy of training-based approaches for image editing.}
    \vspace{-10pt}
    \label{taxonomy_of_training}
\end{figure*}
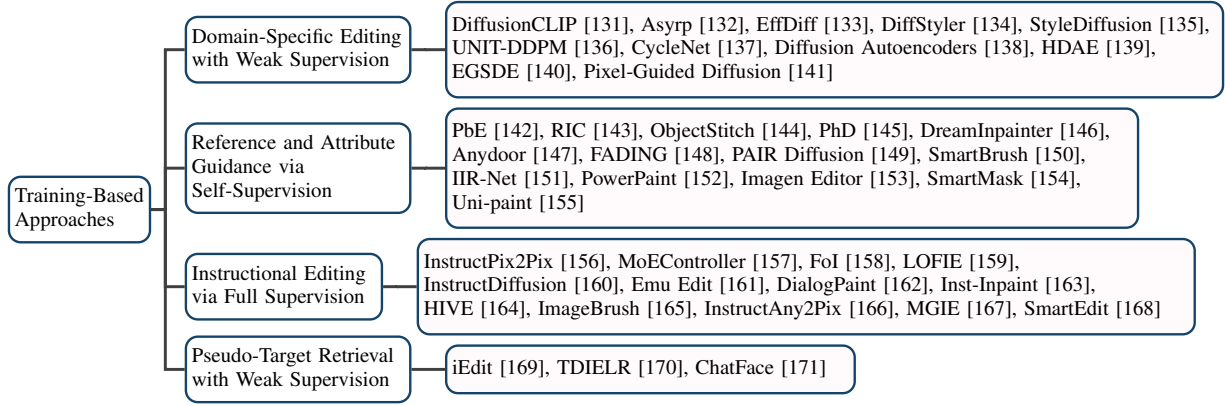

%% file: sec/sec4_train.tex
\section{Training-Based Approaches}\label{train}
In the field of diffusion model-based image editing, training-based approaches refer to methods that require a substantial training phase on large datasets before they can be applied to specific editing tasks. These methods are notable for their stable training processes and effective modeling of data distribution, leading to reliable performance across a variety of editing scenarios. This section categorizes these methods into four main groups based on their application scopes, and the conditions required for training, as shown in Fig.~\ref{taxonomy_of_training}. Further, within each of these primary groups, we classify the approaches into distinct types based on their core editing methodologies. This classification illustrates the range of these methods, from targeted domain-specific applications to broader open-world uses.

\subsection{Domain-Specific Editing}
Over the past several years, Generative Adversarial Networks (GANs) have been widely explored in image editing for their ability to generate high-quality images. However, diffusion models, with their advanced image generation capabilities, emerge as a new focus in this field. One challenge with diffusion models is their need for extensive computational resources when trained on large datasets. To address this, earlier studies train these models on smaller specialized datasets, which highly concentrate on specific domains such as CelebA \cite{karras2018progressive} and FFHQ \cite{karras2019style} for human face manipulation, AFHQ \cite{choi2020stargan} for animal face editing and translation, LSUN \cite{yu2015lsun} for object modification, and WikiArt \cite{WikiArt} for style transfer. To thoroughly understand these approaches, we organize them according to their types of supervision.


\noindent \textbf{CLIP Guidance.}
Drawing inspiration from GAN-based methods \cite{gal2022stylegan, patashnik2021styleclip} that use CLIP \cite{radford2021learning} for guiding image editing with text, several studies incorporate CLIP into diffusion models. 
A key example is DiffusionCLIP \cite{kim2022diffusionclip}, which allows for image manipulation in both trained and new domains using CLIP. Specifically, it first converts a real image into latent noise using DDIM inversion, and then finetunes the pretrained diffusion model during the reverse diffusion process to adjust the image's attributes constrained by a CLIP loss between the source and target text prompts. 
Instead of finetuning on the whole diffusion model, Asyrp \cite{kwon2023diffusion} focuses on a semantic latent space internally, termed \textit{h-space}, where it defines an additional implicit function parameterized by a small neural network. Then it trains the network with the guidance of the CLIP loss while keeping the diffusion model frozen.
To address the time-consuming problem of multi-step optimization in DiffusionCLIP, EffDiff \cite{starodubcev2023towards} introduces a faster method with single-step training and efficient processing. 

Beyond face editing the above methods primarily focus on, DiffStyler \cite{huang2024diffstyler} and StyleDiffusion \cite{wang2023stylediffusion} target artistic style transfer. DiffStyler uses a CLIP instruction loss for the alignment of target text descriptions and generated images, along with a CLIP aesthetic loss for visual quality enhancement. StyleDiffusion, on the other hand, introduces a CLIP-based style disentanglement loss to improve the style-content harmonization.

\noindent \textbf{Cycling Regularization.}
Since diffusion models are capable of domain translation, the cycling framework used in methods like CycleGAN \cite{zhu2017unpaired} is also explored within them.
For example, UNIT-DDPM \cite{sasaki2021unit} defines a dual-domain Markov chain within diffusion models, using cycle consistency to regularize training for unpaired image-to-image translation.
Similarly, CycleNet \cite{xu2023cyclenet} adopts ControlNet \cite{zhang2023adding} with pretrained Stable Diffusion \cite{rombach2022high} as the backbone for text conditioning. It employs consistency regularization throughout the image translation cycle, encompassing both forward translation from the source domain to the target domain and backward translation in the reverse direction.

\noindent \textbf{Projection and Interpolation.}
Another technique frequently used in GANs \cite{shen2020interpreting, abdal2020image2stylegan++} involves projecting two real images into the GAN latent space and then interpolating between them for smooth image manipulation, which is also adopted in some diffusion models for image editing.
For instance, Diffusion Autoencoders \cite{preechakul2022diffusion} introduces a semantic encoder to map an input image to a semantically meaningful embedding, which then serves as a condition of a diffusion model for reconstruction. After training the semantic encoder and the conditional diffusion model, any image can be projected into this semantic space for interpolation. However, HDAE \cite{lu2024hierarchical} points out that this approach tends to miss rich low-level and mid-level features. It addresses this by enhancing the framework to hierarchically exploit the coarse-to-fine features of both the semantic encoder and the diffusion-based decoder, aiming for more comprehensive representations.

\noindent \textbf{Classifier Guidance.}
Some studies enhance image editing performance by introducing an additional pretrained classifier for guidance. EGSDE \cite{zhao2022egsde}, for example, uses an energy function to guide the sampling for realistic unpaired image-to-image translation. This function consists of two log potential functions specified by a time-dependent domain-specific classifier and a low-pass filter, respectively. For fine-grained image editing, Pixel-Guided Diffusion \cite{matsunaga2022fine} trains a pixel-level classifier to both estimate segmentation maps and guide the sampling with its gradient.

\subsection{Reference and Attribute Guided Editing}
This category of works extracts attributes or other information from single images to serve as conditions for training diffusion-based image editing models in a self-supervised manner. They can be classified into two types: reference-based image composition and attribute-controlled image editing.

\noindent \textbf{Reference-Based Image Composition.}
To learn how to composite images, PbE \cite{yang2023paint} is trained in a self-supervised manner using the content inside the object's bounding box of an image as the reference image, and the content outside this bounding box as the source image. To prevent the trivial copy-and-paste solution, it applies strong augmentations to the reference image, creates an arbitrarily shaped mask based on the bounding box, and employs the CLIP image encoder to compress information of the reference image as the condition of the diffusion model. 

Building upon this, RIC \cite{kim2023reference} incorporates sketches of the masked areas as control conditions for training, allowing users to finely tune the effects of reference image synthesis through sketches. ObjectStitch \cite{song2023objectstitch} designs a Content Adaptor to better preserve the key identity information of the reference image. Meanwhile, PhD \cite{zhang2023paste} trains an Inpainting and Harmonizing module on a frozen pretrained diffusion model for efficiently guiding the inpainting of masked areas. To preserve the low-level details of the reference image for inpainting, DreamInpainter \cite{xie2023dreaminpainter} utilizes the downsampling network of U-Net to extract its features. During the training process, it adds noise to the entire image, requiring the diffusion model to learn how to restore the clear image under the guidance of a detailed text description. Furthermore, Anydoor \cite{chen2023anydoor} uses image pairs from video frames as training samples to enhance image composition quality, and introduces modules for capturing identity features, preserving textures, and learning appearance changes.

\noindent \textbf{Attribute-Controlled Image Editing.} This type of papers typically involves augmenting pretrained diffusion models with specific image features as control conditions to learn the generation of corresponding images. This approach allows for image editing by altering these specific control conditions. After training on age-text-face pairs, FADING \cite{Chen_2023_BMVC} edits facial images through null-text inversion and attention control for age manipulation. PAIR Diffusion \cite{goel2023pair} perceives images as a collection of objects, learning to modulate each object's properties, specifically structure and appearance. SmartBrush \cite{xie2023smartbrush} employs masks of varying granularities as control conditions, enabling the diffusion model to inpaint masked regions according to the text and the shape of the mask. To better preserve image information that is irrelevant to the editing text, IIR-Net \cite{zhang2024text} performs color and texture erasure in the required areas. The resulting image, post-erasure, is then used as one of the control conditions for the diffusion model.

\subsection{Instructional Editing}
\begin{figure}[t]
	\small
	\centering
\includegraphics[width=0.48\textwidth]{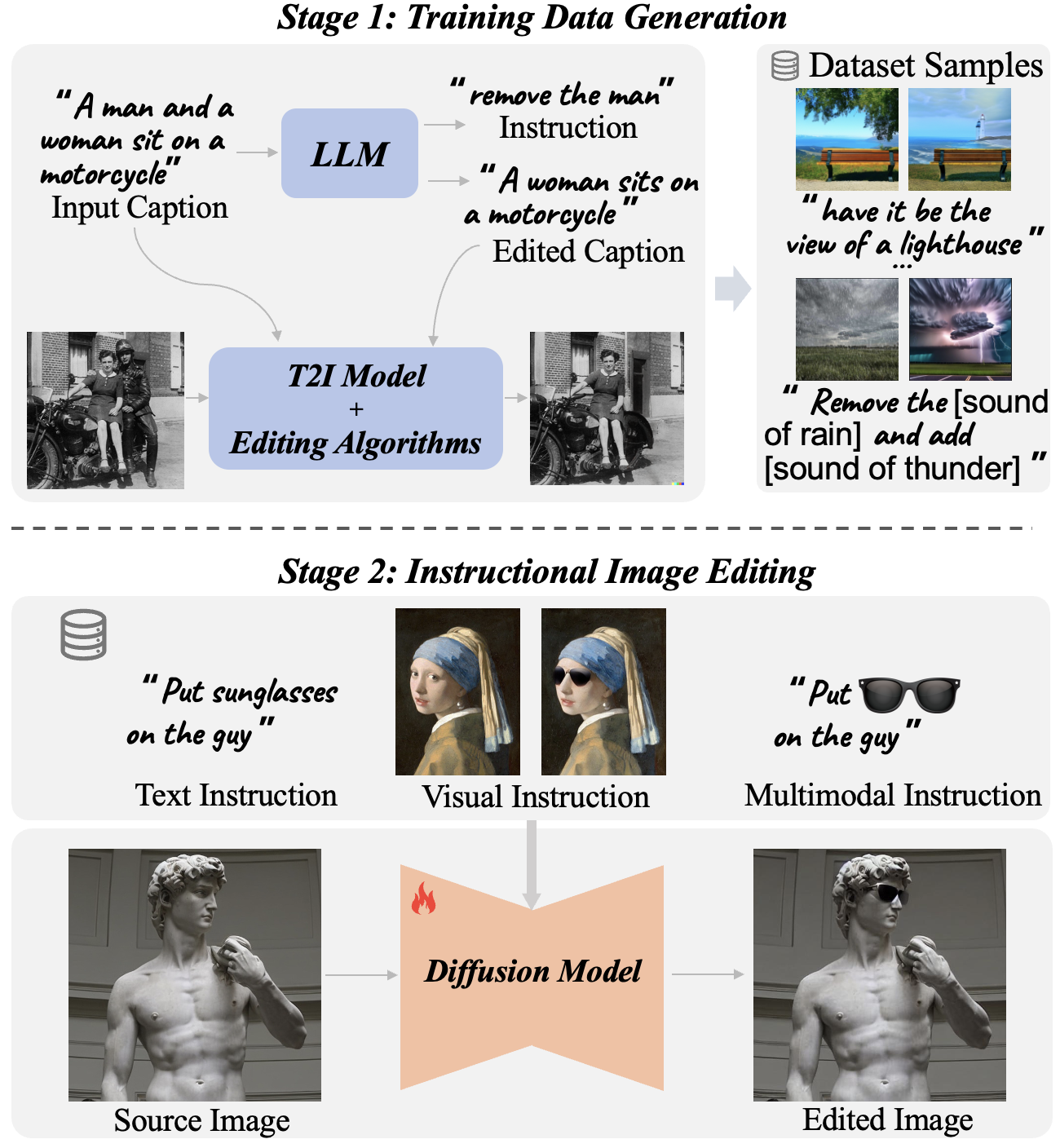}
\vspace{-5pt}
	\caption{Common framework of instructional image editing methods. The sample images are from InstructPix2Pix \cite{brooks2023instructpix2pix}, InstructAny2Pix \cite{li2023instructany2pix} and MagicBrush \cite{zhang2023magicbrush}.}
	\label{fig_instruct}
 \vspace{-10pt}
\end{figure}

Using instructions (e.g., ``Remove the hat") to drive the image editing process as opposed to using a description of the edited image (e.g., ``A puppy with \sout{a hat} and a smile") seems more natural, humanized, and more accurate to the user's needs. InstructPix2Pix \cite{brooks2023instructpix2pix} is the first study that learns to edit images following human instructions. Subsequent works have improved upon it in terms of model architecture, dataset quality, multi-modality, and so on. Therefore, we first describe InstructPix2Pix, and then categorize and present the subsequent works based on their most prominent contributions. Correspondingly, a common framework of these instruction-based methods is depicted in Fig.~\ref{fig_instruct}.

\noindent \textbf{InstructPix2Pix Framework.} One of the major challenges in enabling diffusion models to edit images according to instructions is the construction of the instruction-image paired dataset. InstructPix2Pix generates these image pairs in two steps. First, given an image caption (e.g., ``photograph of a girl riding a horse"), it uses a finetuned GPT-3 \cite{brown2020language} to generate both an instruction (e.g., ``have her ride a dragon.") and an edited image caption (e.g., ``photograph of a girl riding a dragon"). Second, it employs Stable Diffusion and the Prompt-to-Prompt algorithm \cite{hertz2022prompt} to generate edited images, collecting over 450,000 training image pairs. Then it trains an instructional image editing diffusion model in a fully supervised manner, taking into account the conditions of the input image and the instruction.

\noindent \textbf{Model Architecture Enhancement.} MoEController \cite{li2023moecontroller} introduces a Mixture-of-Expert (MOE) architecture, which includes three specialized experts for fine-grained local translation, global style transfer, and complex local editing tasks. On the other hand, FoI \cite{guo2023focus} harnesses the implicit grounding capabilities of InstructPix2Pix to identify and focus on specific editing regions. It also employs cross-condition attention modulation to ensure each instruction targets its corresponding area, reducing interference between multiple instructions. 

\noindent \textbf{Data Quality Enhancement.} 
LOFIE \cite{chakrabarty2023learning} enhances the quality of training datasets by leveraging recent advances in segmentation \cite{kirillov2023segment}, Chain-of-Thought prompting \cite{wei2022chain}, and visual question answering (VQA) \cite{li2023blip}. 
MagicBrush \cite{zhang2023magicbrush} hires crowd workers from Amazon Mechanical Turk (AMT) to manually perform continuous edits using DALL-E 2 \cite{ramesh2022hierarchical}. It comprises 5,313 edit sessions and 10,388 edit turns, thereby establishing a comprehensive benchmark for instructional image editing.
InstructDiffusion \cite{geng2023instructdiffusion} is a unified framework that treats a diverse range of vision tasks as human-intuitive image-manipulating processes, i.e., keypoint detection, segmentation, image enhancement, and image editing. For image editing, it not only utilizes existing datasets but also augmentes them with additional data generated using tools for object removal and replacement, and by collecting image editing pairs from real-world Photoshop requests on the internet. 
Emu Edit \cite{sheynin2023emu} is also trained on both image editing and recognition data, leveraging a dataset comprising 16 distinct tasks with 10 million examples. The dataset is created using Llama 2 \cite{touvron2023llama} and in-context learning to generate diverse and creative editing instructions. During training, the model learns task embeddings in conjunction with its weights, enabling efficient adaptation to new tasks with only a few examples.

DialogPaint \cite{wei2023dialogpaint} aims to extract user's editing intentions during a multi-turn dialogue process and edit images accordingly. It employs the self-instruct technique \cite{wang2022self} on GPT-3 to create a multi-turn dialogue dataset and, in conjunction with four image editing models, generates an instructional image editing dataset. Additionally, the authors finetune the Blender dialogue model \cite{roller2021recipes} to generate corresponding editing instructions based on the dialogue data, which then drives the trained instruction editing model to edit images.

Inst-Inpaint \cite{yildirim2023inst} allows users to specify objects to be removed from an image simply through text commands, without the need for binary masks. It constructs a dataset named GQA-Inpaint, based on the image and scene graph dataset GQA \cite{hudson2019gqa}. It first selects objects and their corresponding relationships from the scene graph, and then uses Detectron2 \cite{wu2019detectron2} and Detic \cite{zhou2022detecting} to extract the segmentation masks of these objects. Afterward, it employs CRFill \cite{zeng2021cr} to generate inpainted target images. The editing instructions are generated through fixed templates. After constructing the dataset, Inst-Inpaint is trained to perform instructional image inpainting.

\noindent \textbf{Human Feedback-Enhanced Learning.} To improve the alignment between edited images and human instructions, HIVE \cite{zhang2023hive} introduces Reinforcement Learning from Human Feedback (RLHF) in instructional image editing. After obtaining a base model following \cite{brooks2023instructpix2pix}, a reward model is then trained on a human ranking dataset. The reward model's estimation is integrated into the training process to finetune the diffusion model, aligning it with human feedback.

\noindent \textbf{Visual Instruction.} ImageBrush \cite{yasheng2023imagebrush} is proposed to learn visual instruction from a pair of transformation images that illustrate the desired manipulation, and to apply this instruction to edit a new image. The method concatenates example images, the source image, and a blank image into a grid, using a diffusion model to iteratively denoise the blank image based on the contextual information provided by the example images. Additionally, a visual prompting encoder is proposed to extract features from the visual instructions to enhance the diffusion process.

\noindent \textbf{Leveraging Multimodal Large-Scale Models.} InstructAny2Pix \cite{li2023instructany2pix} enables users to edit images through instructions that integrate audio, image, and text. It employs ImageBind \cite{girdhar2023imagebind}, a multimodal encoder, to convert diverse inputs into a unified latent space representation. Vicuna-7b \cite{chiang2023vicuna}, a large language model (LLM), encodes the multimodal input sequence to predict two special tokens as the conditions to align the multimodal inputs with the editing results of the diffusion model. 

MGIE \cite{fu2023guiding} inputs the image and instruction, along with multiple [IMG] tokens, into the Multimodal Large Language Model (MLLM), LLaVA \cite{liu2024visual}. It then projects the hidden state of [IMG] in the penultimate layer into the cross attention layer of the UNet in Stable Diffusion. During the training process, the weights of LLaVA and Stable Diffusion are jointly optimized. 

Similarly, SmartEdit \cite{huang2023smartedit} employs LLaVA and additionally introduces a Bidirectional Interaction Module (BIM) to perform image editing in complex scenarios. It first aligns the MLLM's hidden states with the CLIP text encoder using QFormer \cite{li2023blip}, and then it facilitates the fusion between the image feature and the QFormer output through BIM. Besides, it leverages perception-related data, such as segmentation data, to strengthen the model's understanding of spatial and conceptual attributes. Additionally, it incorporates synthetic editing data in complex understanding and reasoning scenarios to activate the MLLM's reasoning ability.

\subsection{Pseudo-Target Retrieval Based Editing}
Since obtaining edited images that accurately represent the ground truth is challenging, the methods in this category aim to retrieve pseudo-target images or directly use CLIP scores \cite{radford2021learning} as the objective to optimize model parameters. iEdit \cite{bodur2023iedit} trains the diffusion model with weak supervision, utilizing CLIP to retrieve and edit the dataset images that are most similar to the editing text, serving as pseudo-target image post-editing. Additionally, it incorporates masks into the image editing process to enable localized preservation by CLIPSeg \cite{luddecke2022image}. To effectively tackle region-based image editing, TDIELR \cite{lin2023text} initially processes the input image using DINO \cite{zhang2023dino} to generate attention maps and features for anchor initialization. It learns a region generation network (RGN) to select the most fitting region proposals. The chosen regions and text descriptions are then fed into a pretrained text-to-image model for editing. TDIELR employs CLIP to calculate scores, assessing the similarity between the text descriptions and editing results, providing a training signal for RGN. Furthermore, ChatFace \cite{yue2023chatface} also utilizes CLIP scores as a metric to learn how to edit on real facial images.

%% file: sec/sec5_finetune.tex
\section{Testing-Time Finetuning Approaches}\label{finetune}
\begin{figure}[!t]
		\small
		\centering
		\includegraphics[width=0.48\textwidth]{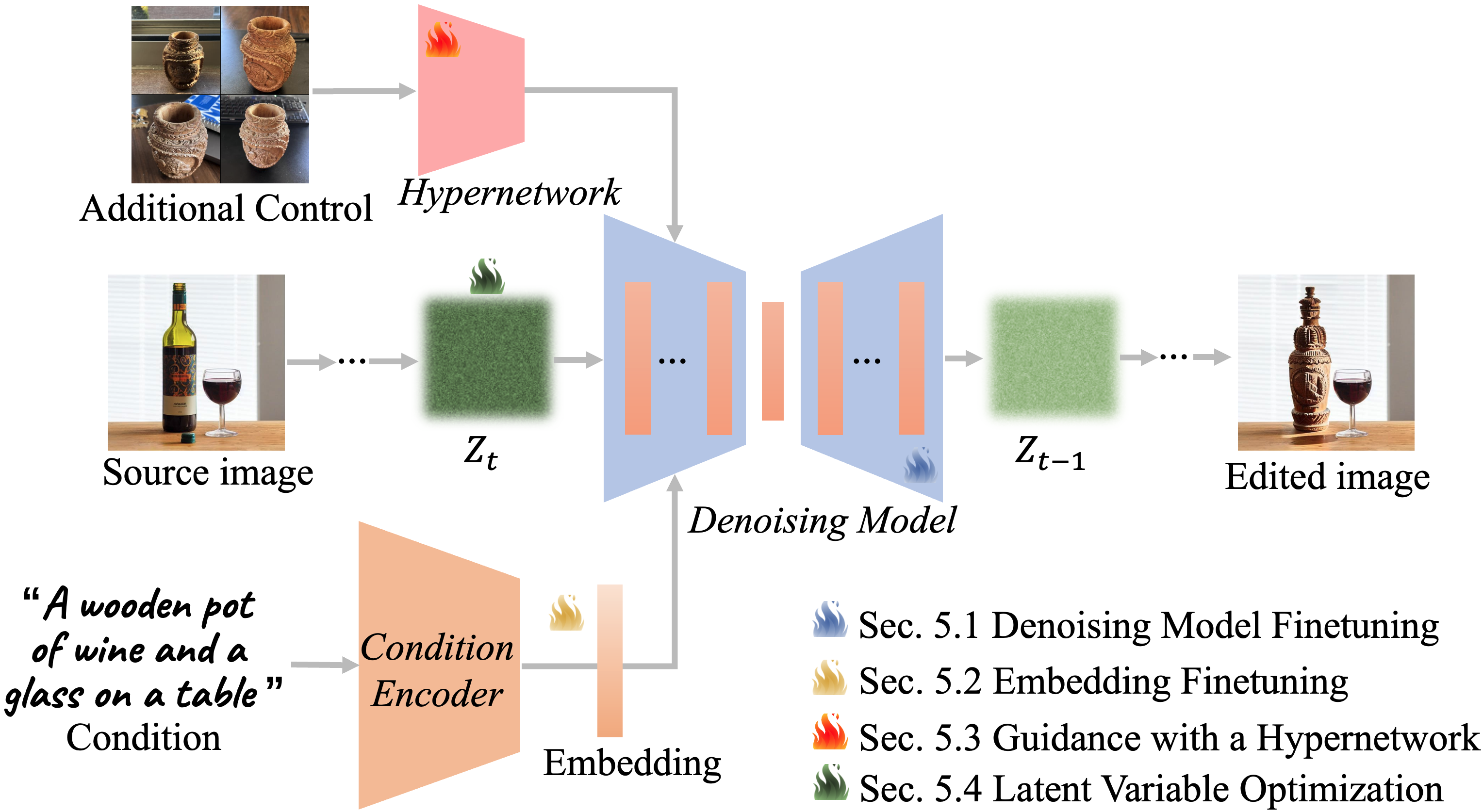}
            \vspace{-5pt}
		\caption{Testing-time finetuning framework with different finetuning components. The sample images are from Custom-Edit~\cite{choi2023custom}.}
		\label{fintune1}
  \vspace{-10pt}
	\end{figure}

Testing-time finetuning approaches usually do not involve a traditional pre-training phase but instead require finetuning during the inference (testing-time) stage for each specific image to be edited. These methods are designed to optimize the model’s performance for individual images, allowing for more precise and personalized edits. This section explores various finetuning strategies, as shown in Fig.~\ref{fintune1} and Fig.~ \ref{taxonomy_of_finetuning}, ranging from finetuning the entire model to adjusting specific layers or embeddings. These methods collectively demonstrate the evolving complexity and effectiveness of finetuning techniques in image editing, catering to various editing requirements and user intentions.

\subsection{Denoising Model Finetuning}
The denoising model is the central component in image generation and editing. Directly finetuning it is a simple and effective approach. Consequently, many editing methods are based on this finetuning process. Some of them involve finetuning the entire denoising model, while others focus on finetuning specific layers within the model.

\noindent \textbf{Finetuning Entire Denoising Models.}
Finetuning the entire denoising model allows the model to better learn specific features of images and more accurately interpret textual prompts, resulting in edits that more closely align with user intent. UniTune~\cite{valevski2022unitune} finetunes the diffusion model on a single base image during the tuning phase, encouraging the model to produce images similar to the base image. During the sampling phase, a modified sampling process is used to balance fidelity to the base image and alignment to the editing prompt. This includes starting sampling from a noisy version of the base image and applying classifier-free guidance during the sampling process. Custom-Edit~\cite{choi2023custom} uses a small set of reference images to personalize the diffusion model, enhancing the similarity of the edited image to the reference images while maintaining the similarity to the source image.
\input{table/taxonomy_of_finetuning}

\noindent \textbf{Partial Parameter Finetuning in Denoising Models.} Some methods focus on finetuning specific parts of the denoising model, such as the self-attention layers, cross-attention layers, encoder, or decoder. This type of finetuning is more precise, largely preserving the capabilities of the pretrained models and building upon them. KV Inversion~\cite{huang2023kv}, by learning keys (K) and values (V), designs an enhanced version of self-attention, termed Content-Preserving Self-Attention (CP-attn). This effectively addresses the issue of action editing on real images while maintaining the content and structure of the original images. It offers an efficient and flexible solution for image editing without the need for model finetuning or training on large-scale datasets.

\subsection{Embedding Finetuning}
Many finetuning approaches opt to target either text or null-text embeddings for refinement, allowing for better integration of embeddings with the generative process to achieve enhanced editing outcomes.

\noindent \textbf{Null-Text Embedding Finetuning.} 
The goal of null-text embedding finetuning is to solve the problem of reconstruction failures in DDIM Inversion~\cite{song2021denoising} and thus improve the consistency with the original image. In Null-Text Inversion~\cite{mokady2023null}, DDIM Inversion is first applied to the original image to obtain the inversion trajectory. Then, during the sampling process, the null-text embedding is finetuned to reduce the distance between the sampling trajectory and the inversion trajectory so that the sampling process can reconstruct the original image. The advantage of this approach is that neither the U-Net weights nor the text embedding are changed, so it is possible to improve reconstruction performance without changing the target prompt set by the user. Similarly, DPL~\cite{wang2023dynamic} dynamically updates nouns in the text prompt with a leakage fixation loss to address cross-attention leakage. The null-text embedding is finetuned to ensure high quality cross-attention maps and accurate image reconstruction.

\noindent \textbf{Text Embedding Finetuning.} Finetuning embeddings derived from input text can enhance image editing, making edited images more aligned with conditional characteristics. DiffusionDisentanglement~\cite{Wu_2023_CVPR} introduces a simple lightweight image editing algorithm that achieves style matching and content preservation by optimizing the blending weights of two text embeddings. This process involves optimizing about 50 parameters, with the optimized weights generalizing well across different images. Prompt Tuning Inversion~\cite{dong2023prompt} designs an accurate and fast inversion technique for text-based image editing, comprising a reconstruction phase and an editing phase. In the reconstruction phase, it encodes the information of the input image into a learnable text embedding. During the editing phase, a new text embedding is computed through linear interpolation, combining the target embedding with the optimized one to achieve effective editing while maintaining high fidelity.

\subsection{Guidance with a Hypernetwork}
Beyond conventional generative frameworks, some methods incorporate a custom network to better align with specific editing intentions. StyleDiffusion~\cite{li2023stylediffusion} introduces a Mapping Network that maps features of the input image to an embedding space aligned with the embedding space of textual prompts, effectively generating a prompt embedding. Cross-attention layers are used to combine textual prompt embeddings with image feature representations. These layers achieve text-image interaction by computing attention maps of keys, values, and queries. InST~\cite{Zhang_2023_inst} integrates a multi-layer cross-attention mechanism in its Textual Inversion segment to process image embeddings. The learned key information is transformed into a text embedding, which can be seen as a ``new word" representing the unique style of the artwork, effectively expressing its distinctive style.

\subsection{Latent Variable Optimization}
The direct optimization of an image's latent variable is also a technique employed during the finetuning process. This approach involves directly optimizing the noisy latents by introducing certain loss function relationships and features of some intermediate layers instead of optimizing the parameters of the generator or the embedded conditional parameters. Utilizing a pretrained diffusion model, most methods acquire its capability to perform image translation without the need of paired training data.

\noindent \textbf{Human-Guided Optimization of Latent Variables.} This approach allows users to participate in the image editing process, guiding the generation of images. Represented by DragGAN~\cite{pan2023drag}, this interactive editing process enables users to specify and move specific points within an image to new locations while keeping the rest of the image unchanged. DragGAN achieves this editing by optimizing the latent space of GANs. Subsequently, there are developments based on diffusion models, such as DragonDiffusion~\cite{mou2023dragondiffusion}, which constructs an energy function in the intermediate features of the diffusion model to guide editing. This enables image editing guidance directly through image features, independent of textual descriptions. DragDiffusion~\cite{shi2023dragdiffusion} concentrates on optimizing the diffusion latent representations at a specific time step, rather than across multiple time steps. This design is based on the observation that the U-Net feature maps at a particular time step offer ample semantic and geometric information to facilitate drag-and-drop editing.

\input{table/taxonomy_of_free}

\noindent \textbf{Utilizing Network Layers and Input for Optimizing Latent Variables.} Some optimization methods utilize embeddings or network features derived from input condition to construct loss functions, thereby enabling direct optimization of latent variables. The DDS~\cite{hertz2023delta} method leverages two image-text pairs: one comprising a source image and its descriptive text, and the other a target image with its corresponding descriptive text. DDS calculates the discrepancy between these two image-text pairs, deriving the loss through this comparison. The loss function in DiffuseIT~\cite{kwon2022diffusion} also incorporates the use of the CLIP model's text and image encoders to compute the similarity between the target text and the source image. CDS~\cite{nam2023contrastive} integrates a contrastive learning loss function into the DDS framework, utilizing the spatially rich features of the self-attention layers of LDM~\cite{rombach2022high} to guide image editing through contrastive loss calculation.

\subsection{Hybrid Finetuning} Some works combine the various finetuning approaches mentioned above, which can be sequential, with stages of tuning occurring in series, or conducted simultaneously as part of a single integrated workflow. Such composite finetuning methods can achieve targeted and effective image editing.

\noindent \textbf{Text Embedding \& Denoising Model Finetuning.} Imagic~\cite{kawar2023imagic} implements its goal in stages, starting with converting target text into a text embedding, which is then optimized to reconstruct an input image by minimizing the difference between the embedding and the image. Simultaneously, the diffusion model is finetuned for better image reconstruction. A midpoint is found through linear interpolation between the optimized text embedding and the target text, combining features of both. This embedding is then used by the finetuned diffusion model to generate the final edited image. LayerDiffusion~\cite{li2023layerdiffusion} optimizes text embeddings to match an input image's background and employs a layered diffusion strategy to finetune the model, enhancing its ability to maintain subject and background consistency. Forgedit~\cite{zhang2023forgedit} focuses on rapid image reconstruction and finding suitable text embeddings for editing, utilizing the diffusion model’s encoder and decoder for learning image layout and texture details, respectively.

\noindent \textbf{Text Encoder \& Denoising Model Finetuning.} SINE~\cite{zhang2023sine} initially finetunes the text encoder and the denoising model to better understand the content and geometric structure of individual images. It introduces a patch-based finetuning strategy, allowing the model to generate images at any resolution, not just the fixed resolution of the pretrained model. Through this finetuning and patch training, SINE is capable of handling single-image editing tasks, including but not limited to style transfer, content addition, and object manipulation.

%% file: table/taxonomy_of_finetuning.tex
\newcommand{\freetaxsize}{\scriptsize}
\tikzstyle{my-box}=[
    rectangle,
    draw=hidden-draw,
    rounded corners,
    text opacity=1,
    minimum height=1.5em,
    minimum width=5em,
    inner sep=2pt,
    align=center,
    fill opacity=.5,
    line width=0.8pt,
]
\tikzstyle{leaf}=[my-box, minimum height=1.5em,
    fill=hidden-pink!80, text=black, align=left,font=\freetaxsize,
    inner xsep=2pt,
    inner ysep=4pt,
    line width=0.8pt,
]
\begin{figure*}[t!]
    \centering
    \scalebox{1}{
        \begin{forest}
            forked edges,
            for tree={
                grow=east,
                reversed=true,
                anchor=base west,
                parent anchor=east,
                child anchor=west,
                base=left,
                font=\freetaxsize,
                rectangle,
                draw=hidden-draw,
                rounded corners,
                align=left,
                minimum width=2em,
                edge+={darkgray, line width=1pt},
                s sep=3pt,
                inner xsep=2pt,
                inner ysep=3pt,
                line width=0.8pt,
                ver/.style={rotate=90, child anchor=north, parent anchor=south, anchor=center},
                transform shape, 
            },
            where level=1{text width=4.5em,font=\freetaxsize,}{},
            where level=2{text width=3.1em,font=\freetaxsize,}{},
            where level=3{text width=3.0em,font=\freetaxsize,}{},
            [
                Testing-Time Finetuning\\ Approaches (Section ~\ref{finetune})
                [
                    Denoising Model Finetuning, text width=9em
                    [
                        UniTune~\cite{valevski2022unitune}{,} Custom-Edit~\cite{choi2023custom}{,} KV-Inversion~\cite{huang2023kv}
                        , leaf, text width=17em
                    ]
                ]
                [
                    Embedding Finetuning, text width=7.5em
                    [
                        Null-Text Inversion~\cite{mokady2023null}{,} DPL~\cite{wang2023dynamic}{,} DiffusionDisentanglement~\cite{Wu_2023_CVPR}{,} Prompt Tuning Inversion~\cite{dong2023prompt}
                        , leaf, text width=32em
                    ]
                ]
                [
                    Guidance with a Hypernetwork, text width=10em
                    [
                        StyleDiffusion~\cite{li2023stylediffusion}{,} InST~\cite{Zhang_2023_inst}
                        , leaf, text width=10.5em
                    ]
                ]
                [
                    Latent Variable Optimization, text width=9em
                    [
                        DragonDiffusion~\cite{mou2023dragondiffusion}{,} DragDiffusion~\cite{shi2023dragdiffusion}{,} DDS~\cite{hertz2023delta}{,} DiffuseIT~\cite{kwon2022diffusion}{,} CDS~\cite{nam2023contrastive}{,} MagicRemover~\cite{yang2023magicremover}
                        , leaf, text width=33em
                    ]
                ]
                [
                     Hybrid Finetuning, text width=6em
                    [
                        Imagic~\cite{kawar2023imagic}{,} LayerDiffusion~\cite{li2023layerdiffusion}{,} Forgedit~\cite{zhang2023forgedit}{,} SINE~\cite{zhang2023sine}
                        , leaf, text width=20em
                    ]
                ]
            ]
        \end{forest}
    }
    \vspace{-5pt}
    \caption{Taxonomy of testing-time finetuning approaches for image editing.}
    \vspace{-10pt}
    \label{taxonomy_of_finetuning}
\end{figure*}
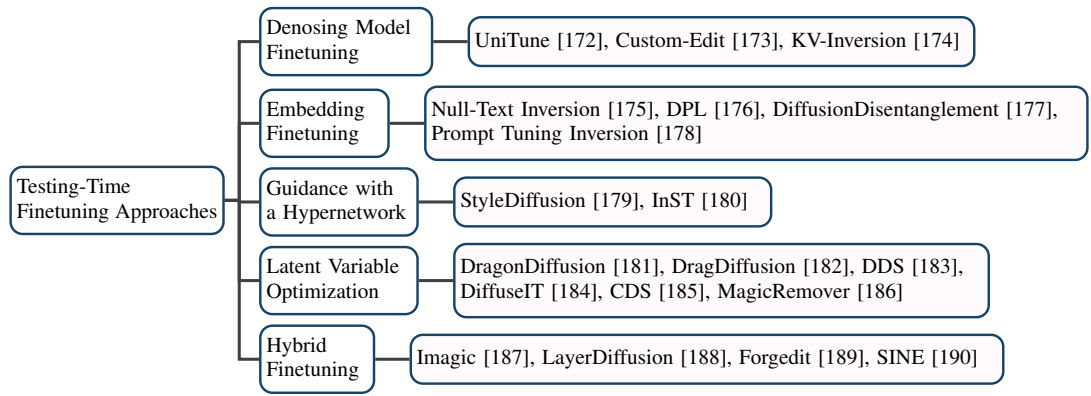

%% file: table/taxonomy_of_free.tex
\tikzstyle{my-box}=[
    rectangle,
    draw=hidden-draw,
    rounded corners,
    text opacity=1,
    minimum height=1.5em,
    minimum width=5em,
    inner sep=2pt,
    align=center,
    fill opacity=.5,
    line width=0.8pt,
]
\tikzstyle{leaf}=[my-box, minimum height=1.5em,
    fill=hidden-pink!80, text=black, align=left,font=\taxsize,
    inner xsep=2pt,
    inner ysep=4pt,
    line width=0.8pt,
]
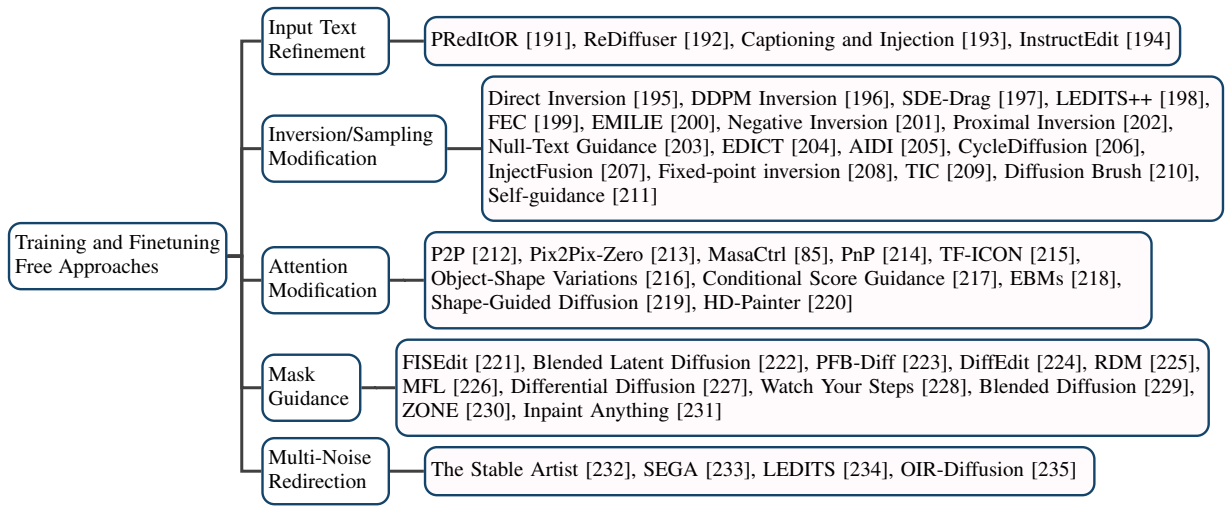
\begin{figure*}[t]
    \centering
    \scalebox{1}{
        \begin{forest}
            forked edges,
            for tree={
                grow=east,
                reversed=true,
                anchor=base west,
                parent anchor=east,
                child anchor=west,
                base=left,
                font=\taxsize,
                rectangle,
                draw=hidden-draw,
                rounded corners,
                align=left,
                minimum width=2em,
                edge+={darkgray, line width=1pt},
                s sep=3pt,
                inner xsep=2pt,
                inner ysep=3pt,
                line width=0.8pt,
                ver/.style={rotate=90, child anchor=north, parent anchor=south, anchor=center},
                transform shape, 
            },
            where level=1{text width=4.5em,font=\taxsize,}{},
            where level=2{text width=3.1em,font=\taxsize,}{},
            where level=3{text width=3.0em,font=\taxsize,}{},
            [
                Training and Finetuning Free\\ Approaches (Section~\ref{free})
                [
                    Input Text Refinement, text width=7em
                    [
                        PRedItOR~\cite{ravi2023preditor}{,} ReDiffuser~\cite{lin2023regeneration}{,} Captioning and Injection~\cite{kim2023user}{,} InstructEdit~\cite{wang2023instructedit}
                        , leaf, text width=26em
                    ]
                ]
                [
                    Inversion/Sampling \\Modification, text width=6em
                    [
                        Direct Inversion~\cite{elarabawy2022direct}{,} DDPM Inversion~\cite{huberman2023edit}{,} SDE-Drag~\cite{nie2023blessing}{,} LEDITS++~\cite{brack2023ledits++}{,} FEC~\cite{chen2023fec}{,} \\  EMILIE~\cite{joseph2024iterative}{,} Negative Inversion~\cite{miyake2023negative}{,} Proximal Inversion~\cite{han2024proxedit}{,} Null-Text Guidance~\cite{zhao2023null}{,} \\EDICT~\cite{wallace2023edict}{,} AIDI~\cite{pan2023effective}{,} CycleDiffusion~\cite{wu2023latent}{,} InjectFusion~\cite{jeong2024training}{,} Fixed-point inversion~\cite{meiri2023fixed}{,}\\ TIC~\cite{duan2023tuning}{,} Diffusion Brush~\cite{gholami2023diffusion}{,} Self-guidance~\cite{epstein2023diffusion}
                        , leaf, text width=30em
                    ]
                ]
                [
                    Attention \\Modification, text width=4em
                    [
                        P2P~\cite{hertz2022prompt}{,} Pix2Pix-Zero~\cite{parmar2023zero}{,} MasaCtrl~\cite{cao2023masactrl}{,} PnP~\cite{tumanyan2023plug}{,} TF-ICON~\cite{lu2023tf}{,} Object-Shape Variations~\cite{patashnik2023localizing}{,}\\ Conditional Score Guidance~\cite{lee2023conditional}{,} EBMs~\cite{park2023energy}{,} Shape-Guided Diffusion~\cite{park2024shape}{,} HD-Painter~\cite{manukyan2023hd}
                        , leaf, text width=33em
                    ]
                ]
                [
                    Mask\\ Guidance, text width=3em
                    [
                        FISEdit~\cite{yu2023fisedit}{,} Blended Latent Diffusion~\cite{avrahami2023blended}{,} PFB-Diff~\cite{huang2023pfb}{,} DiffEdit~\cite{couairon2023diffedit}{,} RDM~\cite{huang2023region}{,} MFL~\cite{liu2023text}{,}  \\Differential Diffusion~\cite{levin2023differential}{,} Watch Your Steps~\cite{mirzaei2023watch}{,} Blended Diffusion~\cite{avrahami2022blended}{,} ZONE~\cite{li2023zone}{,} Inpaint Anything~\cite{yu2023inpaint}
                        , leaf, text width=36em
                    ]
                ]
                [
                    Multi-Noise Redirection, text width=7.5em
                    [
                        The Stable Artist~\cite{brack2022stable}{,} SEGA~\cite{brack2023sega}{,} LEDITS~\cite{tsaban2023ledits}{,} OIR-Diffusion~\cite{yang2023object}
                        , leaf, text width=23em
                    ]
                ]
            ]
        \end{forest}
    }
    \vspace{-5pt}
    \caption{Taxonomy of training and finetuning free approaches for image editing.}
    \vspace{-10pt}
    \label{taxonomy_of_free}
\end{figure*}

%% file: sec/sec6_free.tex
 \section{Training and Finetuning Free Approaches}\label{free}
\begin{figure*}[!th]
    \small
    \setlength{\abovecaptionskip}{-0.1cm}
    \centering
    \includegraphics[width=\textwidth]{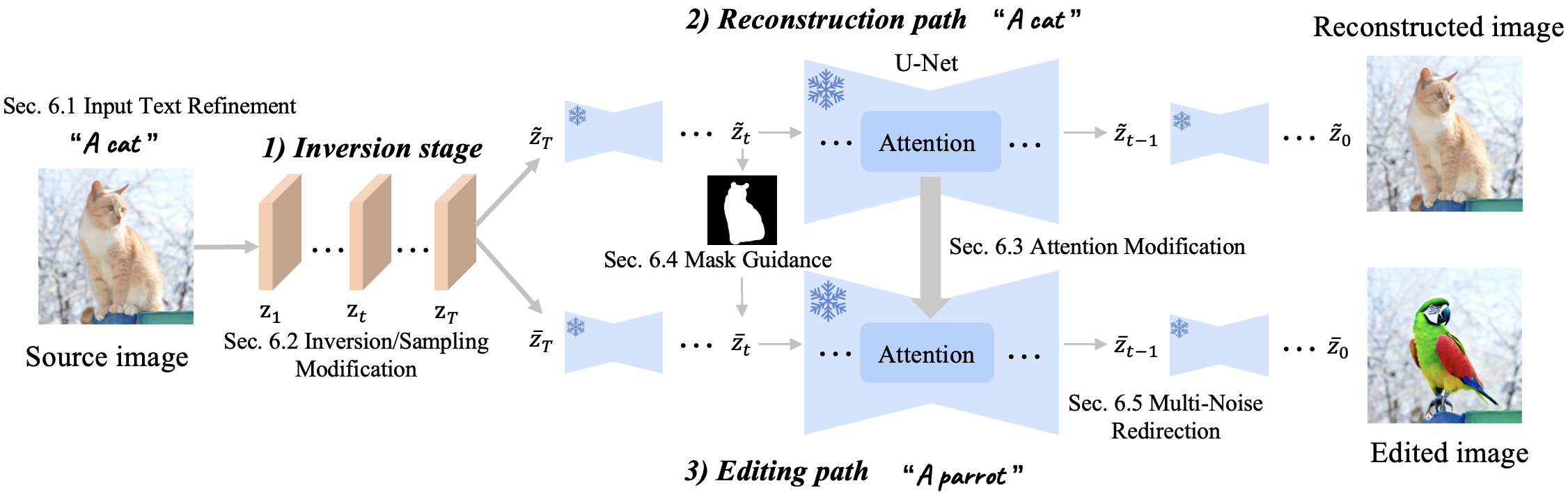}
        \vspace{-5pt}
    \caption{Common framework of training and finetuning free methods, where the modifications described in different sections are indicated. The sample images are from LEDITS++ \cite{brack2023ledits++}.}
    \label{fig:free_pipe}
        \vspace{-10pt}
\end{figure*}

Training and finetuning free approaches start from the premise that they are fast and low-cost. They do not require any form of training (on the dataset) or finetuning (on the source image) during the editing process. This section categorizes these methods into five groups based on the specific elements they modify within the image, as shown in Figs.~\ref{taxonomy_of_free} and~\ref{fig:free_pipe}, highlighting how they leverage the inherent principles of diffusion models for efficient editing.

\subsection{Input Text Refinement}
Input text refinement in the realm of image editing marks a significant progression in refining the text-to-image translation mechanism. This approach emphasizes the enhancement of text embeddings and the streamlining of user inputs, trying to guarantee that images are modified both accurately and contextually in accordance with the given text. It enables conceptual modifications and intuitive user instructions, getting rid of the necessity for complex model modifications.

\noindent \textbf{Enhancement of Input Text Embeddings.}
These methods concentrate on the refinement of text embeddings to modify the text-to-image translation.
PRedItOR \cite{ravi2023preditor} enhances input text embeddings by leveraging a diffusion prior model to perform conceptual edits in the CLIP image embedding space, allowing for more nuanced and context-aware image editing. ReDiffuser \cite{lin2023regeneration} generates rich prompts through regeneration learning, which are then fused to guide the editing process, ensuring that the desired features are enhanced while maintaining the source image structure.
Additionally, Captioning and Injection \cite{kim2023user} introduces a user-friendly image editing method by employing a captioning model such as BLIP \cite{li2022blip} and prompt injection techniques \cite{kim2023user} to augment input text embeddings. This method allows the user to provide minimal text input, such as a single noun, and then automatically generates or optimizes detailed prompts that effectively guide the editing process, resulting in more accurate and contextually relevant image editing.

\noindent \textbf{Instruction-Based Textual Guidance.}
This category enables fine-grained control over the image editing process via user instructions. It represents a significant step forward in that users to steer the editing process with less technical expertise.
InstructEdit \cite{wang2023instructedit} introduces a framework comprising a language processor, a segmentation model like SAM \cite{kirillov2023segment}, and an image editing model. The language processor interprets user instructions into segmentation prompts and captions, which are subsequently utilized to create masks and guide the image editing process. This approach shows the capability of instruction-based textual guidance in achieving detailed and precise image editing.

\subsection{Inversion/Sampling Modification}
Modification of inversion and sampling formulas are common techniques in training and finetuning free methods~\cite{elarabawy2022direct,huberman2023edit,miyake2023negative}. The inversion process is used to invert a real image into a noisy latent space, and then the sampling process is used to generate an edited result given the target prompt. Among them, DDIM Inversion~\cite{song2021denoising} is the most commonly used baseline method, although it often leads to the failure of reconstructing the source image~\cite{wallace2023edict,huberman2023edit,miyake2023negative}. Direct Inversion~\cite{elarabawy2022direct} pioneers a paradigm that edits a real image by changing the source prompt to the target prompt, showcasing its ability to handle diverse tasks. However, it still faces the reconstruction failure problem. Thus, several methods~\cite{wallace2023edict,huberman2023edit,miyake2023negative} modify the inversion and sampling formulas to improve the reconstruction capability.

\noindent \textbf{Reconstruction Information Memory.}
DDPM Inversion~\cite{huberman2023edit}, SDE-Drag~\cite{nie2023blessing}, LEDITS++~\cite{brack2023ledits++}, FEC~\cite{chen2023fec}, and EMILIE~\cite{joseph2024iterative} all belong to the category of reconstruction information memory; they save the information in the inversion stage and use it in the sampling stage to ensure reconstruction. DDPM Inversion explores the latent noise space of DDPM, by preserving all noisy latents during the inversion stage, thus achieving user-friendly image editing while ensuring reconstruction performance. SDE-Drag introduces stochastic differential equations as a unified and improved framework for editing and shows its superiority in drag-conditioned editing. LEDITS++ addresses inefficiencies in DDIM Inversion, presenting a DPM-Solver~\cite{lu2022dpm} inversion approach. FEC focuses on consistency enhancement, providing three techniques for saving the information during inversion to improve the reconstruction consistency. EMILIE pioneers iterative multi-granular editing, offering iterative capabilities and multi-granular control for desired changes. These methods ensure reconstruction performance by memorizing certain inversion information, and on this basis, enhance the diversity of editing operations.

\noindent \textbf{Utilising Null-Text in Sampling.}
Negative Inversion~\cite{miyake2023negative}, ProxEdit~\cite{han2024proxedit}, and Null-Text Guidance~\cite{zhao2023null} all explore the impact of null-text (also known as negative prompts) information on image editing. Negative Inversion uses the source prompt as the negative prompt during sampling, achieving comparable reconstruction quality to optimization-based approaches. ProxEdit addresses degradation in DDIM Inversion reconstruction with larger classifier-free guidance scales, and proposes an efficient solution through proximal guidance and mutual self-attention control. Meanwhile, Null-Text Guidance~\cite{zhao2023null} leverages a disturbance scheme to transform generated images into cartoons without the need for training. Despite the different uses of null-text by these methods, they all explore the role of null-text or negative prompt settings in image editing from different perspectives.

\noindent \textbf{Single-Step Multiple Noise Prediction.}
EDICT~\cite{wallace2023edict} and AIDI~\cite{pan2023effective} propose to predict multiple noises during single-step sampling to solve the problem of reconstruction failure. EDICT, drawing inspiration from affine coupling layers~\cite{wallace2023edict}, achieves mathematically exact inversion of real images. It maintains coupled noise vectors, enabling precise image reconstruction without local linearization assumptions, outperforming DDIM Inversion. On the other hand, AIDI introduces accelerated iterative diffusion inversion, and focuses on improving reconstruction accuracy by predicting multiple noises during single-step inversion. By employing a blended guidance technique, AIDI demonstrates effective results in various image editing tasks with a low classifier-free guidance scale ranging from 1 to 3. 

\subsection{Attention Modification}
Attention modification methods enhance the operations in the attention layers, which is the most common and direct way of training-free image editing. In the U-Net of Stable Diffusion, there are many cross-attention and self-attention layers, and the attention maps and feature maps in these layers contain a lot of semantic information. The common characteristic of attention modification entails discerning the intrinsic principles in the attention layers, and then utilizing them by modifying the attention operations. Prompt-to-Prompt (P2P)\cite{hertz2022prompt} is the pioneering research of attention modification.

\noindent \textbf{Attention Map Replacement.} P2P introduces an intuitive prompt-to-prompt editing framework reliant solely on text inputs by identifying cross-attention layers as pivotal in governing the spatial relationship between image layout and prompt words. As shown in Fig.~\ref{fig:free_pipe}, given the reconstruction and editing paths, P2P replaces the attention maps of the editing path with the corresponding ones of the reconstruction path, resulting in consistency between the edited result and the source image. Unlike P2P, Pix2Pix-Zero~\cite{parmar2023zero} eliminates the need for user-defined text prompts in real image editing. It autonomously discovers editing directions in the text embedding space, preserving the original content structure. Also, it introduces cross-attention guidance in the U-Net to retain input image features. 

\noindent \textbf{Attention Feature Replacement.} Both MasaCtrl~\cite{cao2023masactrl} and PnP~\cite{tumanyan2023plug} emphasize replacing the attention features for consistency. MasaCtrl converts self-attention into mutual self-attention, which replaces the Key and Value features in the self-attention layer for action editing. A mask-guided mutual self-attention strategy further enhances consistency by addressing confusion of foreground and background. PnP enables fine-grained control over generated structures through spatial feature manipulation and self-attention, directly injecting guidance image features. While both contribute to feature replacement, MasaCtrl focuses on the Key and Value in the self-attention layer, while PnP emphasizes the Query and Key in the self-attention layer.

\noindent \textbf{Local Attention Map Modification.}
The TF-ICON~\cite{lu2023tf} framework and Object-Shape Variation~\cite{patashnik2023localizing} share the utilization of local attention map modification. TF-ICON focuses on cross-domain image-guided composition, seamlessly integrating user-provided objects into visual contexts without additional training. It introduces the exceptional prompt for accurate image inversion and deformation mapping of the source image self-attention. Meanwhile, Object-Shape Variation~\cite{patashnik2023localizing} aims to generate collections depicting shape variations of specific objects in text-to-image workflows. It employs prompt-mixing for shape choices and proposes a localization technique using self-attention layers.

\noindent \textbf{Attention Score Guidance.}
Both Conditional Score Guidance~\cite{lee2023conditional} and EBMs~\cite{park2023energy} work on using different attention score guidances. Conditional Score Guidance focuses on image-to-image translation by introducing a conditional score function. This function considers both the source image and text prompt to selectively edit regions while preserving others. The mixup technique enhances the fusion of unedited and edited regions, ensuring high-fidelity translation. EBMs tackle the problem of semantic misalignment with the target prompt. It introduces an energy-based framework for adaptive context control, modeling the posterior of context vectors to enhance semantic alignment. 

\subsection{Mask Guidance}

Mask guidance in diffusion-based image editing represents a kind of techniques for enhancing image editing. 
It includes techniques for enhancing denoising efficiency through selective processing, precise mask auto-generation for targeted image editing, and mask-guided regional focus to ensure localized modifications aligning accurately with specific regions of interest.
These approaches leverage masks to steer and refine the sampling process, offering improved precision, speed, and flexibility. 

\noindent \textbf{Mask-Enhanced Denoising Efficiency.} These methods utilize masks to enhance the efficiency of diffusion-based image editing. The methods \cite{yu2023fisedit, avrahami2023blended, huang2023pfb} prioritize speed improvements while ensuring high-quality results. By selectively processing image regions through masks, they effectively reduce computational demands and enhance overall efficiency.
FISEdit \cite{yu2023fisedit} introduces a cache-enabled sparse diffusion model that leverages semantic mapping between the minor modifications on the input text and the affected regions on the output image. 
It automatically identifies the affected image regions and utilizes the cached unchanged regions’ feature map to accelerate the inference process.
Blended Latent Diffusion \cite{avrahami2023blended} leverages the text-to-image Latent Diffusion Model \cite{rombach2022high}, which accelerates the diffusion process by functioning within a lower-dimensional latent space, and effectively obviates the requirement for gradient computations of CLIP \cite{radford2021learning} at every timestep of the diffusion model.
PFB-Diff \cite{huang2023pfb} seamlessly integrates text-conditioned generated content into the target image through multi-level feature blending and introduces an attention masking mechanism in the cross-attention layers to confine the impact of specific words to desired regions, improving the performance of background editing.

\noindent \textbf{Mask Auto-Generation.} 
These methods enable precise image editing by automatically generating masks.
DiffEdit \cite{couairon2023diffedit} simplifies semantic editing by automatically generating masks that isolate relevant regions for modification. This selective editing approach safeguards unedited regions and preserves their semantic integrity.
RDM \cite{huang2023region} introduces a region-aware diffusion model that seamlessly integrates masks to automatically pinpoint and edit regions of interest based on text-driven guidance.
MFL \cite{liu2023text} proposes a two-stage mask-free training paradigm tailored for text-guided image editing. In the first stage, a unified mask is obtained according to the source prompt, and then several candidate images are generated with the provided mask and the target prompt based on the diffusion model.

\noindent \textbf{Mask-Guided Regional Focus.} This line of research employs masks as navigational tools, steering the sampling process towards specific regions of interest. The methods \cite{levin2023differential, mirzaei2023watch} excel in localizing edits, ensuring that modifications align precisely with designated regions specified in text prompts or editing requirements. 
Differential Diffusion \cite{levin2023differential} pioneers a region-aware editing
paradigm where masks are employed to personalize
the intensity of change within each region of the source image. This allows a granular level of control, fostering exquisite details and artistic expression.
Watch Your Steps \cite{mirzaei2023watch}  utilizes masks as relevance maps to pinpoint desired edit regions, ensuring that only the most relevant regions are edited while preserving the other regions of the original image.

\subsection{Multi-Noise Redirection}
Multi-noise redirection is the process of predicting multiple noises in different directions in a single step and then redirecting them into a single noise. The advantage of this redirection is its capacity to enable the single noise to unify multiple distinct editing directions simultaneously, thereby more effectively meeting the user's editing requirements.

\noindent \textbf{Semantic Noise Steering.} 
These methods semantically guide the noise redirection in the sampling process and enhance fine-grained control over image content.
Stable Artist \cite{brack2022stable} and SEGA \cite{brack2023sega} showcase the ability to steer the diffusion process using semantic guidance. They facilitate refined edits in image composition through control over various semantic directions within the noise redirection process. 
LEDITS \cite{tsaban2023ledits} applies semantic noise steering by combining DDPM inversion~\cite{huberman2023edit} with SEGA, and facilitates versatile editing direction, encompassing both object editing and overall style change.

\noindent \textbf{Object-Aware Noise Redirection.} 

OIR-Diffusion \cite{yang2023object} introduces the object-aware inversion and reassembly method, which enables fine-grained editing of specific objects within an image by determining optimal inversion steps for each sample and seamlessly integrating the edited regions with the other regions. It significantly maintains the fidelity of the non-edited regions and achieves remarkable results in editing diverse object attributes, particularly in intricate multi-object scenarios.

%% file: sec/sec7_inpainting.tex
\section{Inpainting and Outpainting}\label{inpainting}

\begin{figure}[!t]
		\small
		\centering
		\includegraphics[width=0.48\textwidth]{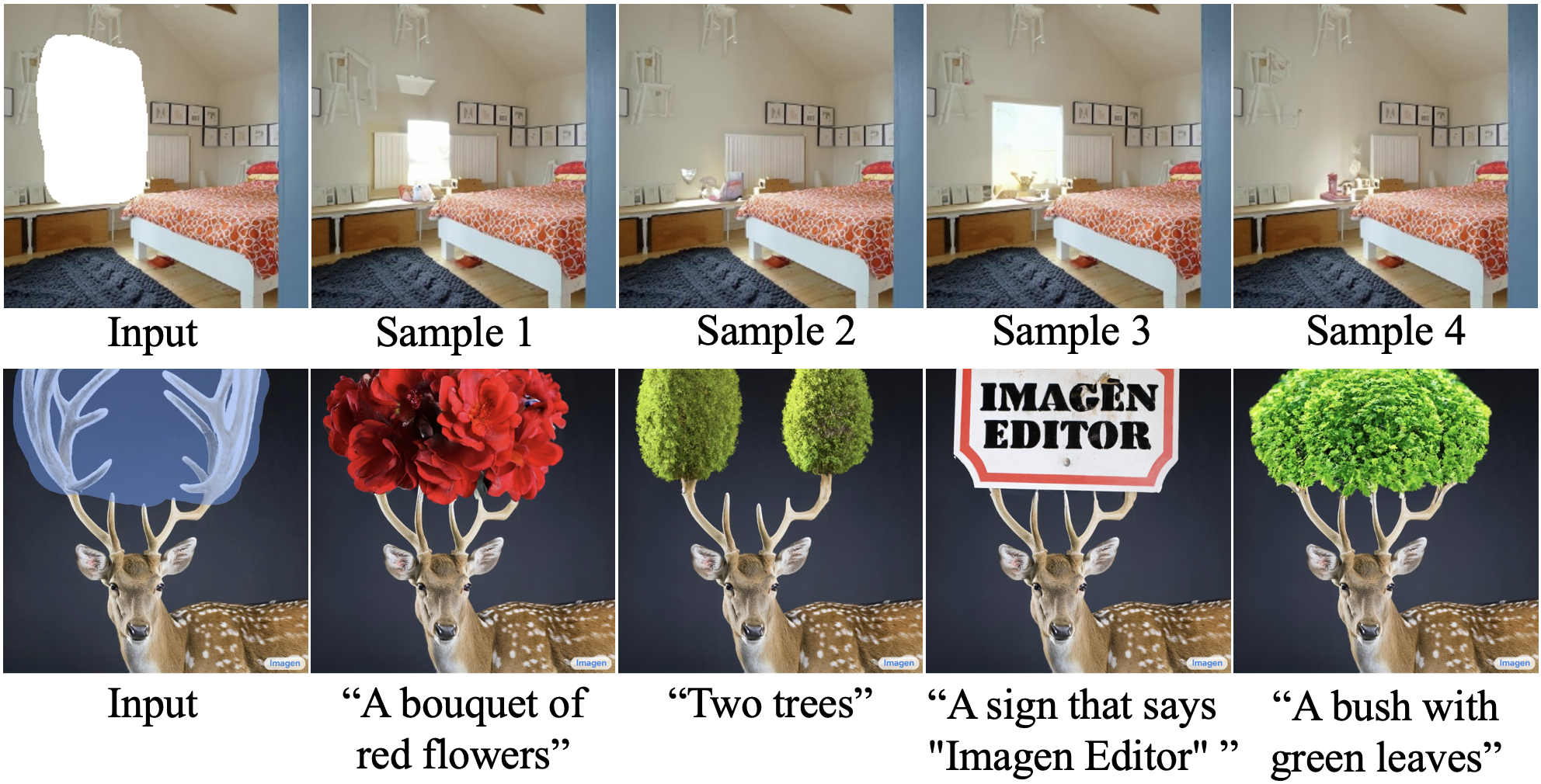}
            \vspace{-5pt}
		\caption{Visual comparison between traditional context-driven inpainting (top) and multimodal conditional inpainting (bottom). The samples in the two rows are from Palette \cite{saharia2022palette} and Imagen Editor \cite{wang2023imagen}, respectively.}
		\label{fig_inpainting}
  \vspace{-10pt}
\end{figure}

Image inpainting and outpainting, often regarded as sub-tasks of image editing, occupy unique positions with distinct objectives and challenges. We divide them into two major types (see Fig.~\ref{fig_inpainting}) for better explanation, as detailed in Sections \ref{sec7_1} and \ref{sec_7_2}.

\subsection{Traditional Context-Driven Inpainting}
\label{sec7_1}

\subsubsection{Supervised Training}
The supervised training-based inpainting \& outpainting methods usually train a diffusion model from scratch with paired corrupted and complete images. Palette \cite{saharia2022palette} develops a unified framework for image-to-image translation based on conditional diffusion models and individually trains this framework on four image-to-image translation tasks, namely colorization, inpainting, outpainting, and JPEG restoration. It utilizes a direct concatenation of low-quality reference images with the denoised result at the $t-1$ step as the condition for noise prediction at the $t$ step. SUD$^2$ \cite{chan2023sud} builds upon and extends the SUD framework \cite{young2023supervision}, integrating additional strategies like correlation minimization, noise injection, and particularly the use of denoising diffusion models to address the scarcity of paired training data in semi-supervised learning.

\subsubsection{Zero-Shot Learning}

\noindent \textbf{Context Prior Integration.} This type of methods extracts structures and textures from uncorrupted image parts to complement masked regions, ensuring global content consistency. Repaint~\cite{lugmayr2022repaint} uses a pretrained unconditional diffusion model, modifying the denoising process by sampling masked areas from the model and unmasked areas from the image at each reverse step.
Similarly, GradPaint \cite{grechka2023gradpaint} employs a pretrained diffusion model to estimate masked content, using a custom loss function to measure coherence with unmasked regions and updating the content based on the loss gradient.

\noindent \textbf{Degradation Decomposition.} Image inpainting can be regarded as a specialized application of general linear inverse problems represented as $\mathbf{y} = \mathbf{H}\mathbf{x} + \mathbf{n}$,
where $\mathbf{H}$ is the degradation operator (e.g., a mask matrix indicating corrupted regions in inpainting), and $\mathbf{n}\sim\mathcal{N}(\mathbf{0},\sigma^2\mathbf{I})$. The goal is to recover the original image $\mathbf{x}$ from the corrupted image $\mathbf{y}$. Some works focus on decomposition on $\mathbf{H}$. DDRM \cite{kawar2022denoising} uses a pretrained diffusion model and Singular Value Decomposition (SVD) of $\mathbf{H}$ to transform this recovery into an efficient iterative diffusion process in the spectral space of $\mathbf{H}$. 
Differently, DDNM \cite{wang2023zero} applys a range-null space decomposition, dividing the image into two distinct components via the pseudo-inverse of the degradation matrix.

\noindent \textbf{Posterior Estimation}. To solve the general noisy linear inverse problem, several studies \cite{chung2022improving,chung2023diffusion, song2023pseudoinverse, fei2023generative, zhang2023towards, fabian2023diracdiffusion} focus on estimating the posterior distribution $p(\mathbf{x} | \mathbf{y})$, leveraging the unconditional diffusion model based on Bayes' theorem. The conditional posterior $ p( \mathbf{x}_t |  \mathbf{y})$ is calculated at each step of the reverse diffusion process: $p( \mathbf{x}_t| \mathbf{y}) = \frac{ p( \mathbf{y}| \mathbf{x}_t) p( \mathbf{x}_t)}{ p( \mathbf{y})}$. The corresponding gradient of the log-likelihood is then computed as $\nabla_{\mathbf{x}_t} \log p_t(\mathbf{x}_t|\mathbf{y}) = \nabla_{\mathbf{x}_t} \log p_t(\mathbf{y}|\mathbf{x}_t) + \nabla_{\mathbf{x}_t} \log p_t(\mathbf{x}_t)$,
where the score function $\nabla_{\mathbf{x}_t} \log p_t(\mathbf{x}_t)$ is estimated by a pretrained score network $\mathbf{s}_{\theta}(\mathbf{x}_t, t)$. Thus, the problem reduces to estimating $\nabla_{\mathbf{x}_t} \log p_t(\mathbf{y}|\mathbf{x}_t)$.
DPS \cite{chung2023diffusion} approximates the posterior $p_t(\mathbf{y}|\mathbf{x}_t)$ with $p_t(\mathbf{y}|\hat{\mathbf{x}}_0)$.
MCG \cite{chung2022improving} further introduces a projection onto the measurement subspace to ensure data consistency.
$\Pi$GDM \cite{song2023pseudoinverse} extends this approach to both linear and non-linear inverse problems through the use of the Moore-Penrose pseudo-inverse $\mathbf{H}^\dagger$ of the degradation function $\mathbf{H}$.
GDP \cite{fei2023generative} offers a different approximation strategy, incorporating an image distance metric and quality loss functions to guide the estimation process.
Similarly, CoPaint \cite{zhang2023towards} uses a one-step prediction of the final image to make inpainting errors more tractable during the denoising steps.

\subsection{Multimodal Conditional Inpainting}
\label{sec_7_2}

\subsubsection{Random Mask Training}

Traditional image inpainting, focusing on filling missing areas in images driven by surrounding context, often lacks precise control over the content. With the advance of text-to-image diffusion models, this limitation is being overcome
by injecting additional multimodal conditions such as text descriptions, segmentation maps, and reference images. These models, e.g., Stable Diffusion and Imagen, can also be adapted to the inpainting task where the noisy background is replaced with a noisy version of the original image in the reverse diffusion process. However, this can sometimes result in unsatisfactory samples due to limited global context visibility. Addressing this, models like GLIDE \cite{nichol2022glide} and Stable Inpainting (inpainting specialist v1.5 from Stable Diffusion \cite{rombach2022high}) further finetune the pretrained text-to-image models specialized for inpainting. They use a randomly generated mask along with the masked image and the full image caption, enabling the model to utilize information from the unmasked region during training. As pioneering works, they provide enlightening insights for subsequent research.

\subsubsection{Precise Control Conditioning}
An essential problem with random masking is that it may cover regions unrelated to the text prompt, leading the model to ignoring the prompt, especially when the masked regions are small. To provide more precise control, SmartBrush \cite{xie2023smartbrush} introduces a precision factor to produce various mask types. It allows users to choose between coarse masks, which include the desired object within the mask, and detailed masks that closely outline the object shape.
Imagen Editor \cite{wang2023imagen} finetunes Imagen \cite{saharia2022photorealistic} by integrating image and mask context into each diffusion stage with three convolutional downsampling image encoders. The mask here is generated on-the-fly by an object detector, SSD Mobilenet v2 \cite{sandler2018mobilenetv2}, instead of randomly masking.

PbE \cite{yang2023paint} and PhD \cite{zhang2023paste} focus on using reference images for users' personalized inpainting. Specifically, PbE uses CLIP to only extract subject information from the reference image, while PhD integrates both text and reference images in the training pipeline same as PbE's.
Uni-paint \cite{yang2023uni} first finetunes Stable Diffusion on the unmasked regions of the input images with null text and then uses various modes of guidance including unconditional, text-driven, stroke-driven, exemplar-driven inpainting, and a combination of them to perform multimodal inpainting during inference via different condition integration techniques.
PVA \cite{xu2024personalized} focuses on facial inpainting by encoding a few reference images with the same identity as the masked image through Transformer blocks. The encoded image embedding, combined with the CLIP text embedding, then serves as extra conditions of the Stable Inpainting model through a parallel visual attention mechanism.
SmartMask \cite{singh2023smartmask} utilizes semantic amodal segmentation data to create high-quality annotations for mask-free object insertion. It trains a diffusion model to predict instance maps from these annotations including both semantic maps and scene context.

\subsubsection{Pretrained Diffusion Prior Exploiting}
Despite significant advancements in training-based methods, there still exist challenges, particularly in gathering large-scale realistic data and the extensive resources required for model training. Consequently, some research shifts focus to specific inpainting tasks, harnessing the generative capabilities of pretrained text-to-image diffusion models via integrating various techniques.

Blended Diffusion \cite{avrahami2022blended} uses CLIP to compute the difference between the denoised image embedding from a pretrained diffusion model and the guided text embedding to refine generation within masked regions. It then blends the noisy masked regions with the noisy unmasked parts of the input image during each sampling step. Blended Latent Diffusion \cite{avrahami2023blended} takes this further by applying the technique in the LDM \cite{rombach2022high} for higher efficiency.

Inpaint Anything \cite{yu2023inpaint} combines SAM \cite{kirillov2023segment} with inpainting models like LaMa \cite{suvorov2022resolution}, providing a user-friendly framework for removing, replacing, or filling image parts with simple clicks.
MagicRemover \cite{yang2023magicremover} employs null-text inversion \cite{mokady2023null} to project an image into a noisy latent code and then optimizes the codes during sampling. It introduces an attention-based guidance strategy and a classifier optimization algorithm to enhance inpainting precision and stability with fewer steps.

HD-Painter \cite{manukyan2023hd} is designed for high-resolution image inpainting and operates in two stages: image completion and inpainting-specialized super-resolution. The first stage is carried out in the latent space of Stable Diffusion, where a prompt-aware introverted attention mechanism replaces the original self-attention, and an additional reweighting attention score guidance is employed. This modification ensures better alignment of the generated content with the input prompt. The second stage uses the Stable Diffusion Super-Resolution variant with the low-resolution inpainted image conditioned and mask blended.

\subsection{Outpainting}
Image outpainting, similar to but not the same as inpainting, aims to generate new pixels that seamlessly extend an image's boundaries. Existing T2I models like Stable Diffusion \cite{rombach2022high} and DALL-E \cite{ramesh2022hierarchical} can be generalized to address this task due to their training on extensive datasets including images of various sizes and shapes. It is often considered a special form of inpainting with similar implementations. For instance, Palette \cite{saharia2022palette} trains a diffusion model by directly combining cropped images with their original versions as input in a supervised manner.
PowerPaint \cite{zhuang2023task} divides inpainting and related tasks into four specific categories, viewing outpainting as a type of context-aware image inpainting. It enhances Stable Diffusion with learnable task-specific prompts and finetuning strategies to precisely direct the model towards distinct objectives.

%% file: sec/sec8_benchmark.tex
\section{Benchmark and Evaluation}\label{benchmark}

\subsection{Benchmark Construction}
In the previous sections, we delve into the methodology aspects of diffusion model-based image editing approaches. Beyond this analysis, it is crucial to assess these methods, examining their capabilities across various editing tasks. However, existing benchmarks for image editing are limited and do not fully meet the needs identified in our survey. For instance, EditBench \cite{wang2023imagen} primarily aims at text and mask-guided inpainting, and ignores broader tasks that involve global editing like style transfer. TedBench \cite{kawar2023imagic}, while it expands the range of tasks, lacks detailed instructions, which is essential for evaluating methods that rely on textual instructions rather than descriptions. Moreover, the EditVal benchmark \cite{basu2023editval}, while attempting to offer more comprehensive coverage of tasks and methods, is limited by the quality of its images from the MS-COCO dataset \cite{lin2014microsoft}, which are often low-resolution and blur.

To address these problems, we introduce EditEval, a benchmark designed to evaluate general diffusion model-based image editing methods. EditEval includes a carefully curated dataset of 150 high-quality images, each accompanied by text prompts. EditEval evaluates performance on 7 common editing tasks selected from Table~\ref{tab_taxonomy}. Additionally, we propose LMM Score, a quantitative evaluation metric that utilizes the capabilities of large multimodal models (LMMs) to assess the editing performance on different tasks. In addition to the objective evaluation provided by LMM Score, we also carry out a user study to incorporate subjective evaluation. The details of EditEval's construction and its application are described as follows.

\subsubsection{Task Selection}
When selecting evaluation tasks, we take into account the capabilities indicated in Table~\ref{tab_taxonomy}. It is observed that most methods are capable of handling semantic and stylistic tasks while encountering challenges with structural editing. The potential reason is that many current T2I diffusion models, which most editing methods depend on, have difficulty in accurate spatial awareness \cite{betker2023improving}. For example, they often generate inconsistent images with the prompts containing terms like ``to the left of", ``underneath", or ``behind".
Considering these factors and the practical applications, we choose seven common tasks for our benchmark: object addition, object replacement, object removal, background replacement, style change, texture change, and action change, which aim to provide a thorough evaluation of the editing method's performance from simple object edits to complex scene changes.

\subsubsection{Dataset Construction}
For image selection, we manually choose 150 images from \href{https://unsplash.com/}{Unsplash}’s online repository of professional photographs, ensuring a wide diversity of subjects and scenes. These images are cropped to a square format, aligning with the input ratio for most editing models. We then categorize the chosen images into 7 groups, each associated with one of the specific editing tasks mentioned above.

In generating prompts for each image, we employ an LMM to create a source prompt that describes the image's content, a target prompt outlining the expected result of the editing, and a corresponding instruction to guide the editing process. This step is facilitated by providing GPT-4V, one of the most widely-used LMMs, with a detailed template that includes an example pair and several instructions. The example pair consists of a task indicator (e.g., ``Object Removal"), a source image, a source description (e.g., ``A cup of coffee on an unmade bed"), a target description (e.g., ``An unmade bed"), and an instruction (e.g., ``Remove the cup of coffee"). Along with this example, we feed GPT-4V clear instructions on our expectations for generating source descriptions, target descriptions, and instructions for new images and task indicators we upload. This method ensures that GPT-4V produces prompts and instructions that not only align with the specified editing task but also maintain diversity in scenes and subjects. After GPT-4V generates the initial prompts and instructions, we undertake a meticulous examination to ensure each prompt and the set of instructions is specific, clear, and directly applicable to the corresponding image and task.

The final dataset, including the selected images, their source and target prompts, and the editing instructions, is available in the accompanying GitHub repository.

\subsubsection{Metric Design and Selection}
The field of image editing has traditionally relied on CLIPScore \cite{hessel2021clipscore} as a major quantitative evaluation metric. Despite its effectiveness in assessing the alignment between images and corresponding textual prompts, CLIPScore may struggle in complicated scenes with many details and specific spatial relationships \cite{basu2023editval}. This limitation urges the need for a more versatile and encompassing metric that can be applied to broader image editing tasks.
Hence, we propose LMM Score, a new metric by harnessing advanced visual-language understanding capabilities of large multimodal models (LMMs).

To develop this metric, we first direct GPT-4 to conceive a quantitative metric and outline a framework that allows for objective evaluation for general image editing tasks through appropriate user prompts. Based on GPT-4's recommendations, the evaluation framework encompasses the following elements, using a source image $I_{src}$ along with its text description $t_{src}$, a collection of edited images $I_{tgt}^i, i={1, 2, ..., N}$ generated by $N$ methods, an editing prompt $t_{edit}$, an editing instruction $t_{inst}$, and a task indicator $o$. The criteria integrates four crucial factors:
\begin{itemize}
    \item \textit{Editing Accuracy}: Evaluate how closely the edited image $I_{tgt}^i$ adheres to the specified editing prompt $t_{edit}$ and 
    instruction $t_{inst}$, measuring the precision of the editing.

    \item \textit{Contextual Preservation}: Assess how well the edited image $I_{tgt}^i$ maintains the context of the source image  $I_{src}$ that requires no changes.

    \item \textit{Visual Quality}: Assess the overall quality of the edited image $I_{tgt}^i$, including factors like resolution, absence of artifacts, color accuracy, sharpness, etc.

    \item \textit{Logical Realism}: Evaluate how logically realistic the edited image $I_{tgt}^i$ is, in terms of adherence to natural laws like lighting consistency, texture continuity, etc.
\end{itemize}
As for evaluation, each edited image $I_{tgt}^i$ is evaluated on these factors, yielding four sub-scores (ranging from 1 to 10) denoted as $S_{acc}$, $S_{pre}$, $S_{qua}$, and $S_{real}$. An overall score, $S_{LMM}$, is computed as a weighted average of these sub-scores to reflect the overall editing quality:
\begin{align}
S_{LMM}=0.4×S_{acc}+0.3×S_{pre}+0.2×S_{qua}+0.1×S_{real}.
\label{eq_lmm}
\end{align}
This weighted formula, suggested by GPT-4, aims to balance the relative importance of each evaluation factor.

Following the proposed framework, we employ an LMM, specifically GPT-4V in this work, to carry out this evaluation. GPT-4V follows user-predefined instructions to calculate the sub-scores across the four evaluation factors for each edited image, obtaining the value of $S_{LMM}$ as in Eq.~\ref{eq_lmm}. To enable a user-friendly application of LMM Score, we provide a template that includes predefined instructions and necessary materials, which is accessible in our GitHub repository, facilitating its adoption in image editing evaluation.

In addition to the objective evaluation provided by LMM Score, we conduct a user study to collect subjective feedback. This study engages 50 participants from diverse backgrounds, ensuring a wide range of perspectives. Each participant is presented with the source image, its description, the editing prompt and instruction, and a series of edited images. They are then asked to score each edited image according to the same four evaluation factors utilized by LMM Score. Each factor is scored from 1 to 10. The participants' scores are then aggregated to calculate an overall \textit{User Score} ($S_{User}$) for each image, with the same calculation in Eq.~\ref{eq_lmm}. The results from this user study complement the LMM Score evaluations, providing a comprehensive assessment of the performance of different editing methods.

\subsection{Evaluation}
\subsubsection{Method Selection}
For the evaluation of each editing task, we carefully select between 4 to 8 methods from Table \ref{tab_taxonomy}, encompassing a range of
\noindent 
\begin{minipage}{\columnwidth}
  \centering
\includegraphics[width=0.94\columnwidth]{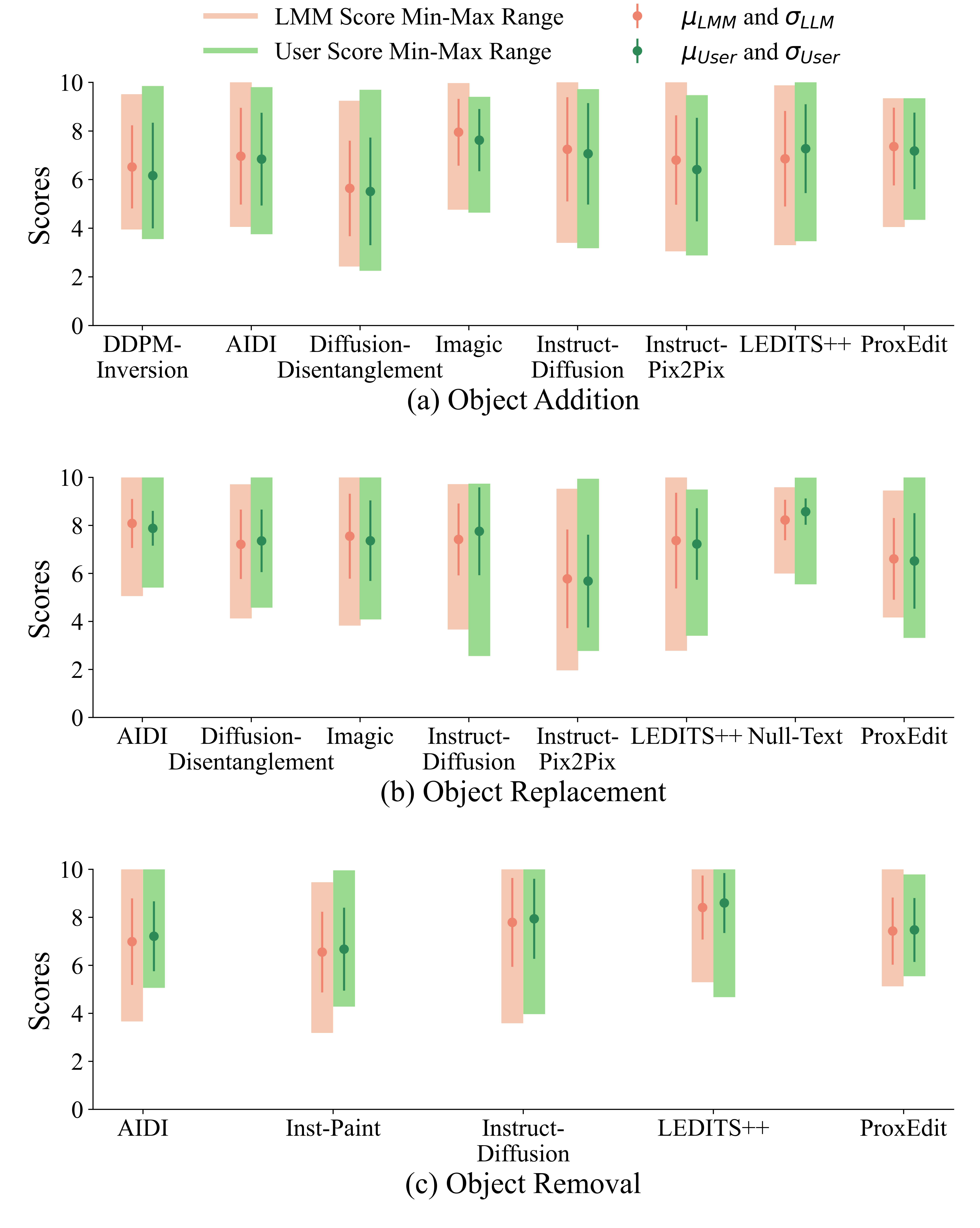}
\newline
\includegraphics[width=0.94\columnwidth]{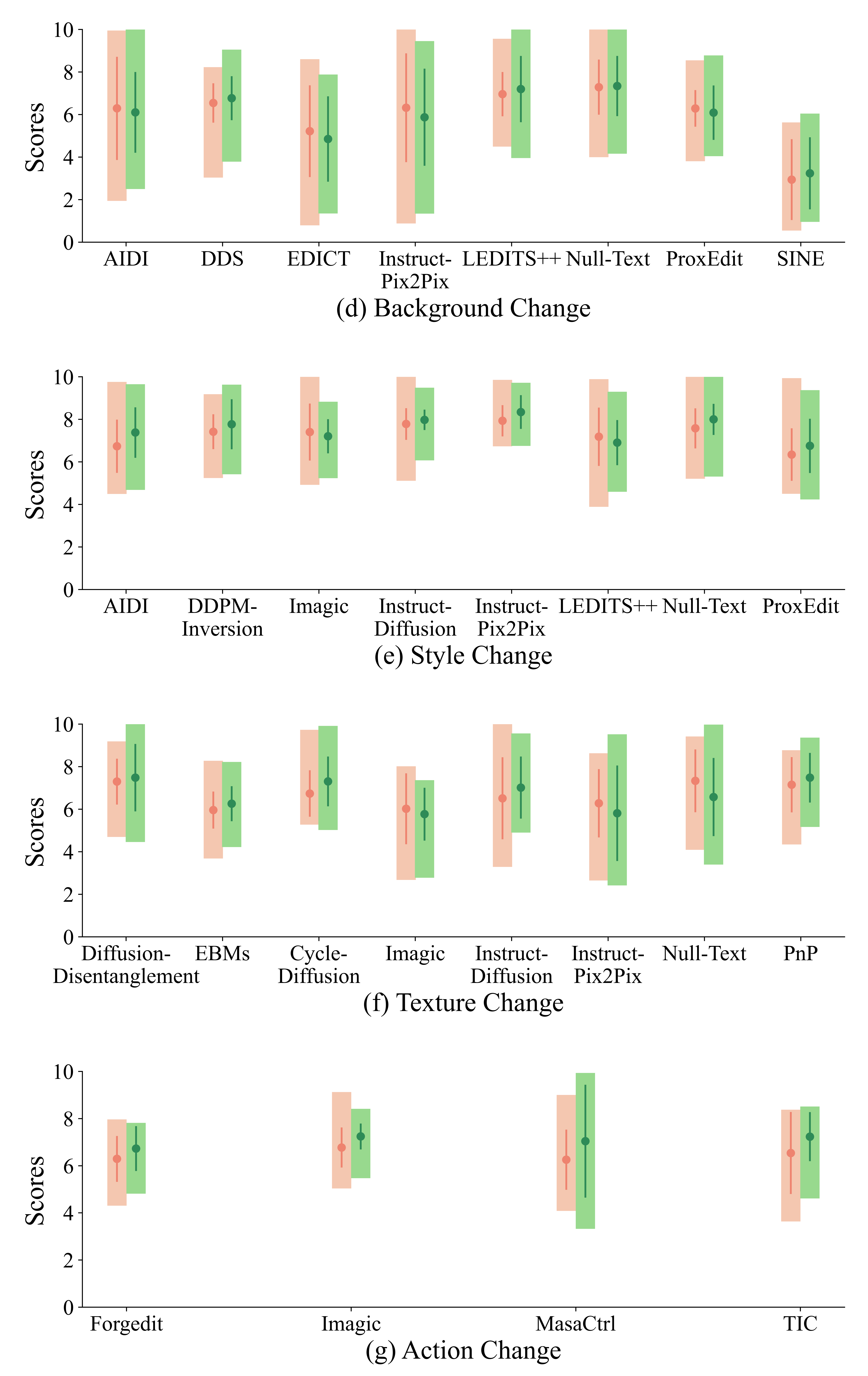}
\vspace{-0.275cm}
  \captionof{figure}{Quantitative performances of various methods across 7 selected editing tasks. $\mu$ and $\sigma$ denote the mean value and standard deviation of the scores, respectively.}
  \label{fig_evaluation_all}
\end{minipage}
\newline
\noindent
 approaches including training-based, testing-time finetuning, and training \& finetuning free methods. To ensure a fair and consistent comparison, our selection criteria are as follows: the method must require only text conditions, own the capability to handle specific tasks, and have open-source code available for implementation. We exclude domain-specific methods to avoid the constraints imposed by their limited applicability across diverse domains. These selected methods are specified in subsequent subsections.

\subsubsection{Comparative Analysis}

\noindent \textbf{Performance Comparison.}
To present a thorough evaluation of the selected methods across the 7 selected editing tasks, we compute the mean and standard deviation of both $S_{LMM}$ and $S_{User}$ on all evaluation samples, as shown in Fig.~\ref{fig_evaluation_all}. The color bars illustrate the range of scores from minimum to maximum. From the results, we can have several insights.
First, there is no single method that outperforms others across all tasks.
Second, most methods exhibit a wide range of fluctuating scores in each task, as evidenced by the broad span between the minimum and maximum values and the large standard deviations of both $S_{LMM}$ and $S_{User}$.
Such variability points to the unrobustness of these methods, suggesting that a method's performance can be sample-dependent and may not consistently meet the editing requirements in various scenes.
Third, some methods do exhibit stable and impressive performances for certain tasks. 
Last, the mean values of $S_{LMM}$ and $S_{User}$ are closely matched for each method within each task, suggesting that LMM Score well reflects users' preferences. Although the methods' performances fluctuate, the consistency between LMM Score and user perception across the tasks indicates that LMM Score is a reliable metric for image editing. Besides the quantitative results, we also present several visual examples for qualitative comparison in Fig.~\ref{fig_visual_editing}.

\begin{figure}[!t]
		\small
		\centering
		\includegraphics[width=0.495\textwidth]{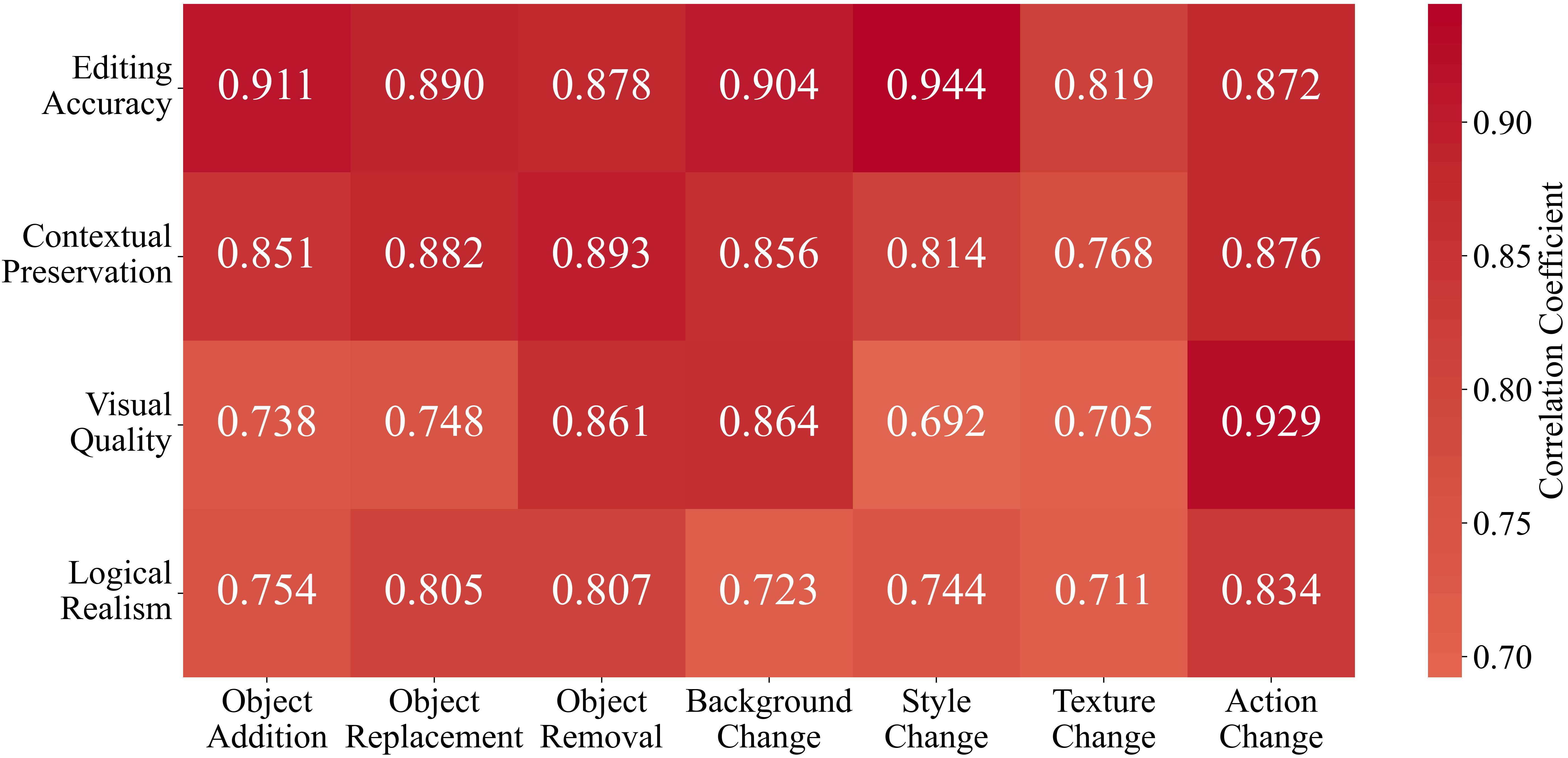}
        \vspace{-5pt}
		\caption{Pearson correlation coefficients between LMM Score and the user study.}
		\label{heatmap}
			\vspace{-10pt}
\end{figure}

\noindent \textbf{Correlation Between LMM Score and User Study.}
To examine the effectiveness of LMM Score, we investigate the correlation between it and the user study. Specifically, we first calculate the 4 sub-scores' means of the collected user data on the 4 factors. Then, within each editing task, we compute the Pearson correlation coefficient on each factor between the LMM sub-scores and the user study sub-scores on each sample.
Finally, these coefficients are averaged across all samples to obtain the overall correlation for each factor per task. The results, as illustrated in Fig.~\ref{heatmap}, reveal a significant alignment across the editing tasks, indicating a strong concordance between the objective LMM Score evaluation and the subjective human judgment.

\begin{figure}[!t]
		\small
		\centering
		\includegraphics[width=0.38\textwidth]{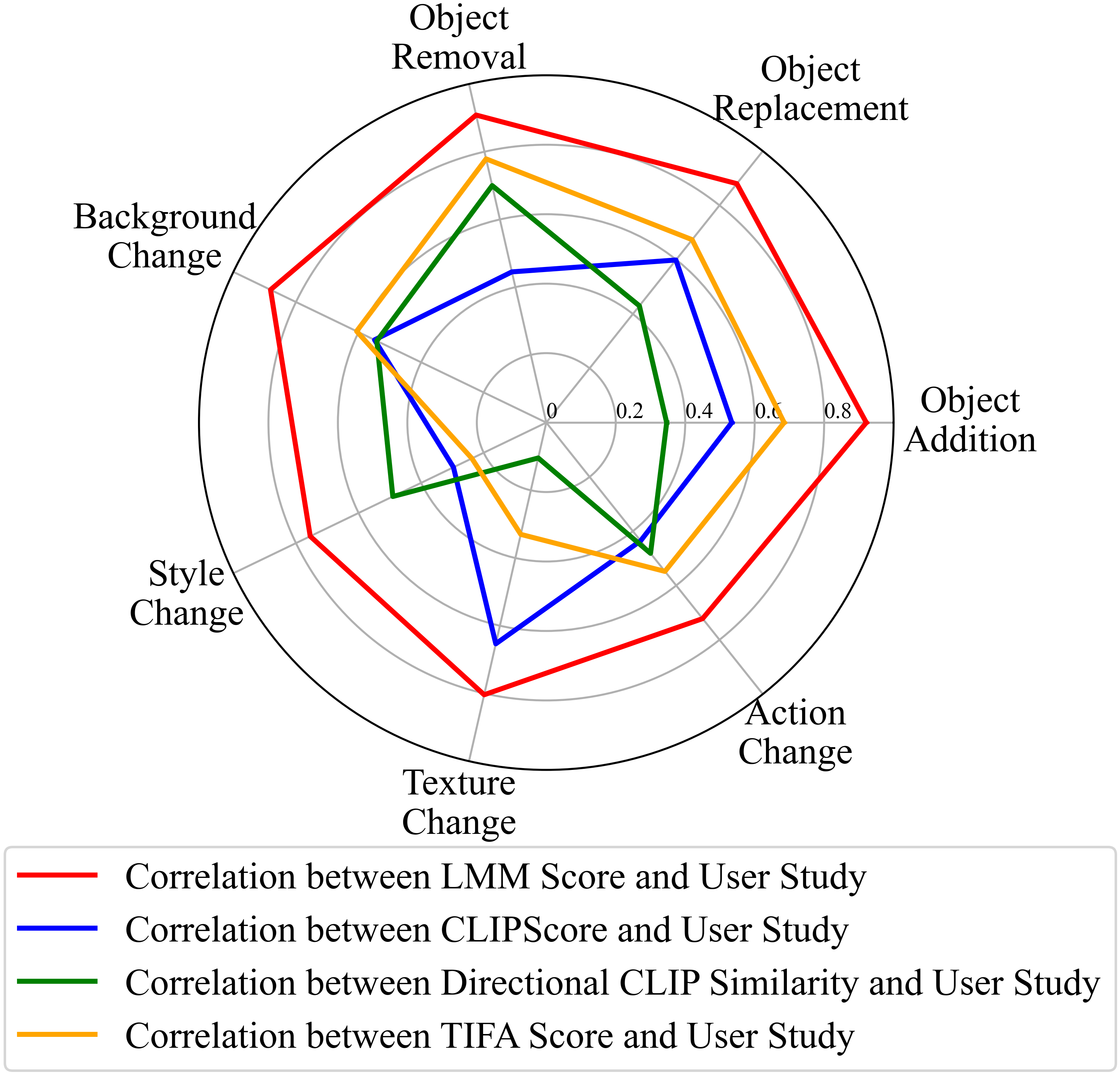}
            \vspace{-5pt}
		\caption{Comparison of Pearson correlation coefficients between four metrics and the user study.}
		\label{radar}
			\vspace{-10pt}
\end{figure}
\input{fig_comp}

\noindent \textbf{LMM Score vs. Other Metrics.} 
To further assess the effectiveness of LMM Score, we compare it against three other metrics—CLIPScore~\cite{hessel2021clipscore}, Directional CLIP Similarity~\cite{gal2022stylegan}, and TIFA Score~\cite{hu2023tifa}—by examining their Pearson correlation coefficients with the user study results. Unlike CLIPScore and Directional CLIP Similarity, which measure the feature distance between text and image directly in the CLIP feature space, TIFA Score evaluates the faithfulness of a generated image to its text input using visual question answering (VQA). Specifically, TIFA Score generates several question-answer pairs based on the text input using a language model and then determines image faithfulness by verifying whether the current VQA model can accurately answer these questions using the generated image.
Fig.~\ref{radar} presents these comparisons. It reveals that both CLIPScore and Directional CLIP Similarity show limited correlation with user judgments. While TIFA Score demonstrates better performance in semantic editing tasks such as object addition, it struggles with stylistic editing like style change. In contrast, LMM Score consistently exhibits the highest correlation with user evaluations across all tasks. This indicates that LMM Score more accurately captures the quality and relevance of image edits in alignment with human judgment, making it a more reliable metric for diverse editing scenarios.

%% file: fig_comp.tex
\begin{figure*}[p]
    \centering
    
\begin{tabular}{c c c c c c c c c}

\toprule
\multicolumn{9}{c}{\textbf{Editing Type:} \textit{Object Addition}\quad(\textit{``Add a glass of milk next to the cookies"})} \\
\begin{minipage}[t]{0.0885\linewidth}
\footnotesize
    \centerline{\includegraphics[width=1.98cm]{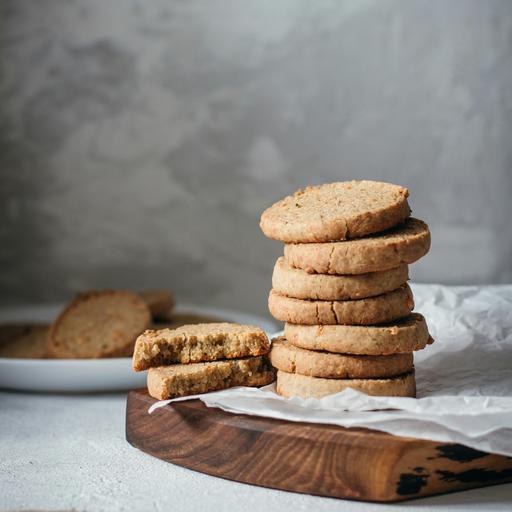}}
    \centerline{Source}
\end{minipage} &
\begin{minipage}[t]{0.0885\linewidth}
\footnotesize
    \centerline{\includegraphics[width=1.98cm]{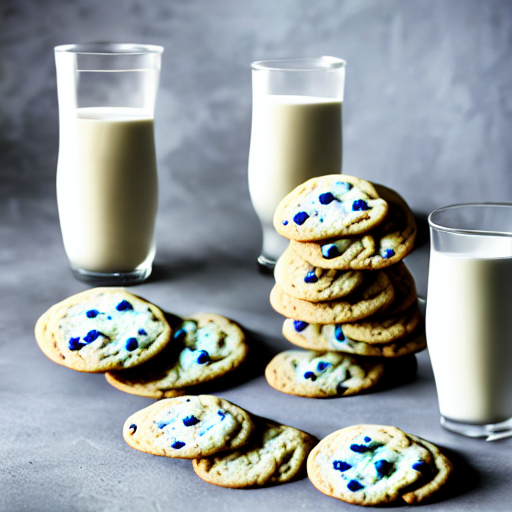}}
    \centerline{InstructPix2Pix \cite{brooks2023instructpix2pix}}
\end{minipage} &
\begin{minipage}[t]{0.0885\linewidth}
\footnotesize
    \centerline{\includegraphics[width=1.98cm]{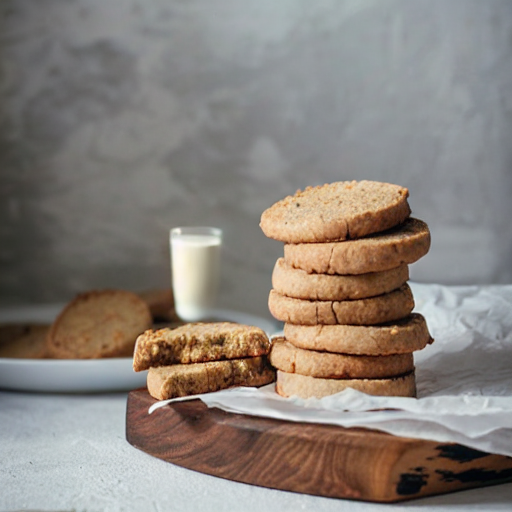}}
        \centerline{Instruct-}     \centerline{Diffusion \cite{geng2023instructdiffusion}}
\end{minipage} &
\begin{minipage}[t]{0.0885\linewidth}
\footnotesize
    \centerline{\includegraphics[width=1.98cm]{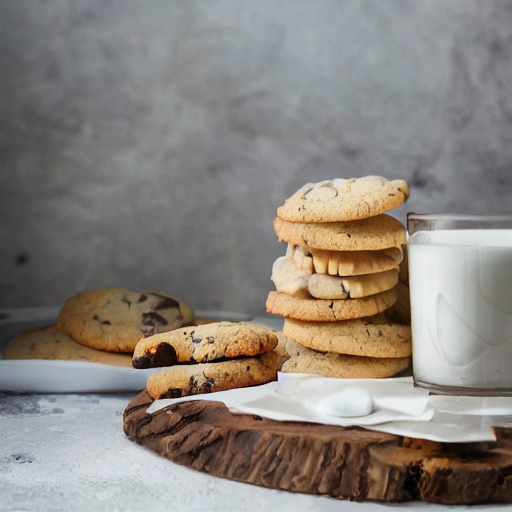}}
\centerline{DiffusionDisen-} \centerline{tanglement \cite{dong2023prompt}}
\end{minipage} &
\begin{minipage}[t]{0.0885\linewidth}
\footnotesize
    \centerline{\includegraphics[width=1.98cm]{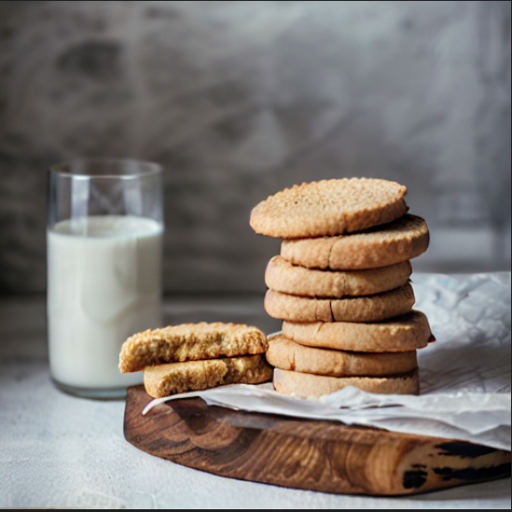}}
    \centerline{Imagic \cite{kawar2023imagic}}
\end{minipage} &
\begin{minipage}[t]{0.0885\linewidth}
\footnotesize
    \centerline{\includegraphics[width=1.98cm]{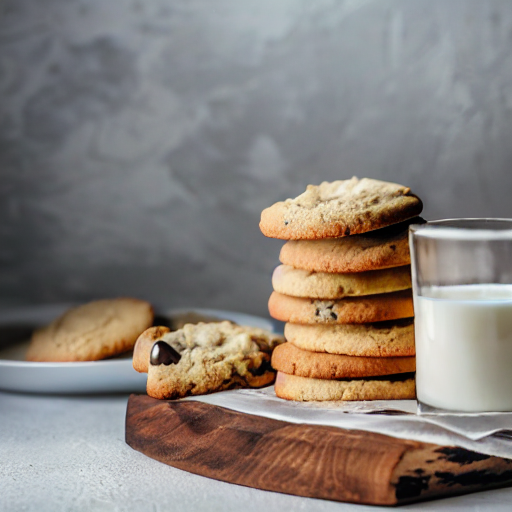}}
    \centerline{DDPM} \centerline{Inversion \cite{huberman2023edit}}
\end{minipage} &
\begin{minipage}[t]{0.0885\linewidth}
\footnotesize
    \centerline{\includegraphics[width=1.98cm]
    {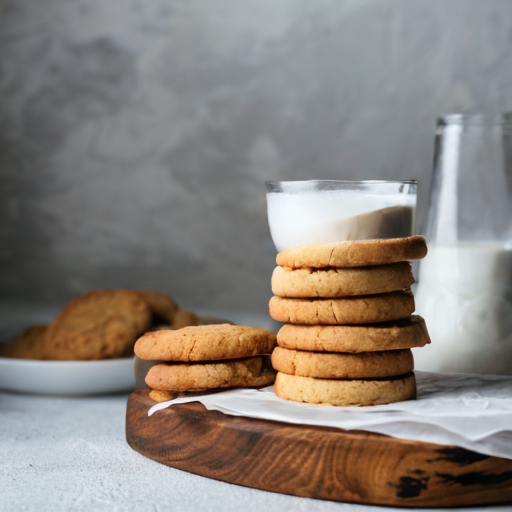}}
    \centerline{LEDITS++ \cite{brack2023ledits++}}
\end{minipage} &
\begin{minipage}[t]{0.0885\linewidth}
\footnotesize
    \centerline{\includegraphics[width=1.98cm]{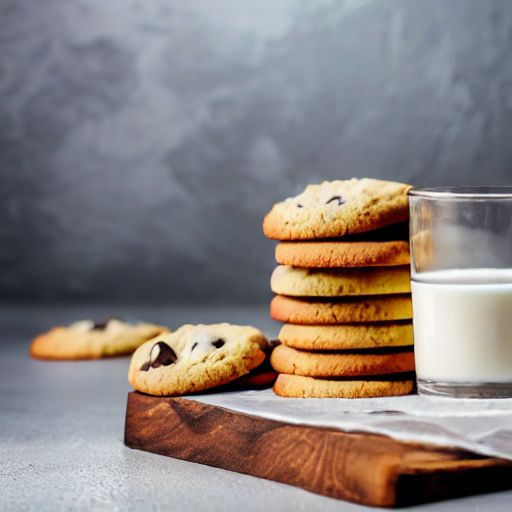}}
    \centerline{ProxEdit \cite{han2024proxedit}}
\end{minipage} &
\begin{minipage}[t]{0.0885\linewidth}
\footnotesize
    \centerline{\includegraphics[width=1.98cm]{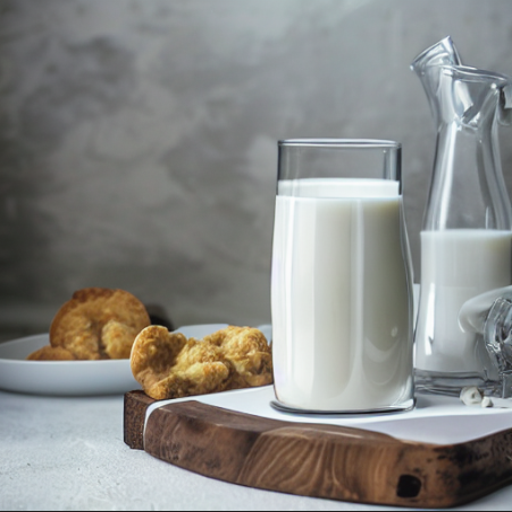}}
    \centerline{AIDI \cite{pan2023effective}}
\end{minipage} \\

\specialrule{0em}{1.0pt}{0.6pt}
\toprule
\end{tabular}
\vspace{0.05cm}

\begin{tabular}{c c c c c c}

\multicolumn{6}{c}{\textbf{Editing Type:} \textit{Object Removal}\quad(\textit{``Remove the wooden house and dock"})} \\
\begin{minipage}[t]{0.0885\linewidth}
\footnotesize
    \centerline{\includegraphics[width=1.98cm]{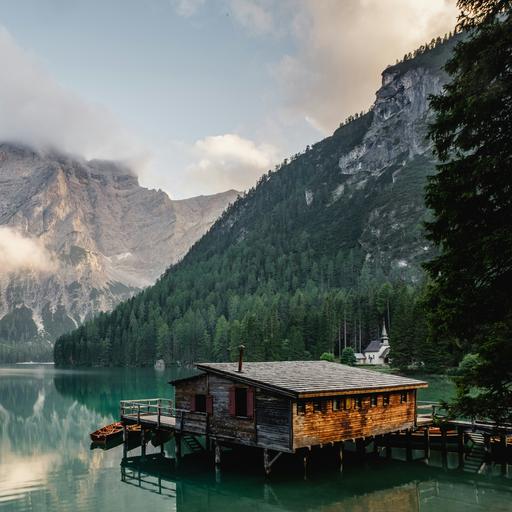}}
    \centerline{Source}
\end{minipage} &
\hspace{1.122cm}
\begin{minipage}[t]{0.0885\linewidth}
\footnotesize
    \centerline{\includegraphics[width=1.98cm]{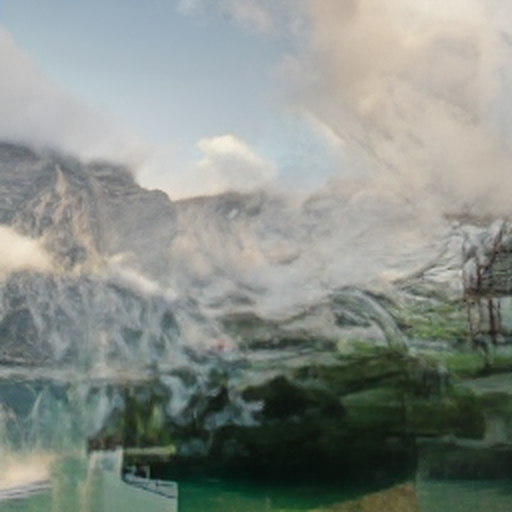}}
    \centerline{Inst-Paint \cite{yildirim2023inst}}
\end{minipage} &
\hspace{1.122cm}
\begin{minipage}[t]{0.0885\linewidth}
\footnotesize
    \centerline{\includegraphics[width=1.98cm]{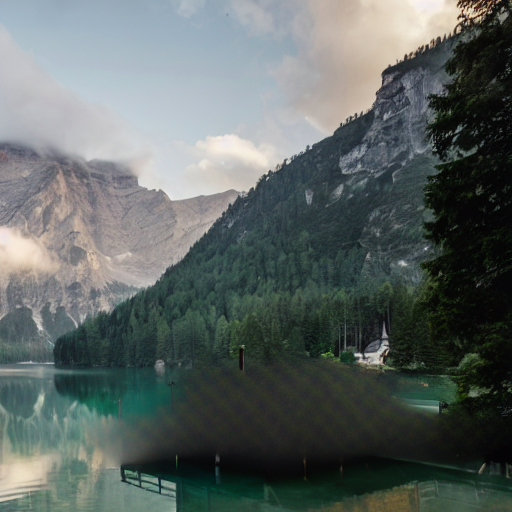}}
    \centerline{InstructDiffusion \cite{geng2023instructdiffusion}}
\end{minipage} &
\hspace{1.122cm}
\begin{minipage}[t]{0.0885\linewidth}
\footnotesize
    \centerline{\includegraphics[width=1.98cm]{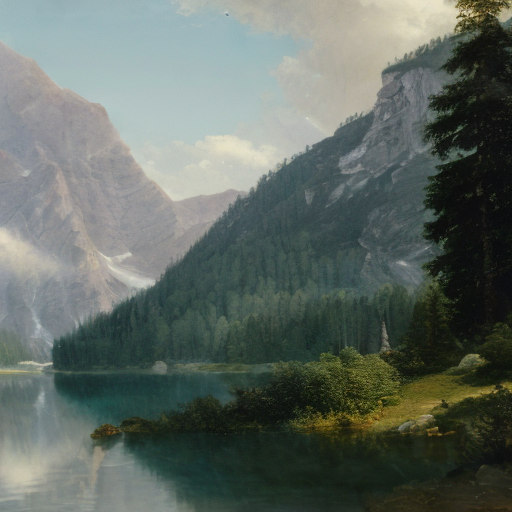}}
    \centerline{LEDITS++ \cite{brack2023ledits++}}
\end{minipage} &
\hspace{1.122cm}
\begin{minipage}[t]{0.0885\linewidth}
\footnotesize
    \centerline{\includegraphics[width=1.98cm]{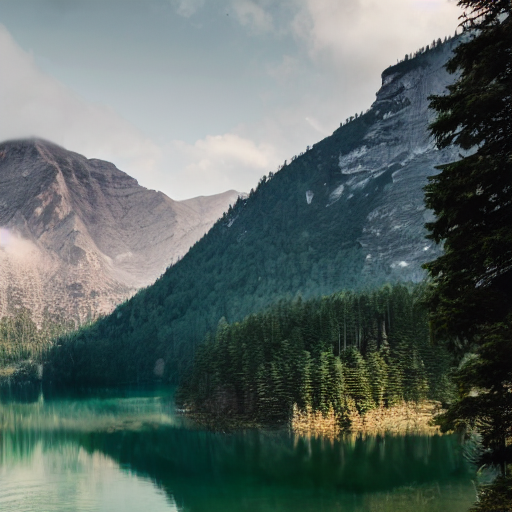}}
    \centerline{ProxEdit \cite{han2024proxedit}}
\end{minipage} &
\hspace{1.122cm}
\begin{minipage}[t]{0.0885\linewidth}
\footnotesize
    \centerline{\includegraphics[width=1.98cm]{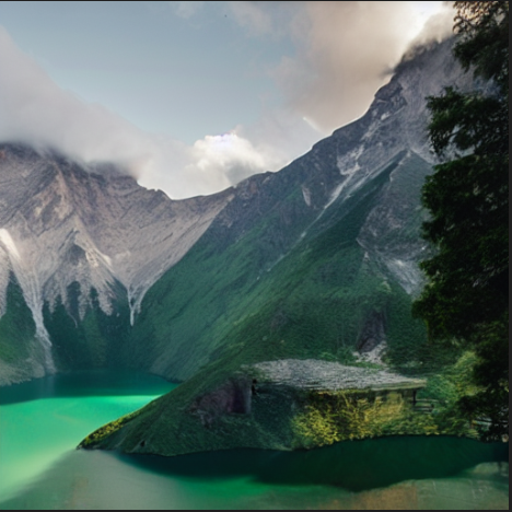}}
    \centerline{AIDI \cite{pan2023effective}}
\end{minipage}
\\
\specialrule{0em}{0.6pt}{0.6pt}
\toprule
\end{tabular}
\vspace{0.05cm}

\begin{tabular}{c c c c c c c c c}
\multicolumn{9}{c}{\textbf{Editing Type:} \textit{Object Replacement}\quad( \textit{``Replace the llama toy with a floor lamp"})} \\

\begin{minipage}[t]{0.0885\linewidth}
\footnotesize
    \centerline{\includegraphics[width=1.98cm]{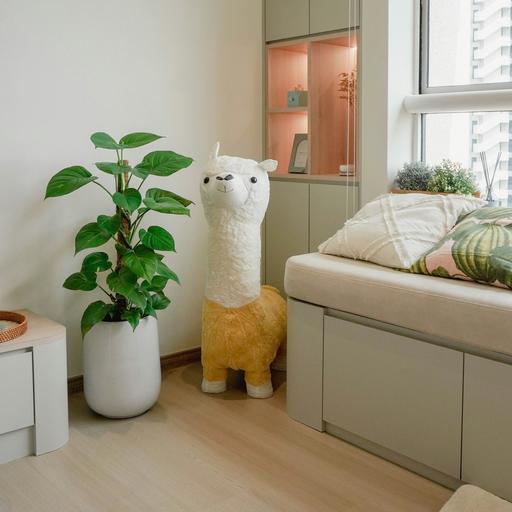}}
    \centerline{Source}
\end{minipage} &
\begin{minipage}[t]{0.0885\linewidth}
\footnotesize
    \centerline{\includegraphics[width=1.98cm]{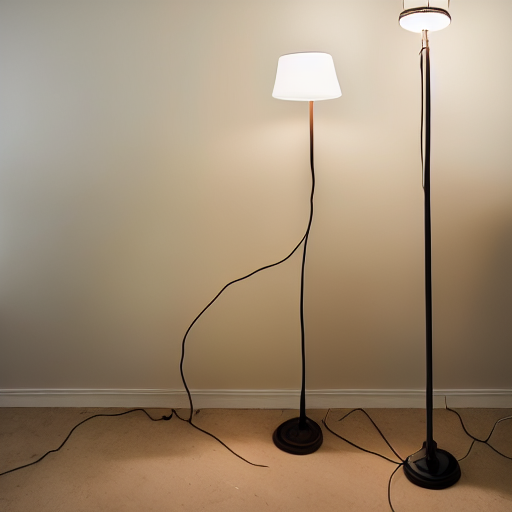}}
\centerline{InstructPix2Pix \cite{brooks2023instructpix2pix}}
\end{minipage} &
\begin{minipage}[t]{0.0885\linewidth}
\footnotesize
    \centerline{\includegraphics[width=1.98cm]{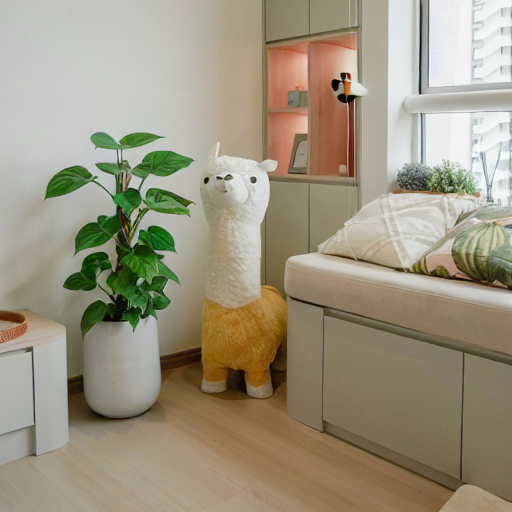}}
        \centerline{Instruct-}     \centerline{Diffusion \cite{geng2023instructdiffusion}}
\end{minipage} &
\begin{minipage}[t]{0.0885\linewidth}
\footnotesize
    \centerline{\includegraphics[width=1.98cm]{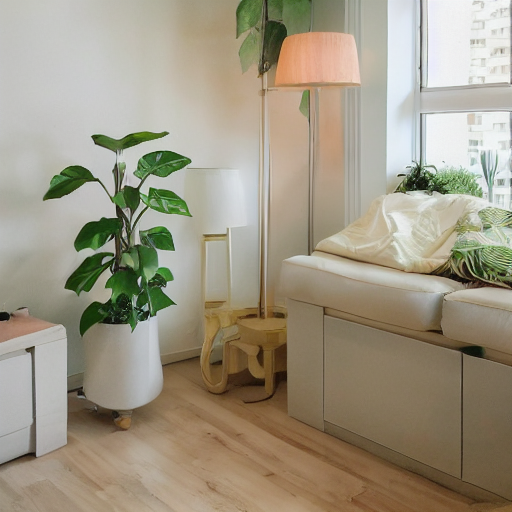}}
\centerline{DiffusionDisen-} \centerline{tanglement \cite{dong2023prompt}}
\end{minipage} &
\begin{minipage}[t]{0.0885\linewidth}
\footnotesize
    \centerline{\includegraphics[width=1.98cm]{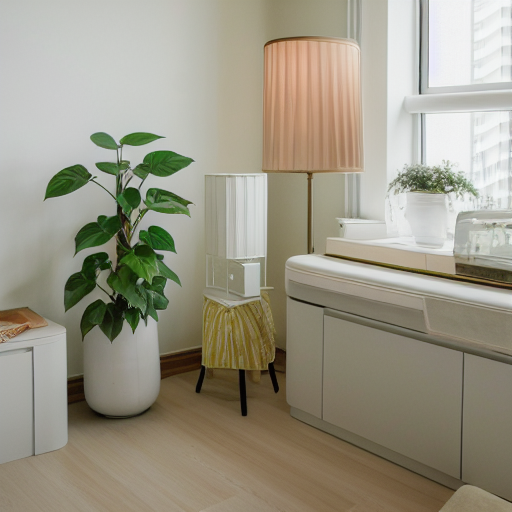}}
    \centerline{Null-Text \cite{mokady2023null}}
\end{minipage} &
\begin{minipage}[t]{0.0885\linewidth}
\footnotesize
    \centerline{\includegraphics[width=1.98cm]{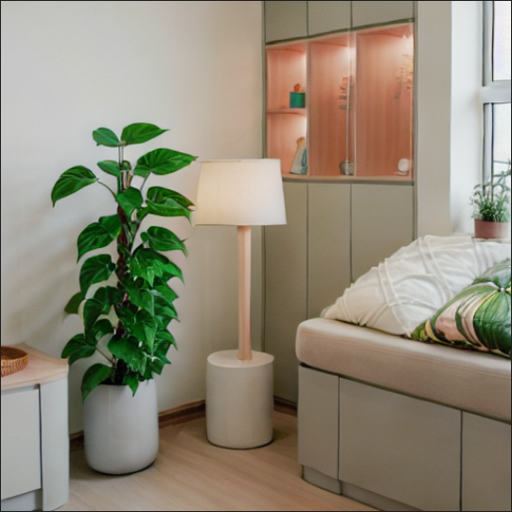}}
    \centerline{Imagic \cite{kawar2023imagic}}
\end{minipage} &
\begin{minipage}[t]{0.0885\linewidth}
\footnotesize
    \centerline{\includegraphics[width=1.98cm]
    {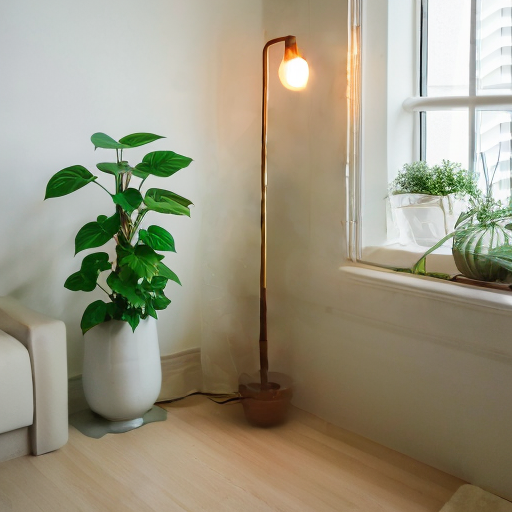}}
    \centerline{LEDITS++ \cite{brack2023ledits++}}
\end{minipage} &
\begin{minipage}[t]{0.0885\linewidth}
\footnotesize
    \centerline{\includegraphics[width=1.98cm]{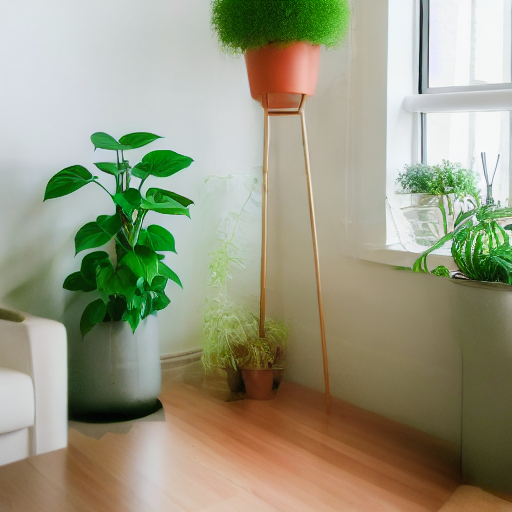}}
    \centerline{ProxEdit \cite{han2024proxedit}}
\end{minipage} &
\begin{minipage}[t]{0.0885\linewidth}
\footnotesize
    \centerline{\includegraphics[width=1.98cm]{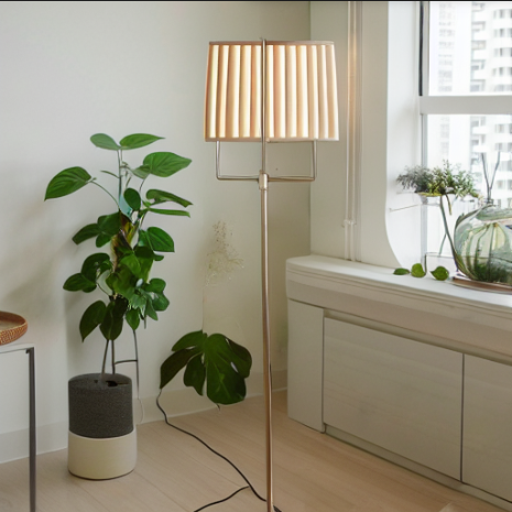}}
    \centerline{AIDI \cite{pan2023effective}}
\end{minipage} \\
\specialrule{0em}{0.6pt}{0.6pt}
\toprule
\end{tabular}
\vspace{0.05cm}

\begin{tabular}{c c c c c c c c c}
\multicolumn{9}{c}{\textbf{Editing Type:} \textit{Background Change}\quad (\textit{``Change the city street to a dense jungle"})} \\

\begin{minipage}[t]{0.0885\linewidth}
\footnotesize
    \centerline{\includegraphics[width=1.98cm]{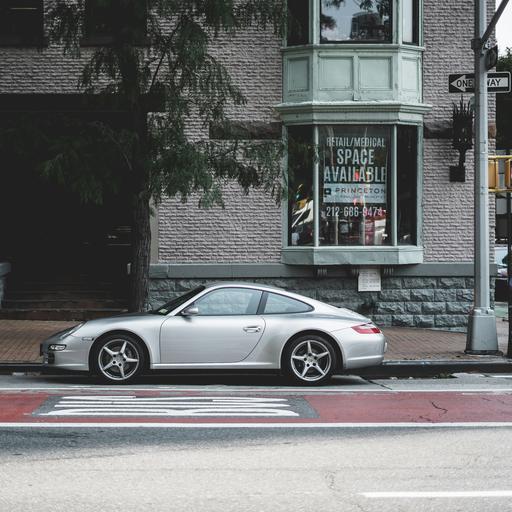}}
    \centerline{Source}
\end{minipage} &
\begin{minipage}[t]{0.0885\linewidth}
\footnotesize
    \centerline{\includegraphics[width=1.98cm]{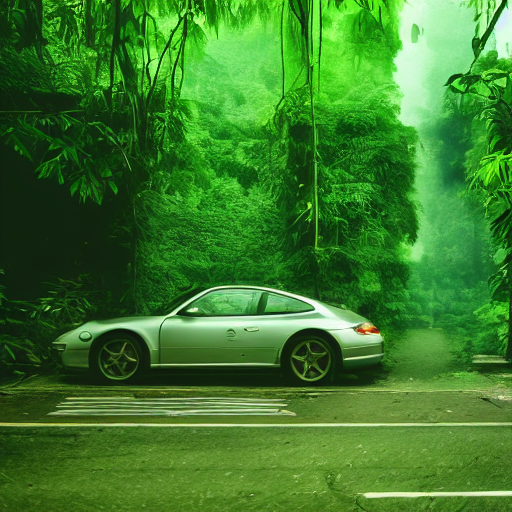}}
\centerline{InstructPix2Pix \cite{brooks2023instructpix2pix}}
\end{minipage} &
\begin{minipage}[t]{0.0885\linewidth}
\footnotesize
    \centerline{\includegraphics[width=1.98cm]{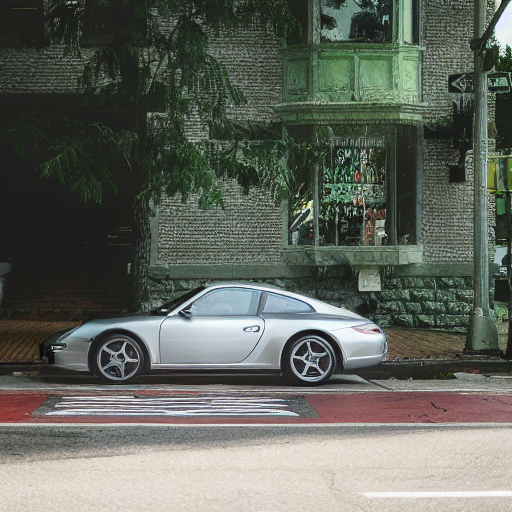}}
    \centerline{DDS \cite{hertz2023delta}}
\end{minipage} &
\begin{minipage}[t]{0.0885\linewidth}
\footnotesize
    \centerline{\includegraphics[width=1.98cm]{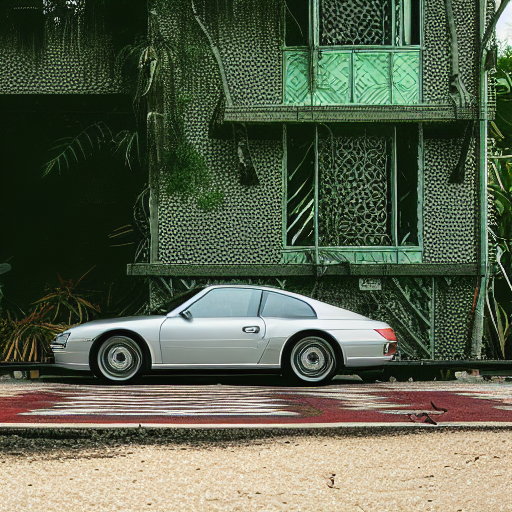}}
    \centerline{Null-Text \cite{mokady2023null}}
\end{minipage} &
\begin{minipage}[t]{0.0885\linewidth}
\footnotesize
    \centerline{\includegraphics[width=1.98cm]{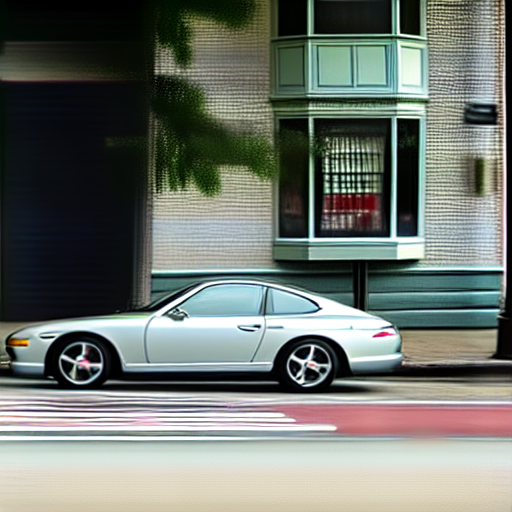}}
    \centerline{SINE \cite{zhang2023sine}}
\end{minipage} &
\begin{minipage}[t]{0.0885\linewidth}
\footnotesize
    \centerline{\includegraphics[width=1.98cm]{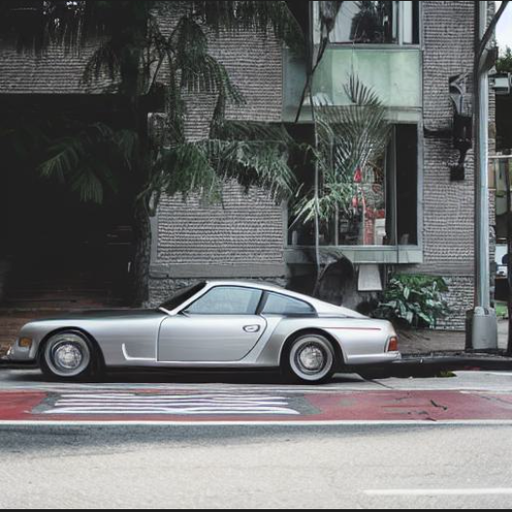}}
    \centerline{EDICT \cite{wallace2023edict}}
\end{minipage} &
\begin{minipage}[t]{0.0885\linewidth}
\footnotesize
    \centerline{\includegraphics[width=1.98cm]
    {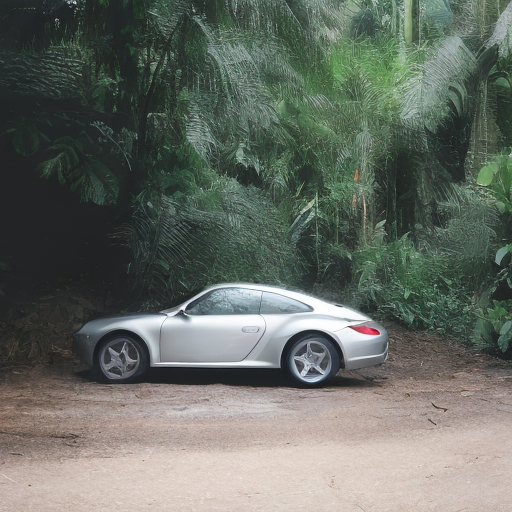}}
    \centerline{LEDITS++ \cite{brack2023ledits++}}
\end{minipage} &
\begin{minipage}[t]{0.0885\linewidth}
\footnotesize
    \centerline{\includegraphics[width=1.98cm]{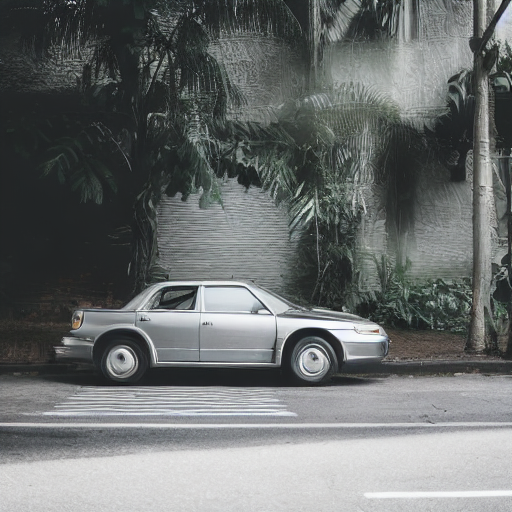}}
    \centerline{ProxEdit \cite{han2024proxedit}}
\end{minipage} &
\begin{minipage}[t]{0.0885\linewidth}
\footnotesize
    \centerline{\includegraphics[width=1.98cm]{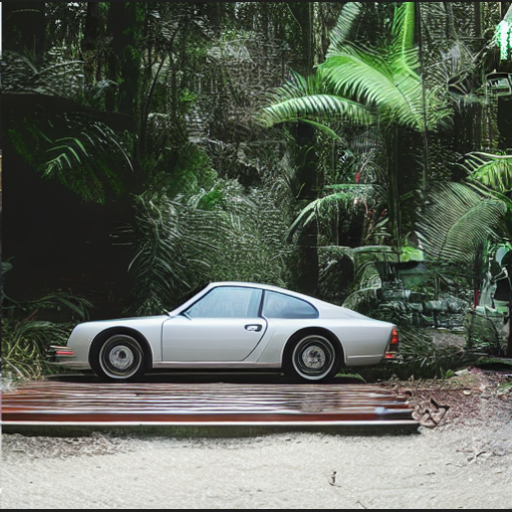}}
    \centerline{AIDI \cite{pan2023effective}}
\end{minipage} \\
\specialrule{0em}{0.6pt}{0.6pt}
\toprule
\end{tabular}
\vspace{0.05cm}


\begin{tabular}{c c c c c c c c c}

\multicolumn{9}{c}{\textbf{Editing Type:} \textit{Style Change}\quad(\textit{``Transform it to the Van Gogh style"})} \\

\begin{minipage}[t]{0.0885\linewidth}
\footnotesize
    \centerline{\includegraphics[width=1.98cm]{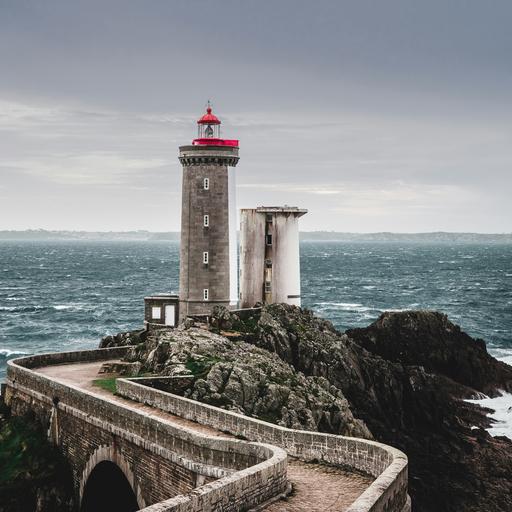}}
    \centerline{Source}
\end{minipage} &
\begin{minipage}[t]{0.0885\linewidth}
\footnotesize
    \centerline{\includegraphics[width=1.98cm]{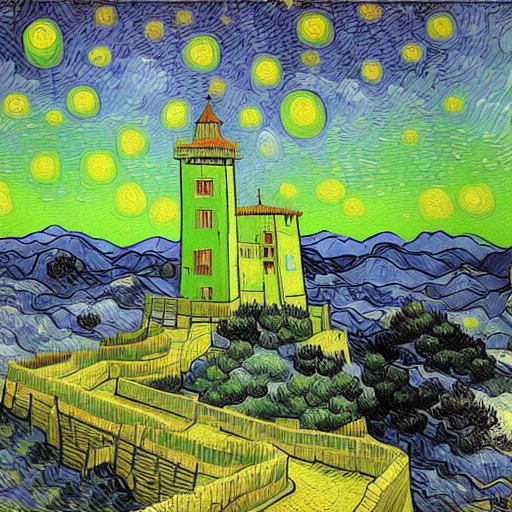}}
\centerline{Instruct} \centerline{Pix2Pix \cite{brooks2023instructpix2pix}}
\end{minipage} &
\begin{minipage}[t]{0.0885\linewidth}
\footnotesize
    \centerline{\includegraphics[width=1.98cm]{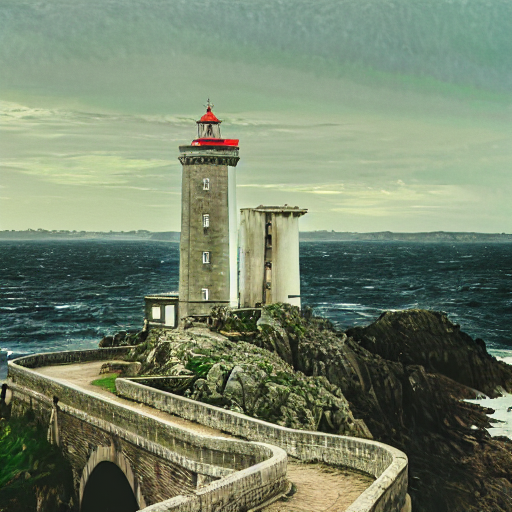}}
        \centerline{Instruct-}     \centerline{Diffusion \cite{geng2023instructdiffusion}}
\end{minipage} &
\begin{minipage}[t]{0.0885\linewidth}
\footnotesize
    \centerline{\includegraphics[width=1.98cm]{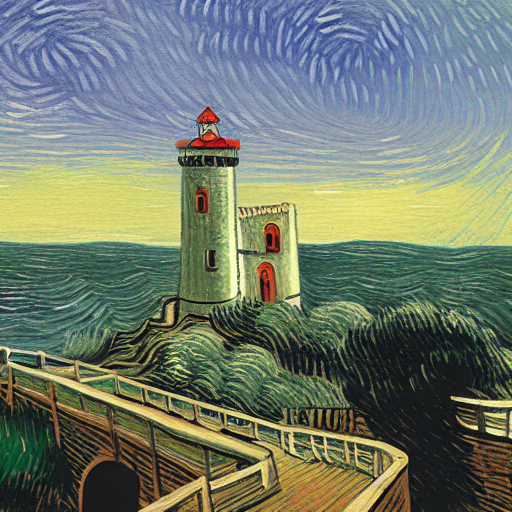}}
    \centerline{Null-Text \cite{mokady2023null}}
\end{minipage} &
\begin{minipage}[t]{0.0885\linewidth}
\footnotesize
    \centerline{\includegraphics[width=1.98cm]{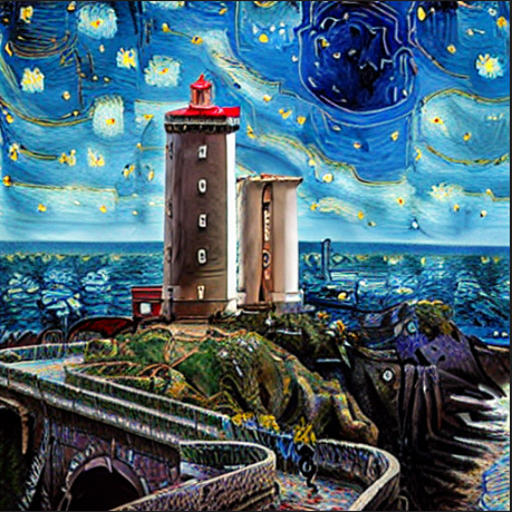}}
    \centerline{Imagic \cite{kawar2023imagic}}
\end{minipage} &
\begin{minipage}[t]{0.0885\linewidth}
\footnotesize
    \centerline{\includegraphics[width=1.98cm]{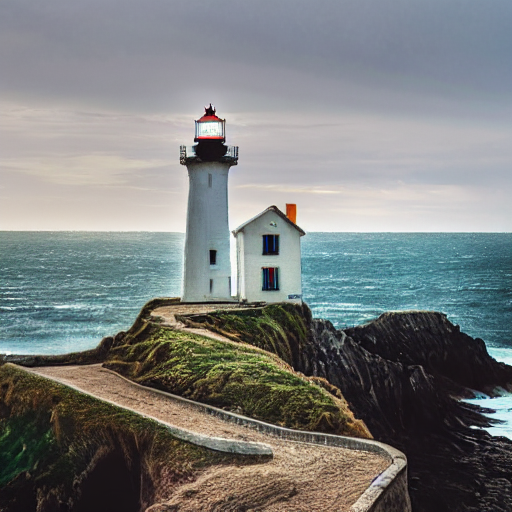}}
    \centerline{DDPM} \centerline{Inversion \cite{huberman2023edit}}
\end{minipage} &
\begin{minipage}[t]{0.0885\linewidth}
\footnotesize
    \centerline{\includegraphics[width=1.98cm]
    {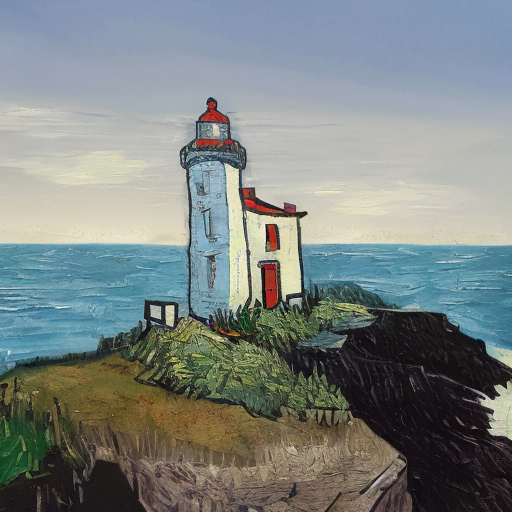}}
    \centerline{LEDITS++ \cite{brack2023ledits++}}
\end{minipage} &
\begin{minipage}[t]{0.0885\linewidth}
\footnotesize
    \centerline{\includegraphics[width=1.98cm]{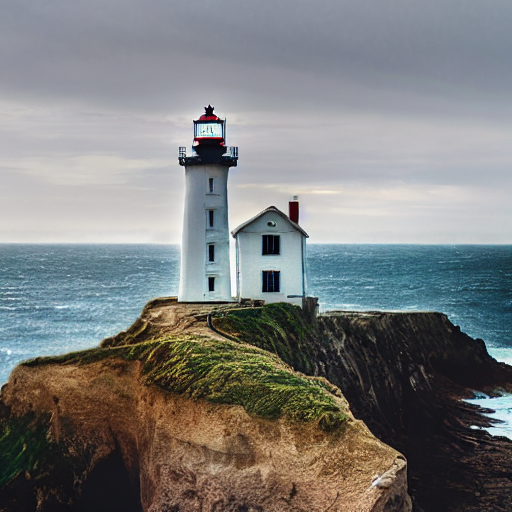}}
    \centerline{ProxEdit \cite{han2024proxedit}}
\end{minipage} &
\begin{minipage}[t]{0.0885\linewidth}
\footnotesize
    \centerline{\includegraphics[width=1.98cm]{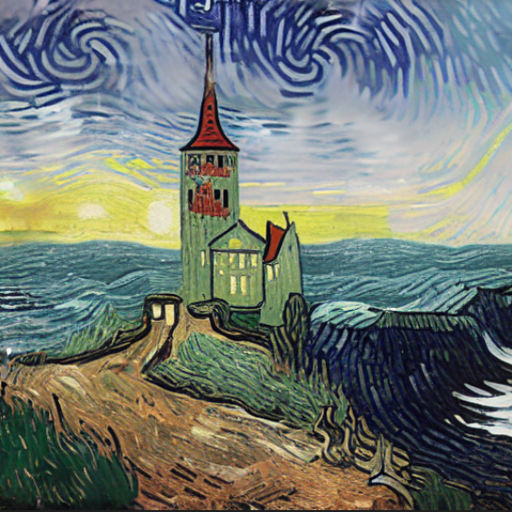}}
    \centerline{AIDI \cite{pan2023effective}}
\end{minipage} \\
\specialrule{0em}{0.6pt}{0.6pt}
\toprule
\end{tabular}
\vspace{0.05cm}

\begin{tabular}{c c c c c c c c c}

\multicolumn{9}{c}{\textbf{Editing Type:} \textit{Texture Change}\quad(\textit{``Turn the horse into a statue"})} \\

\begin{minipage}[t]{0.0885\linewidth}
\footnotesize
    \centerline{\includegraphics[width=1.98cm]{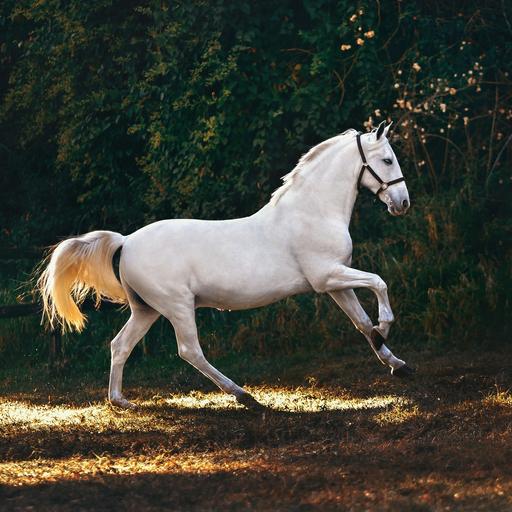}}
    \centerline{Source}
\end{minipage} &
\begin{minipage}[t]{0.0885\linewidth}
\footnotesize
    \centerline{\includegraphics[width=1.98cm]{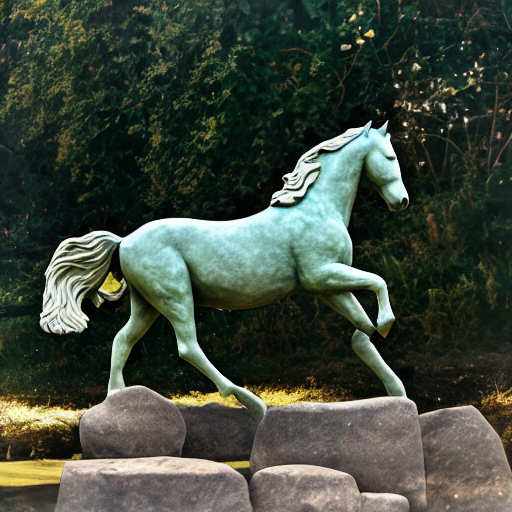}} \centerline{InstructPix2Pix \cite{brooks2023instructpix2pix}}
\end{minipage} &
\begin{minipage}[t]{0.0885\linewidth}
\footnotesize
    \centerline{\includegraphics[width=1.98cm]{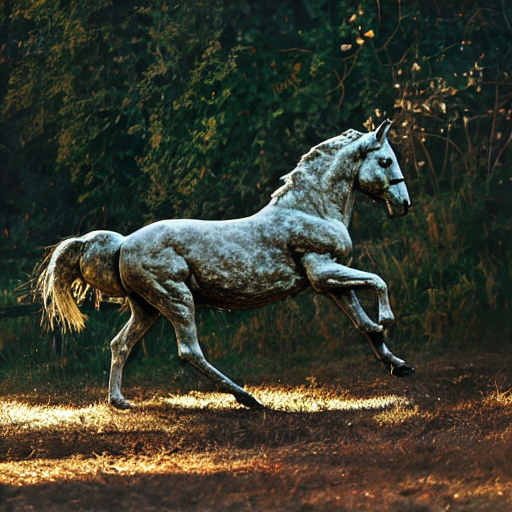}}
        \centerline{Instruct-}     \centerline{Diffusion \cite{geng2023instructdiffusion}}
\end{minipage} &
\begin{minipage}[t]{0.0885\linewidth}
\footnotesize
    \centerline{\includegraphics[width=1.98cm]{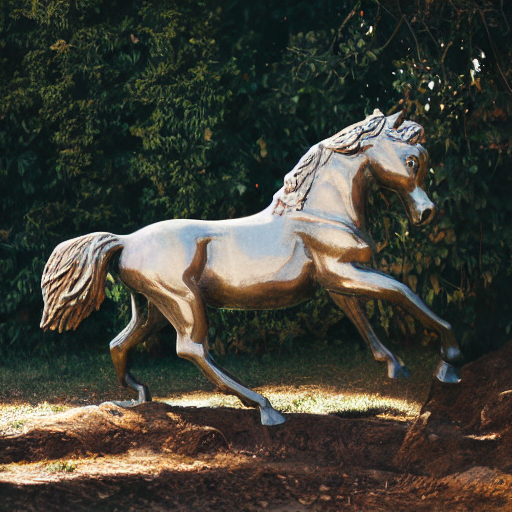}}
\centerline{DiffusionDisen-} \centerline{tanglement \cite{dong2023prompt}}
\end{minipage} &
\begin{minipage}[t]{0.0885\linewidth}
\footnotesize
    \centerline{\includegraphics[width=1.98cm]{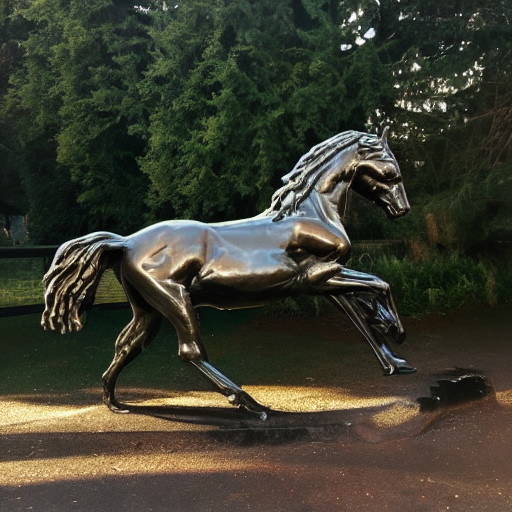}}
    \centerline{Null-Text \cite{mokady2023null}}
\end{minipage} &
\begin{minipage}[t]{0.0885\linewidth}
\footnotesize
    \centerline{\includegraphics[width=1.98cm]{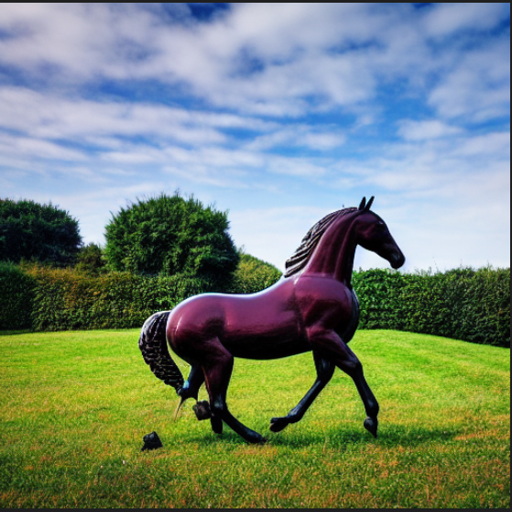}}
    \centerline{Imagic \cite{kawar2023imagic}}
\end{minipage} &
\begin{minipage}[t]{0.0885\linewidth}
\footnotesize
    \centerline{\includegraphics[width=1.98cm]{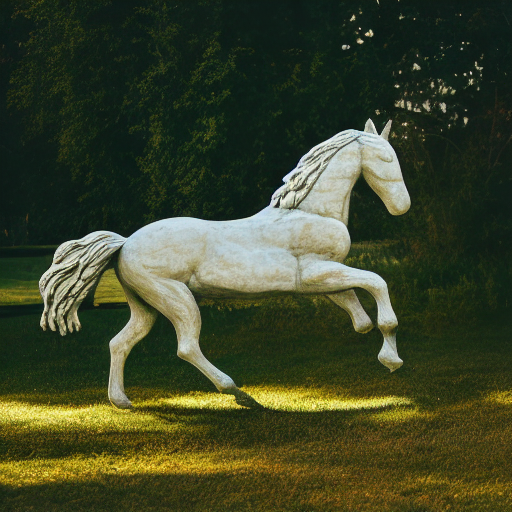}}
    \centerline{PnP \cite{tumanyan2023plug}}
\end{minipage} &
\begin{minipage}[t]{0.0885\linewidth}
\footnotesize
    \centerline{\includegraphics[width=1.98cm]
    {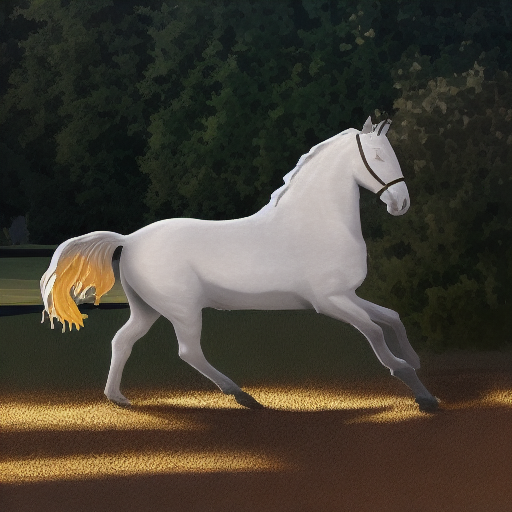}}
    \centerline{EBMs \cite{park2023energy}}
\end{minipage} &
\begin{minipage}[t]{0.0885\linewidth}
\footnotesize
    \centerline{\includegraphics[width=1.98cm]{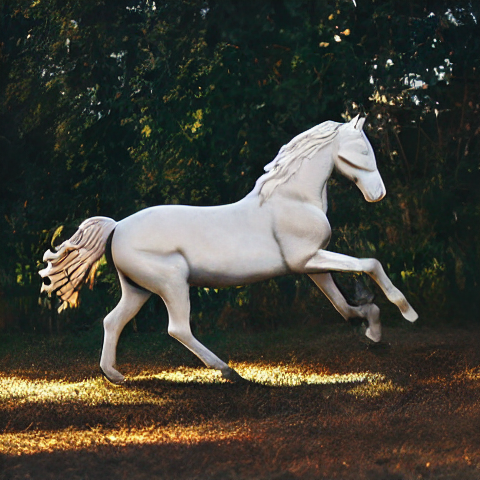}}
  \centerline{CycleDiffusion \cite{wu2023latent}}
\end{minipage} \\
\specialrule{0em}{0.6pt}{0.6pt}
\toprule
\end{tabular}
\vspace{0.05cm}

\begin{tabular}{c c c c c}

\multicolumn{5}{c}{\textbf{Editing Type:} \textit{Action Change}\quad(\textit{``Let the bear raise its hand"})} \\
\begin{minipage}[t]{0.0885\linewidth}
\footnotesize
    \centerline{\includegraphics[width=1.98cm]{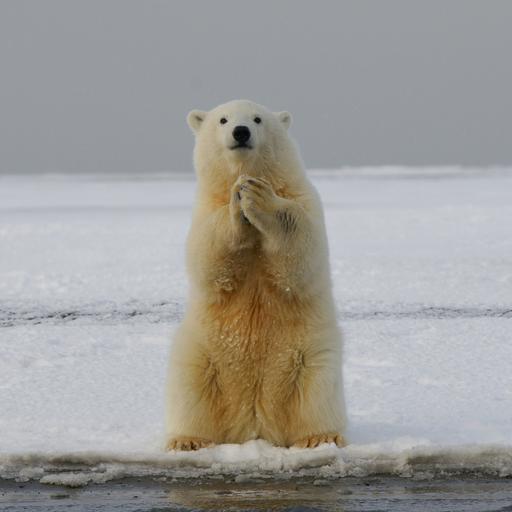}}
    \centerline{Source}
\end{minipage} &  
\hspace{1.93cm}
\begin{minipage}[t]{0.0885\linewidth}
\footnotesize
    \centerline{\includegraphics[width=1.98cm]{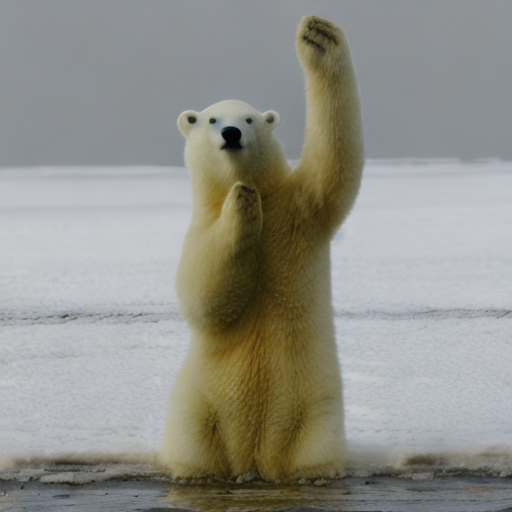}}
    \centerline{Forgedit \cite{zhang2023forgedit}}
\end{minipage} &  
\hspace{1.93cm}
\begin{minipage}[t]{0.0885\linewidth}
\footnotesize
    \centerline{\includegraphics[width=1.98cm]{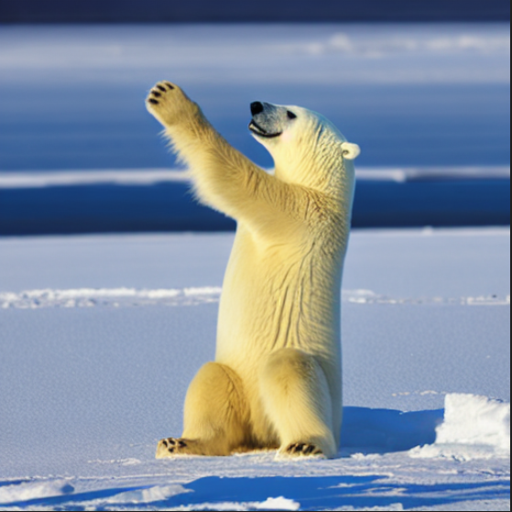}}
    \centerline{Imagic \cite{kawar2023imagic}}
\end{minipage} &  
\hspace{1.93cm}
\begin{minipage}[t]{0.0885\linewidth}
\footnotesize
    \centerline{\includegraphics[width=1.98cm]{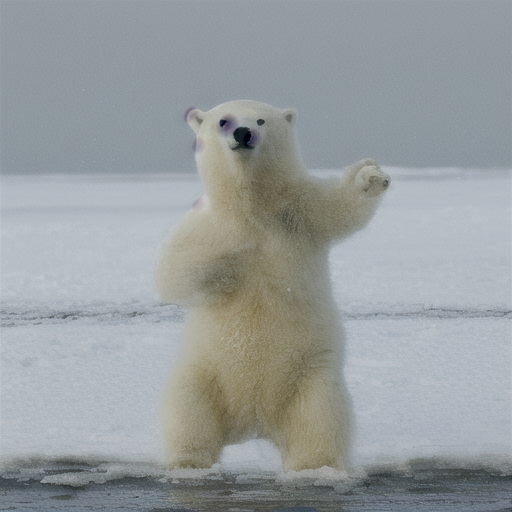}}
    \centerline{MasaCtrl \cite{cao2023masactrl}}
\end{minipage} &  
\hspace{1.93cm}
\begin{minipage}[t]{0.0885\linewidth}
\footnotesize
    \centerline{\includegraphics[width=1.98cm]
    {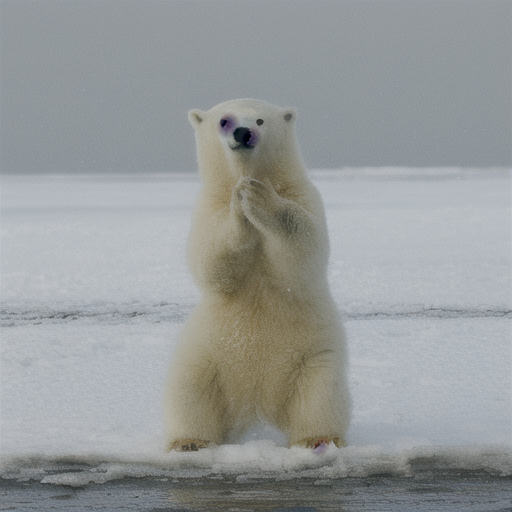}}
    \centerline{TIC \cite{duan2023tuning}}
\end{minipage} 
\\
\toprule
\specialrule{0em}{0.6pt}{0.6pt}
\end{tabular}
\caption{Visual comparison on 7 selected editing types.}
\label{fig_visual_editing}
\end{figure*}

%% file: sec/sec9_future.tex
\section{Challenges and Future Directions}\label{future}
Despite the success achieved in image editing with diffusion models, there are still limitations to address in future work.

\noindent \textbf{Fewer-Step Model Inference.} 
Most diffusion-based models require a significant number of steps to obtain the final image during inference, which is both time-consuming and computationally costly, bringing challenges in model deployment and user experience. To improve the inference efficiency, few-step or one-step generation diffusion models have been studied~\cite{sauer2023adversarial,geng2023one,salimans2022progressive}. Recent methods in this field reduce the number of steps by distilling knowledge from a pretrained strong diffusion model so that the few-step model can mimic the behavior of the strong model. A more interesting yet challenging direction is to develop few-step models directly without relying on the pretrained models, such as consistency models~\cite{song2023consistency,luo2023latent}. 

\noindent \textbf{Efficient Models.} 
Training a diffusion model that can generate realistic results is computationally intensive and requires a large amount of high-quality data. This complexity makes developing diffusion models for image editing very challenging. To reduce the training cost, recent works design more efficient network architectures as the backbones of diffusion models~\cite{zhao2023mobilediffusion,li2023snapfusion}. Besides, another important direction is to train only a portion of the parameters or freeze the original parameters and add a few new layers on top of the pretrained diffusion model~\cite{hu2021lora,shi2023instantbooth}. 

\noindent \textbf{Complex Object Structure Editing.} 
Existing works can synthesize realistic colors, styles or textures when editing images. However, they still produce noticeable artifacts when dealing with complex structures, such as fingers, logos, and scene text. Attempts have been made to address these issues. Previous methods usually use negative prompting such as ``six fingers, bad leg, etc.'' to make the model avoid generating such images, which are effective in certain cases but not robust enough~\cite{schramowski2023safe}. Recent works start to use layouts, edges, or dense labels as guidance for editing the global or local structures of images~\cite{zhang2023adding,qin2023unicontrol}.

\noindent \textbf{Complex Lighting and Shadow Editing.} 
Editing the lighting or illumination of an image remains a challenging task~\cite{ren2023relightful}, especially when aiming for realistic and consistent results. Traditional methods like Total Relighting~\cite{pandey2021total} estimate normals, albedo, and shading to achieve realistic effects. Recent diffusion-based methods have advanced this area. For instance, DiffusionRig \cite{ding2023diffusionrig} and DiFaReli \cite{ponglertnapakorn2023difareli} estimate 3D parameters (e.g., shape, expression, SH lighting) from portraits and use these to condition diffusion models for face relighting. Other approaches \cite{papantoniou2023relightify, zhang2024facednerf} reconstruct relightable 3D face models from a single image using diffusion models, enabling 3D editing. However, since these methods focus primarily on portrait relighting, their applicability to broader natural images is limited.
More recent methods \cite{bashkirova2023lasagna, zeng2024dilightnet, kocsis2024lightit, jin2024neural} focus on training image-based, lighting-conditioned diffusion models on synthetic datasets. Specifically, Lasagna \cite{bashkirova2023lasagna} is trained on a self-rendered 3D object dataset relit from multiple light source locations to learn two lighting layers for luminosity adjustment. Similarly, both DiLightNet \cite{zeng2024dilightnet} and Neural Gaffer \cite{jin2024neural} train light-conditioned ControlNet models on synthetic datasets, where 3D objects are rendered with different lighting using Blender. Although these methods are not restricted to portrait relighting, they still rely heavily on strong lighting priors from 3D models. LightIt \cite{kocsis2024lightit} focuses on outdoor relighting using shading maps estimated by determining the Sun’s direction, but it is less suitable for indoor scenes.
In contrast, Retinex-Diffusion \cite{xing2024retinex} offers a training-free approach by reformulating the energy function of diffusion models based on the Retinex theory, allowing for lighting alterations without the need for 3D priors. Similarly, ShadowDiffusion~\cite{guo2023shadowdiffusion} explores generating visually pleasing shadows of objects using degradation priors. However, accurately editing the lighting or shadow of an object under different background conditions using diffusion models remains an unsolved problem.

\noindent \textbf{Unrobustness of Image Editing.}
Existing diffusion-based image editing models can synthesize realistic visual contents for a portion of given conditions. However, they still fail in many real-world scenarios~\cite{kawar2023imagic}. The fundamental cause for this problem is that the model is not capable of accurately modeling all possible samples in the conditional distribution space. How to improve the models to generate artifact-free contents consistently remains a challenge.  There are several ways to relieve this problem. First, scale up the data for model training to cover the challenging scenarios. This is an effective yet costly approach. In some situations, it is even very challenging to collect sufficient amount of data, such as medical images, visual inspection data, etc.  Second, adapt the model to accept more conditions such as structural guidance~\cite{zhang2023adding}, 3D-aware guidance~\cite{xiang20233d}, and textual guidance for more controllable and deterministic content creation. Third, adopt iterative refinement or multi-stage training to improve the initial results of the model progressively~\cite{saharia2022image,lugmayr2022repaint,ackermann2022high}.

\noindent \textbf{High-Resolution Image Generation and Editing.}
High-resolution images, typically those with resolutions of 1024x1024 pixels or higher, have become increasingly important in various applications. While several diffusion-based models focus on high-resolution image generation, such as SDXL \cite{podell2023sdxl} and PixArt \cite{chen2023pixart}, specialized diffusion models for high-resolution image editing are relatively scarce. Some studies \cite{sheynin2024emu,zhang2023sine,kawar2023imagic} have claimed the capability to perform high-resolution image editing, but few have demonstrated concrete results.
One of the key challenges in high-resolution image editing is ensuring precise modifications. The model must achieve a high degree of accuracy, particularly in tasks involving pixel-level adjustments and edge refinement, where maintaining fine details is essential. This level of precision becomes increasingly demanding with higher resolutions.
Another major challenge is maintaining computational efficiency. As resolution increases, the demand for GPU memory and processing power grows substantially, which puts pressure on both hardware resources and the model’s runtime performance. Therefore, it becomes essential to strike a balance between precision and speed to ensure the editing process remains efficient and smooth.

\noindent \textbf{Faithful Evaluation Metrics.}
Accurate evaluation is crucial for image editing to ensure that the edited contents are well aligned with the given conditions. However, although some quantitative metrics such as FID~\cite{heusel2017gans}, KID~\cite{binkowski2018demystifying}, LPIPS~\cite{zhang2018unreasonable}, CLIP Score~\cite{radford2021learning}, PSNR, and SSIM have been used for this task as a reference, most of existing works still heavily rely on user study to provide relatively accurate perceptual evaluation for visual results, which is neither efficient nor scalable. Faithful quantitative evaluation metrics are still an open problem. Recently, more accurate metrics for quantifying the perceptual similarity of objects have been proposed.  DreamSim~\cite{fu2023dreamsim} measures the mid-level similarity of two images considering layout, pose, and semantic contents and outperforms LPIPS. Similarly, foreground feature averaging (FFA)~\cite{kotar2024these} provides a simple yet effective method for measuring the similarity of objects despite its pose, viewpoint, lighting conditions, or background. In this paper, we also propose an effective image editing metric LMM Score with the help of an LMM.

\section{Conclusion}\label{conclusion}
We have extensively overviewed diffusion model-based image editing methods, examining the field from multiple perspectives. Our analysis begins by categorizing over 100 methods into three main groups according to their learning strategies: training-based, test-time fine-tuning, and training and fine-tuning free approaches. We then classify image editing tasks into three distinct categories: semantic, stylistic, and structural editing, encompassing 12 specific types in total. We explore these methods and their contributions towards enhancing editing performance. An evaluation of 7 tasks alongside recent state-of-the-art methods is conducted within our image editing benchmark EditEval. Additionally, a new metric LMM Score 
is introduced for comparative analysis of these methods. Concluding our review, we highlight the broad potential within the image editing domain and suggest directions for future research.